\documentclass{article}

\usepackage{log_2022}            % for camera-ready version
\usepackage[numbers,compress,sort]{natbib}
\usepackage[utf8]{inputenc} % allow utf-8 input
\usepackage[T1]{fontenc}    % use 8-bit T1 fonts
\usepackage{url}            % simple URL typesetting
\usepackage{booktabs}       % professional-quality tables
\usepackage{amsfonts}       % blackboard math symbols
\usepackage{nicefrac}       % compact symbols for 1/2, etc.
\usepackage{microtype}      % microtypography
\usepackage[table]{xcolor}  % colors

\usepackage{mathtools}
\usepackage{caption}
\usepackage{subcaption}
\usepackage{lipsum}
\usepackage{graphicx}
\usepackage{flushend}
\usepackage{changepage,threeparttable} % for wide tables
\usepackage{wrapfig}
\usepackage{flushend}
\usepackage{enumitem}

\usepackage{algorithm}
\usepackage{algpseudocode}
\usepackage{amsmath}
\usepackage{MnSymbol}
\usepackage{multirow}
\usepackage{graphbox}
\usepackage{duckuments}         % sample images

\graphicspath{{figures/}}

\newcommand{\myparagraph}[1]{\smallbreak\noindent\textbf{#1}}
\newcommand{\pert}[1]{\emph{#1}}
\newcommand{\ds}[1]{\texttt{#1}}
\newcommand{\I}[1]{\texttt{I-#1}}
\newcommand{\T}[1]{\texttt{T-#1}}

\definecolor{lightgray}{gray}{0.95}
\definecolor{myblue}{RGB}{0, 0, 165}
\newcommand{\red}[1]{\textcolor{purple}{#1}}
\newcommand{\blue}[1]{\textcolor{myblue}{#1}}

\definecolor{purpleheart}{rgb}{0.41, 0.21, 0.61}

\DeclareCaptionFormat{mycaptionformat}{\fontsize{8}{8}\selectfont#1#2#3}

\title{Taxonomy of Benchmarks in Graph Representation Learning}
\articleno{106}

\author[Liu \& Cantürk et al.]{%
   Renming Liu\thanks{Equal contribution.} \\
   \institute{Michigan State University} \\
   \email{liurenmi@msu.edu}
   \And
   Semih Cantürk\footnotemark[1] \\
   \institute{Mila, Université de Montréal} \\
   \email{semih.canturk@mila.quebec}
   \And
   Frederik Wenkel \\
   \institute{Mila, Université de Montréal}
   \And
   Sarah McGuire \\
   \institute{Michigan State University}
   \And
   Xinyi Wang \\
   \institute{Michigan State University}
   \And
   Anna Little \\
   \institute{University of Utah}
   \And
   Leslie O'Bray \\
   \institute{ETH Zürich}
   \And
   Michael Perlmutter\thanks{Co-PI.} \\
   \institute{University of California, Los Angeles}
   \And
   Bastian Rieck\footnotemark[2] \\
   \institute{Helmholtz Munich \& TU Munich}
   \AND
   Matthew Hirn\footnotemark[2] \\
   \institute{Michigan State University}
   \And
   Guy Wolf \footnotemark[2] \\
   \institute{Mila, Université de Montréal} \\
   \email{wolfguy@mila.quebec}
   \And
   Ladislav Rampášek\footnotemark[1] \textsuperscript{ ,}\footnotemark[2] \\
   \institute{Mila, Université de Montréal} \\
   \email{ladislav.rampasek@mila.quebec}
}

\begin{document}

\maketitle

\setcounter{footnote}{0}
\begin{abstract}
Graph Neural Networks (GNNs) extend the success of neural networks to graph-structured data by accounting for their intrinsic geometry. While extensive research has been done on developing GNN models with superior performance according to a collection of graph representation learning benchmarks, it is currently not well understood what aspects of a given model are probed by them. For example, to what extent do they test the ability of a model to leverage graph structure vs.\ node features?
Here, we develop a principled approach to taxonomize benchmarking datasets according to a \emph{sensitivity profile} that is based on how much GNN performance changes due to a collection of graph perturbations.
Our data-driven analysis provides a deeper understanding of which benchmarking data characteristics are leveraged by GNNs. 
Consequently, our taxonomy can aid in selection and development of adequate graph benchmarks, and better informed evaluation of future GNN methods. Finally, our approach and implementation in \texttt{GTaxoGym} package\footnote{\url{https://github.com/G-Taxonomy-Workgroup/GTaxoGym}}
are extendable to multiple graph prediction task types and future datasets.

% \noindent\emph{Anonymized source code is provided in Supplementary Files.}
\end{abstract}

%###########################################################
\section{Introduction}
%###########################################################

Machine learning for graph representation learning (GRL) has seen rapid development in recent years~\cite{hamilton2020}.
Originally inspired by the success of convolutional neural networks in regular Euclidean domains, thanks to their ability to leverage data-intrinsic geometries, classical graph neural network (GNN) models~\cite{defferrard2016convolutional,kipf2016GCN, velivckovic2017graph} extend those principles to irregular graph domain.
Further advances in the field have led to a wide selection of complex and powerful GNN architectures. Some models are provably more expressive than others \cite{xu2018gin, morris2019kGNN}, can leverage multi-resolution views of graphs \cite{min_scattering_2020}, or can account for implicit symmetries in graph data \cite{bronstein_geometric_2021}. Comprehensive surveys of graph neural networks can be found in~\citet{Bronstein_2017,wu2020,ZHOU202057}.

Most graph-structured data encode information in \emph{graph structures} and \emph{node features}. The structure of each graph represents relationships (i.e., edges) between different nodes, while the node features represent quantities of interest at each node. For example, in citation networks, nodes represent papers, and edges represent citations between the papers. On such networks, node features often capture the presence or absence of certain keywords in each paper, encoded in binary feature vectors. In graphs modeling social networks, each node represents a user, and the corresponding node features often include user statistics like gender, age, or binary encodings of personal interests. 

Intuitively, the power of GNNs lies in relating local node-feature information to global graph structure information, typically achieved by applying a cascade of feature \emph{aggregation} and \emph {transformation} steps. In aggregation steps, information is exchanged between neighboring nodes, while transformation steps apply a (multi-layer) perceptron to feature vectors of each node individually. Such architectures are commonly referred to as \emph{Message Passing Neural Networks (MPNN)}~\cite{gilmer2017neural}.

Historically, GNN methods have been evaluated on a small collection of datasets \cite{morris2020tudataset}, many of which originated from the development of graph kernels. The limited quantity, size and variety of these datasets have rendered them insufficient to serve as distinguishing benchmarks \cite{dwivedi_benchmarking_2020,palowitch2022graphworld}. Therefore, recent work has focused on compiling a set of large(r) benchmarking datasets across diverse graph domains \cite{dwivedi_benchmarking_2020, hu_open_2021}. Despite these efforts and the introduction of new datasets, it is still not well understood what aspects of a dataset most influence the performance of GNNs. Which is more important, the geometric structure of the graph or the node features? Are long-range interactions crucial, or are short-range interactions sufficient for most tasks?
This lack of understanding of the dataset properties and of their similarities makes it difficult to select a benchmarking suit that would enable comprehensive evaluation of GNN models. Even when an array of seemingly different datasets is used, they may probe similar aspects of graph representation learning.

% This lack of understanding in dataset properties make it difficult to determine subsets of graph datasets that are able to statistically separate GNN model performance despite the steady increase in the number of available datasets. This phenomenon stems from redundancies in the properties tested for even when an array of datasets are used, as seemingly different datasets may be employing similar pathways for the propagation of information, leading to quickly diminishing marginal returns as the number of datasets used for benchmarking increases. In turn, the small subset of datasets used for benchmarking have stayed the same for the most part, and have resulted in impaired benchmarking practices in the field of graph learning \cite{palowitch2022graphworld}, and arguably even hindered further progress in the field.

\begin{figure}[t]
    \centering
    \includegraphics[width=0.9\textwidth]{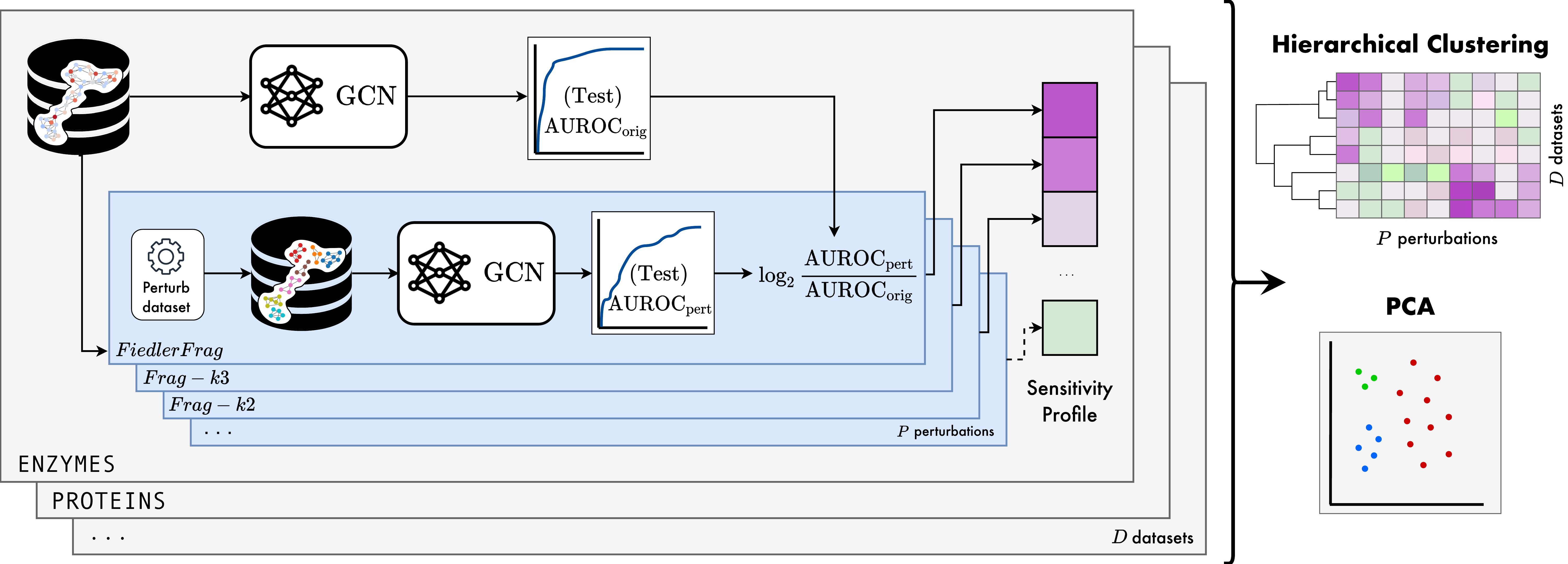}
    \vspace{-2pt}
    \caption{Overview of our pipeline to taxonomize graph learning datasets. %; described in Sec.~\ref{ss:taxonomization}.
    }
    \label{fig:viz_abs}
    \vspace{-10pt}
\end{figure}

Leveraging symmetries and other geometric priors in graph data is crucial for generalizable learning~\cite{bronstein_geometric_2021}. While invariance or equivariance to some transformations is inherent, invariance to others may only be empirically or partially apparent. Motivated by this observation, we propose to use the lens of empirical transformation sensitivity to gauge \emph{how} task-related information is encoded in graph datasets and subsequently taxonomize their use as benchmarks in graph representation learning. Our approach is illustrated in Figure~\ref{fig:viz_abs}.
% The goal of this paper is to propose and apply a framework through which we can methodologically define and evaluate the characteristics of benchmarking datasets in terms of their reliance on particular types of information encoding and propagation. 
Namely, we list our contributions in this study as:
%\vspace{-5pt}
\begin{enumerate}[leftmargin=2em, itemsep=0.2em]
    \item We develop a graph dataset taxonomization framework that is extendable to both new datasets and evaluation of additional graph/task properties.
    \item Using this framework, we provide the first taxonomization of GNN (and GRL) benchmarking datasets, collected from TUDatasets \cite{morris2020tudataset}, OGB \cite{hu_open_2021} and other sources.
    \item Through the resulting taxonomy, we provide insights about existing datasets and guide better dataset selection in future benchmarking of GNN models.
\end{enumerate}

% We believe that our methodology and the resulting taxonomy can alleviate the problems mentioned above by acting both as a tool to understand existing and future graph datasets and models better, and as a guide to aid in the selection of benchmarking datasets that sufficiently express the variation and complexity of real-world derived graph data.

% This is achieved via a series of experiments in which we \emph{perturb} certain types of graph properties while leaving others intact. Based upon these experiments, we propose a taxonomy that categorizes datasets according to what type of information has a large impact on GNN performance.
% These insights improve our understanding of empirical evaluations of GNNs and will lead to appropriate empirical validation of future models.
% In particular, we address the two questions above by demonstrating that (i) eliminating graph structure has limited effects on many datasets, and (ii) smooth local information is crucial for the majority of datasets, while long-range dependencies are only realized in a few cases.
% We found that a diverse categorization of graph datasets can be obtained by measuring their sensitivities to perturbations that alter graph structure and ones that eliminate or modify node features. 

%###########################################################
% \newpage
\vspace{-3pt}
\section{Methods}
\vspace{-2pt}
%###########################################################

\begin{figure}[t]
% 2 x 6 layout version
\vspace{-10pt}
\captionsetup[sub]{format=mycaptionformat,labelfont={bf}}
\centering
     \begin{subfigure}[b]{0.12\textwidth}
         \centering
         \includegraphics[width=\textwidth]{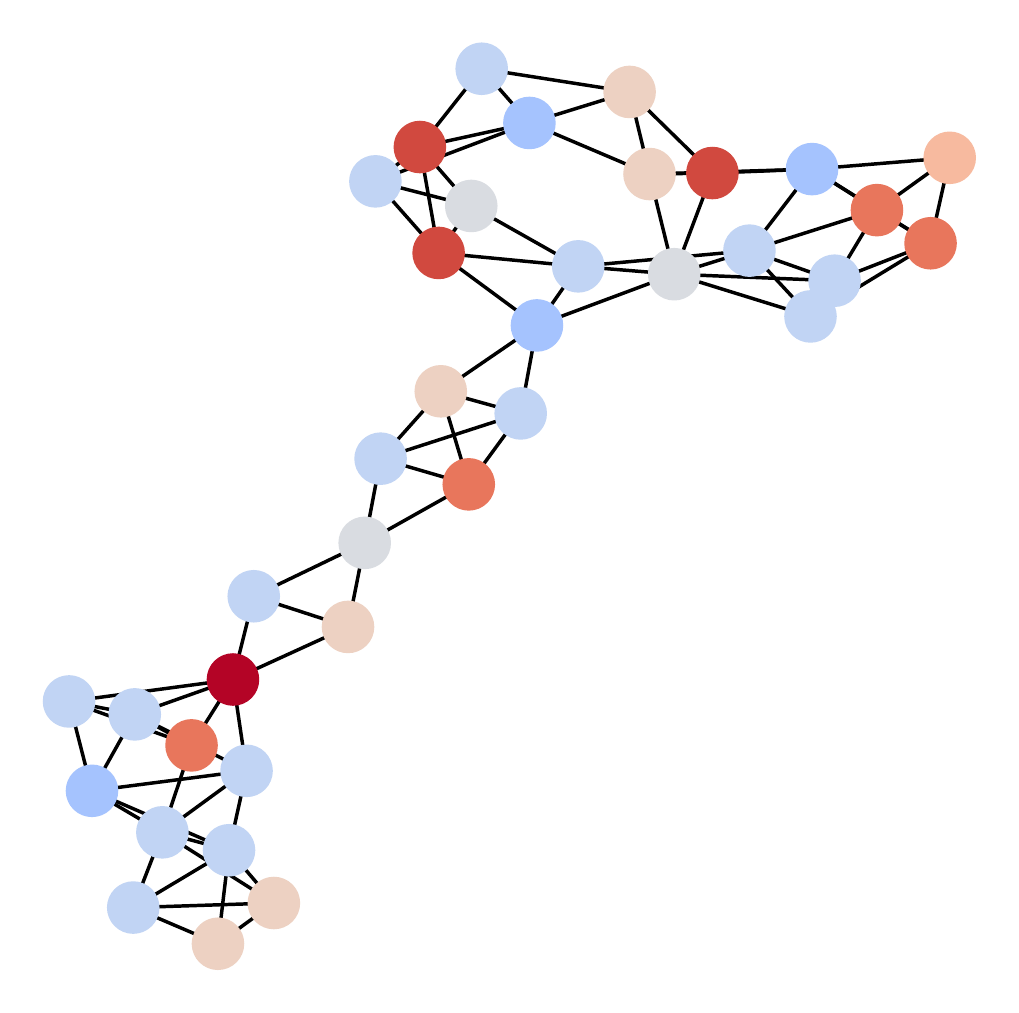}
         \caption{original}
         \label{fig:perts-og}
     \end{subfigure}
     \hfill
     \begin{subfigure}[b]{0.12\textwidth}
         \centering
         \includegraphics[width=\textwidth]{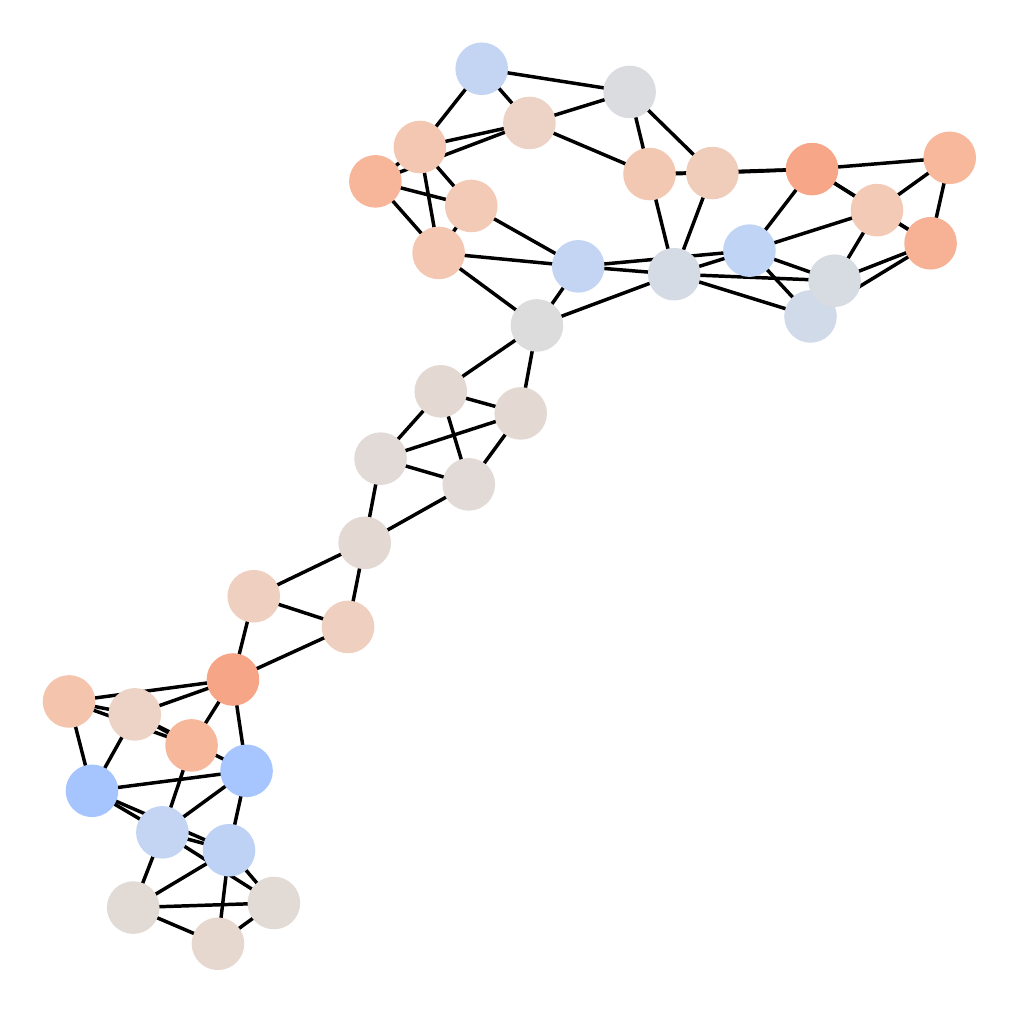}
         \caption{\red{LowPass}}
         \label{fig:perts-low}
     \end{subfigure}
     \hfill
     \begin{subfigure}[b]{0.12\textwidth}
         \centering
         \includegraphics[width=\textwidth]{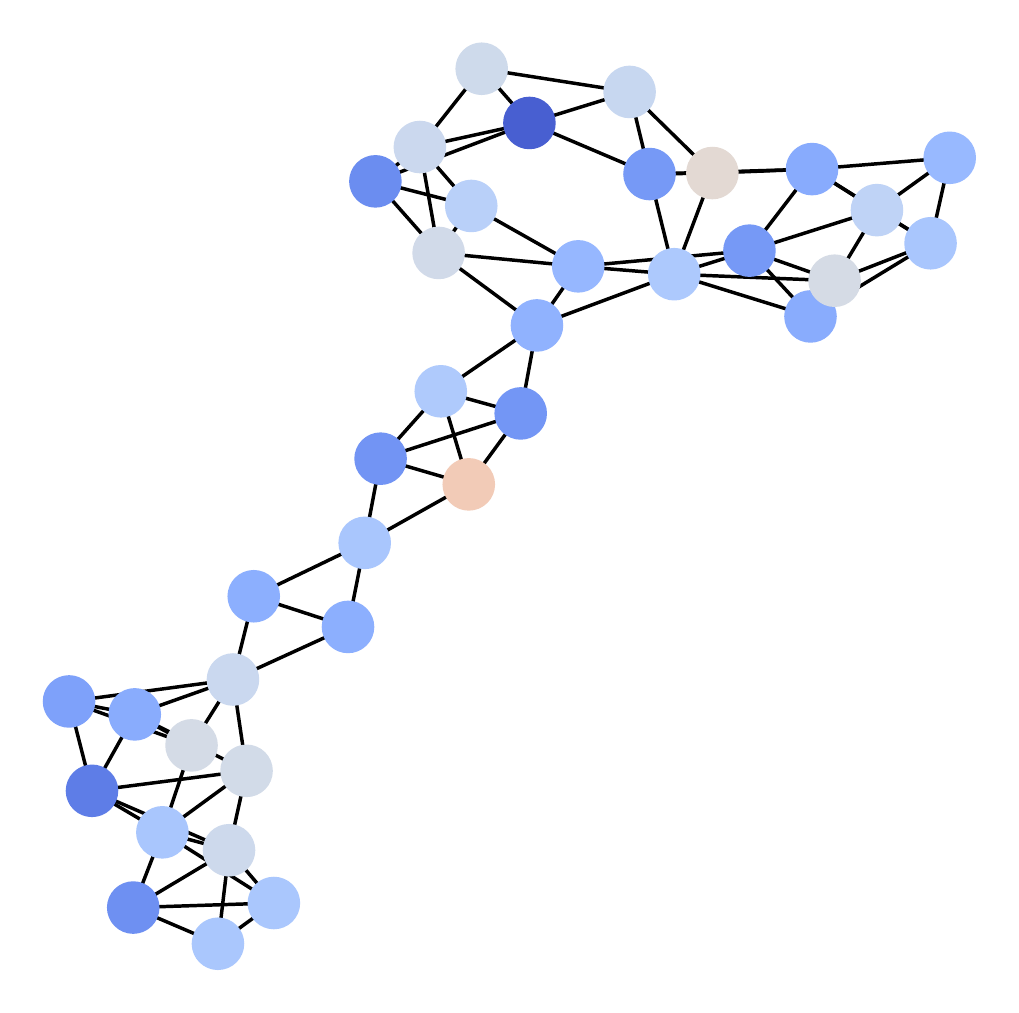}
         \caption{\red{MidPass}}
         \label{fig:perts-mid}
     \end{subfigure}
     \hfill
     \begin{subfigure}[b]{0.12\textwidth}
         \centering
         \includegraphics[width=\textwidth]{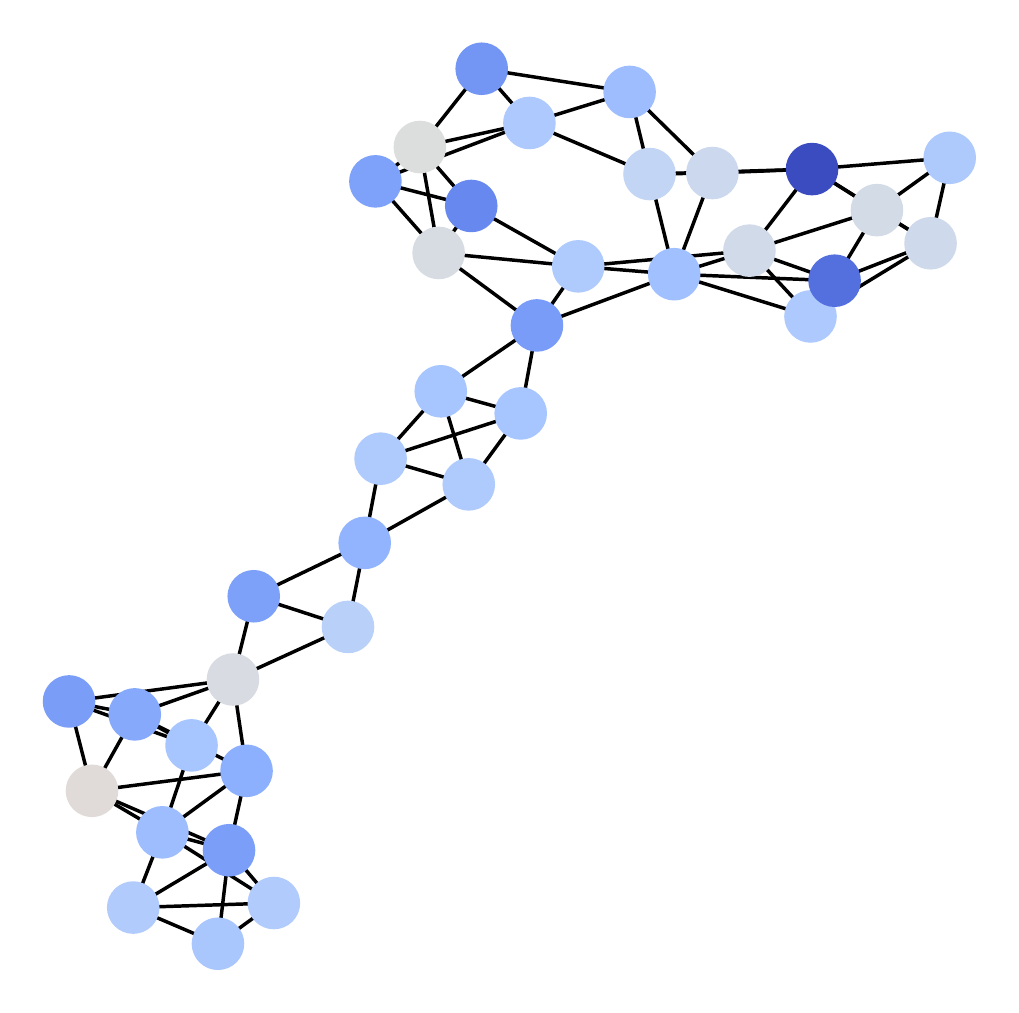}
         \caption{\red{HighPass}}
         \label{fig:perts-high}
     \end{subfigure}
     \hfill
     \begin{subfigure}[b]{0.12\textwidth}
         \centering
         \includegraphics[width=\textwidth]{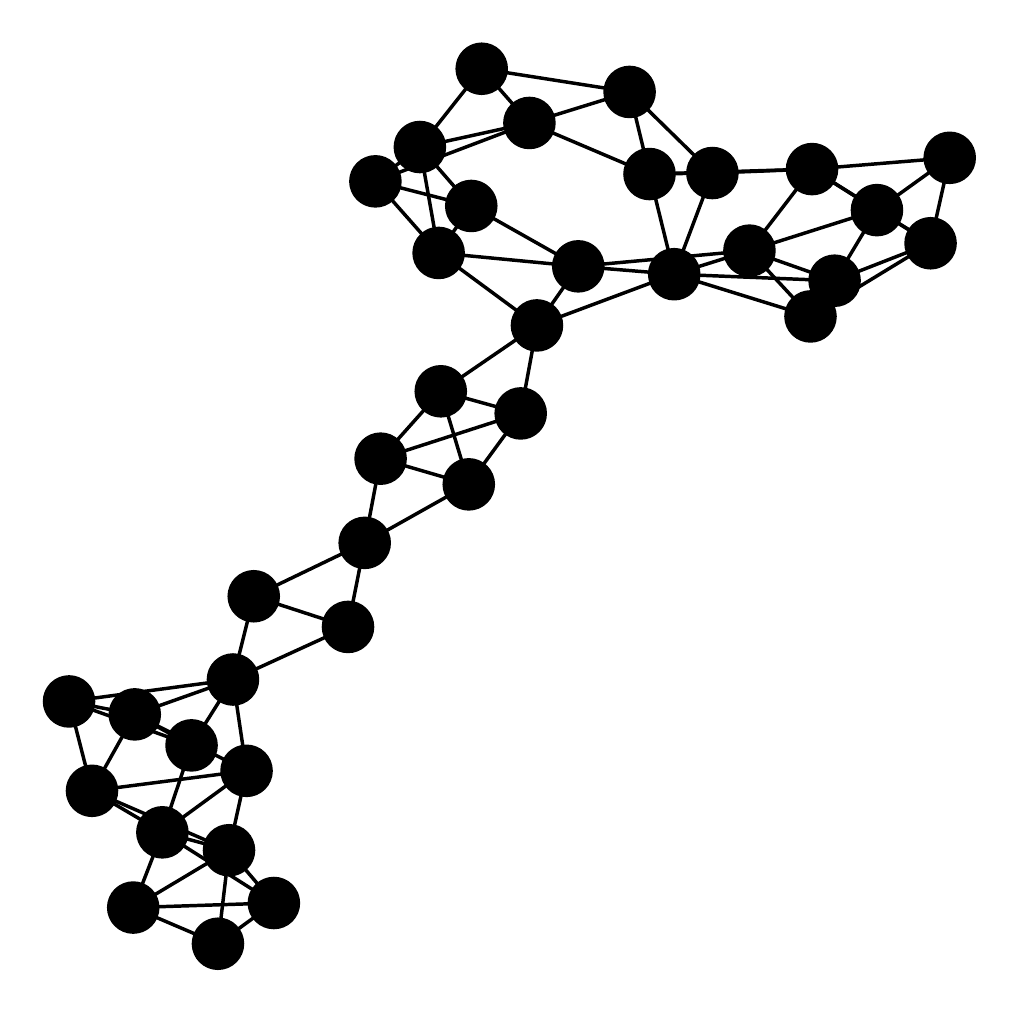}
         \caption{\fontsize{7.5}{8}\selectfont \red{NoNodeFtrs}}
         \label{fig:perts-nonodeftr}
     \end{subfigure}
     \hfill
     \begin{subfigure}[b]{0.12\textwidth}
         \centering
         \includegraphics[width=\textwidth]{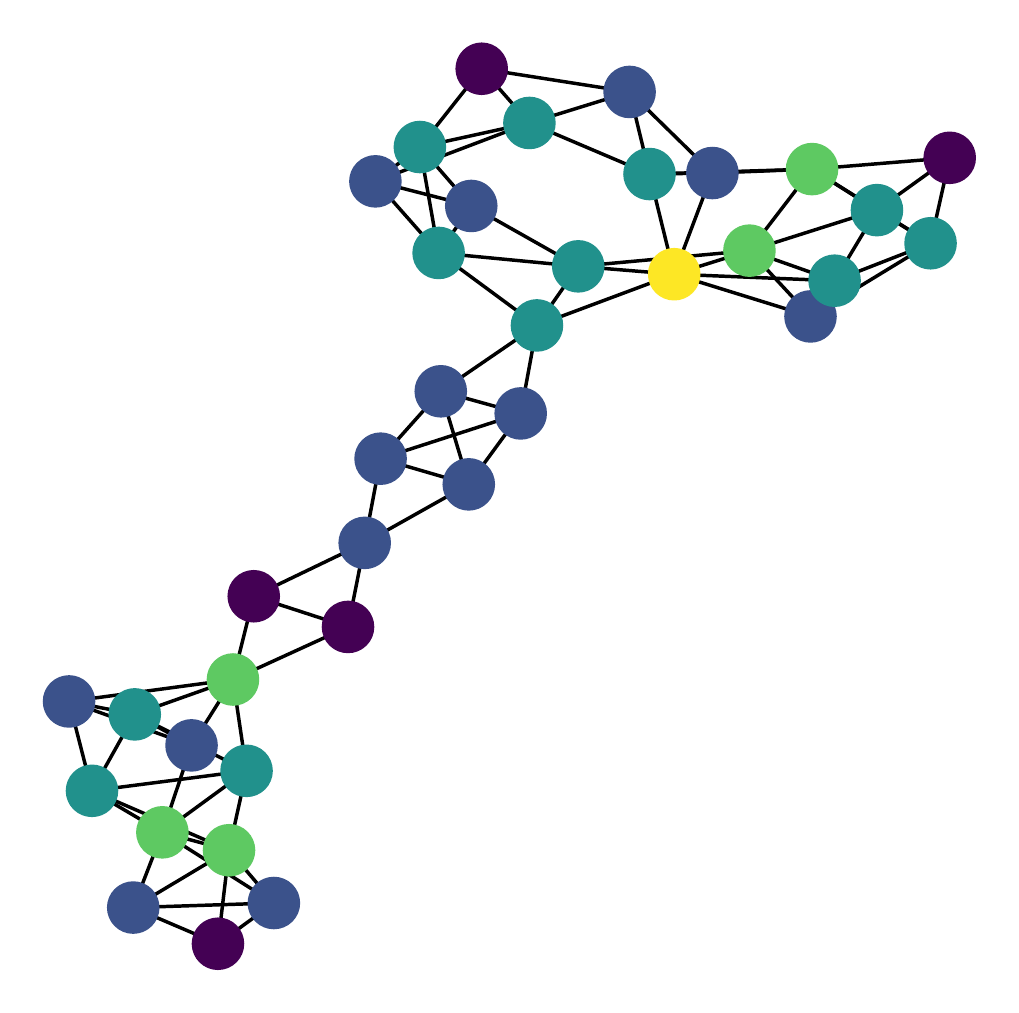}
         \caption{\red{NodeDeg}}
         \label{fig:perts-nodedeg}
     \end{subfigure}
     \hfill
     \begin{subfigure}[b]{0.12\textwidth}
         \centering
         \includegraphics[width=\textwidth]{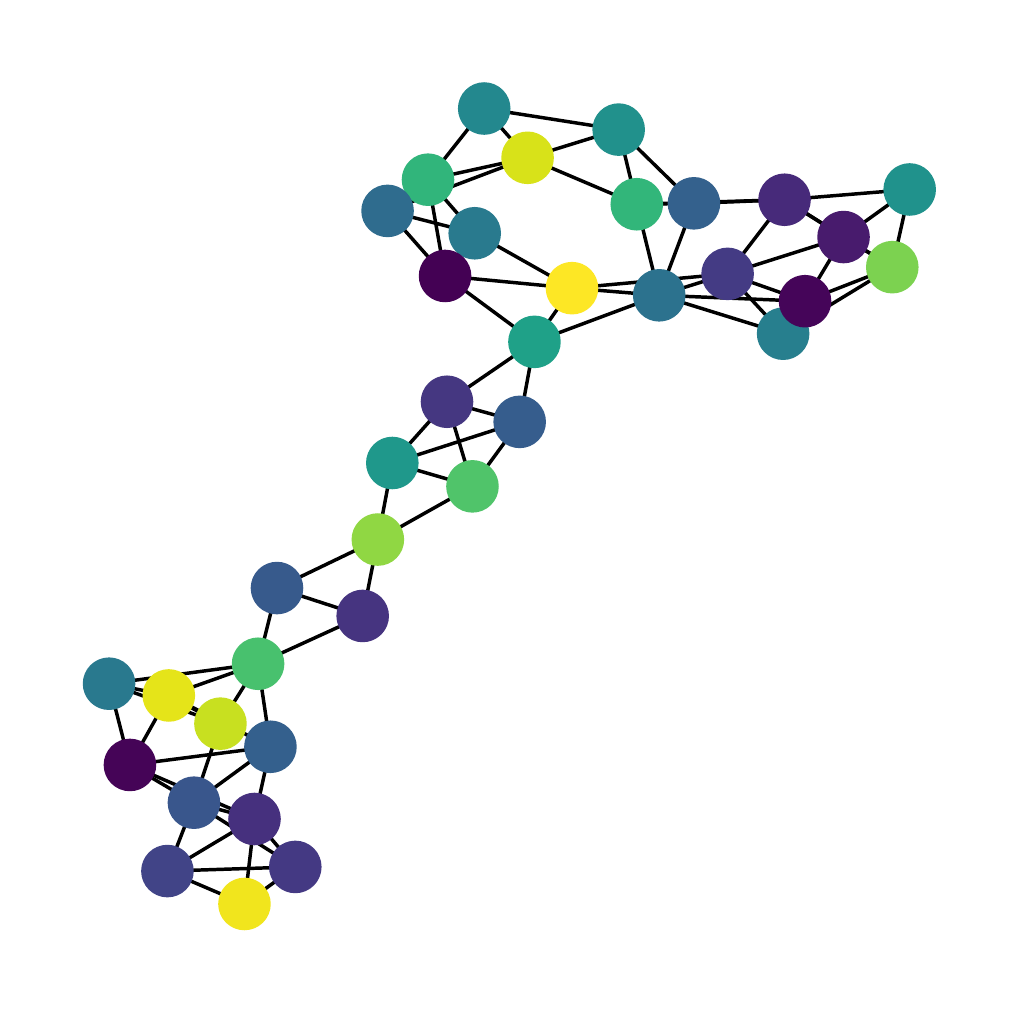}
         \caption{\red{RandFtrs}}
         \label{fig:perts-rf}
     \end{subfigure}
     \hfill
     \begin{subfigure}[b]{0.12\textwidth}
         \centering
         \includegraphics[width=\textwidth]{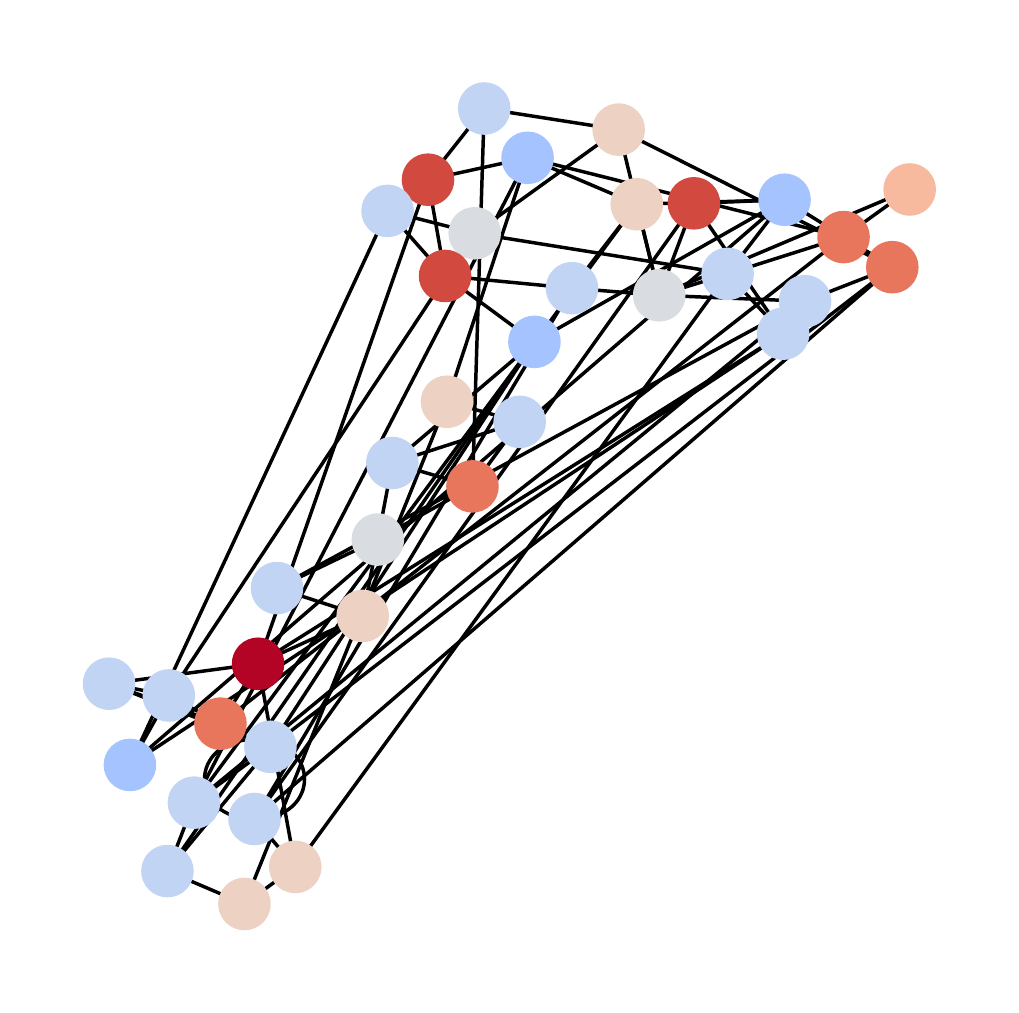}
         \caption{\fontsize{7}{7.5}\selectfont \blue{RandRewire}}
         \label{fig:perts-rr}
     \end{subfigure}
     \hfill
     \begin{subfigure}[b]{0.12\textwidth}
         \centering
         \includegraphics[width=\textwidth]{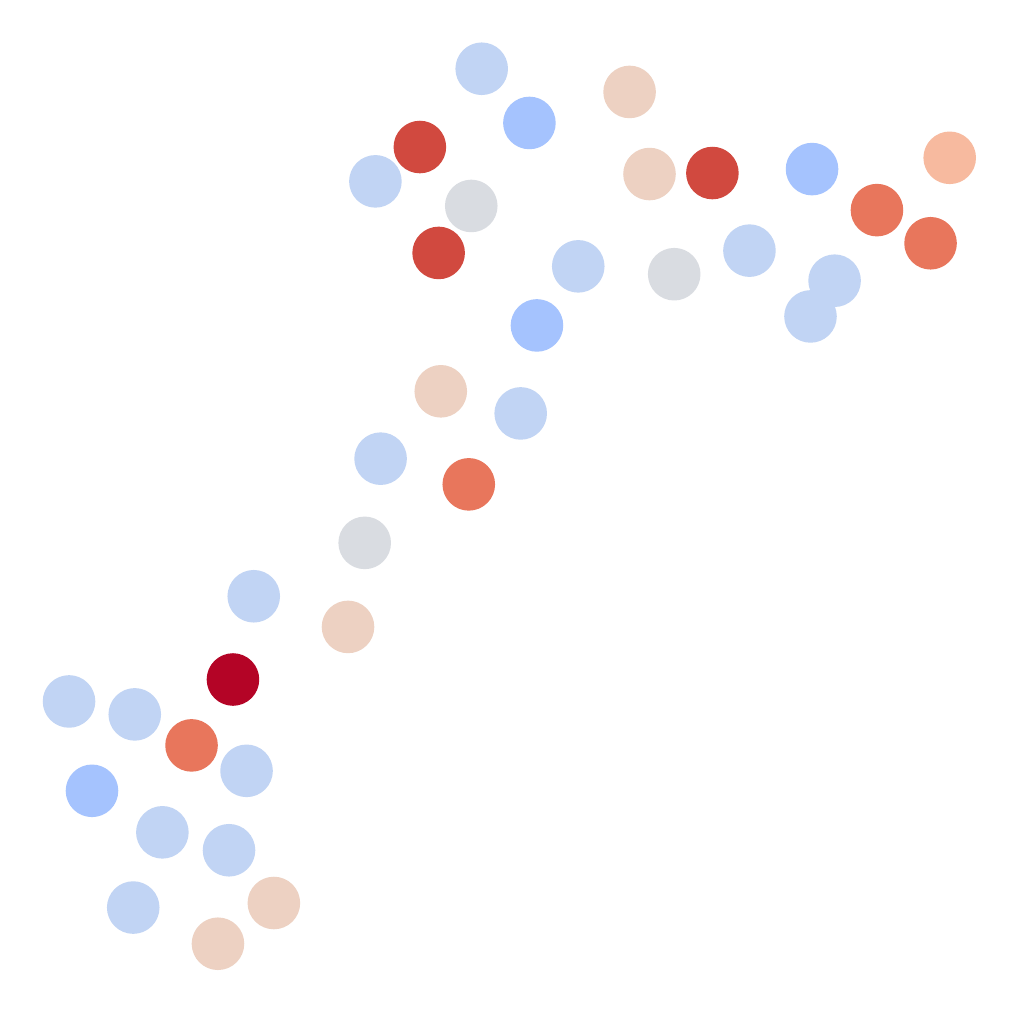}
         \caption{\blue{NoEdges}}
         \label{fig:perts-ne}
     \end{subfigure}
     \hfill
     \begin{subfigure}[b]{0.12\textwidth}
         \centering
         \includegraphics[width=\textwidth]{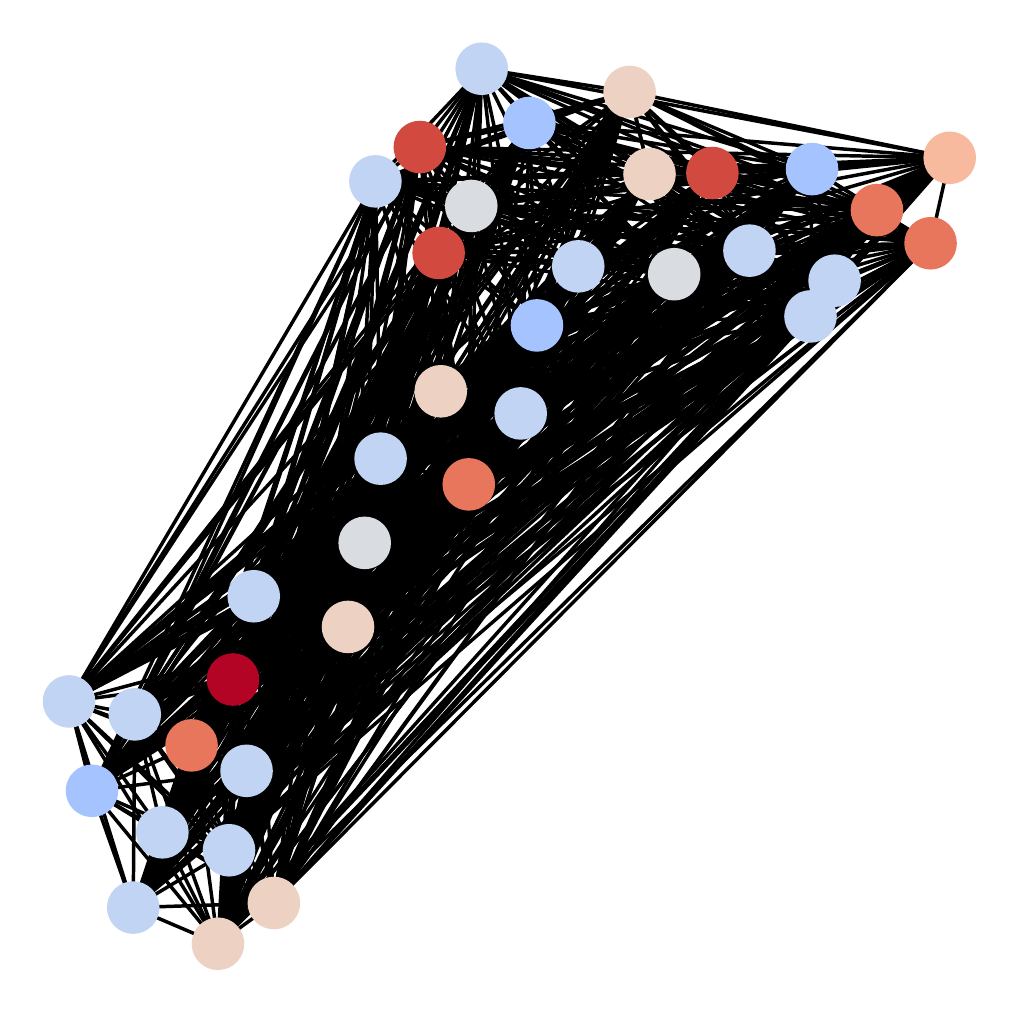}
         \caption{\blue{FullyConn}}
         \label{fig:perts-fc}
     \end{subfigure}
     \hfill
     \begin{subfigure}[b]{0.12\textwidth}
         \centering
         \includegraphics[width=\textwidth]{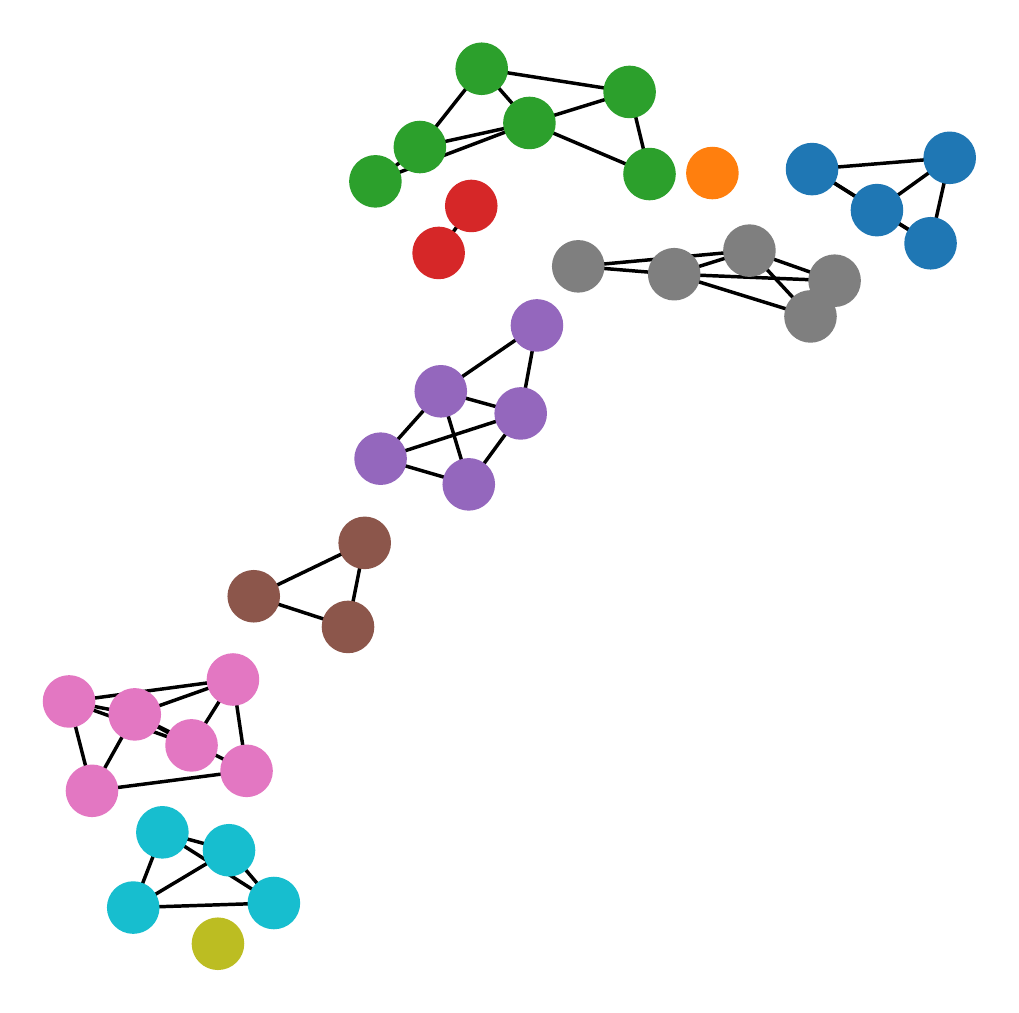}
         \caption{\blue{Frag. $k{=}1$}}
         \label{fig:perts-frag1}
     \end{subfigure}
     \hfill
     \begin{subfigure}[b]{0.12\textwidth}
         \centering
         \includegraphics[width=\textwidth]{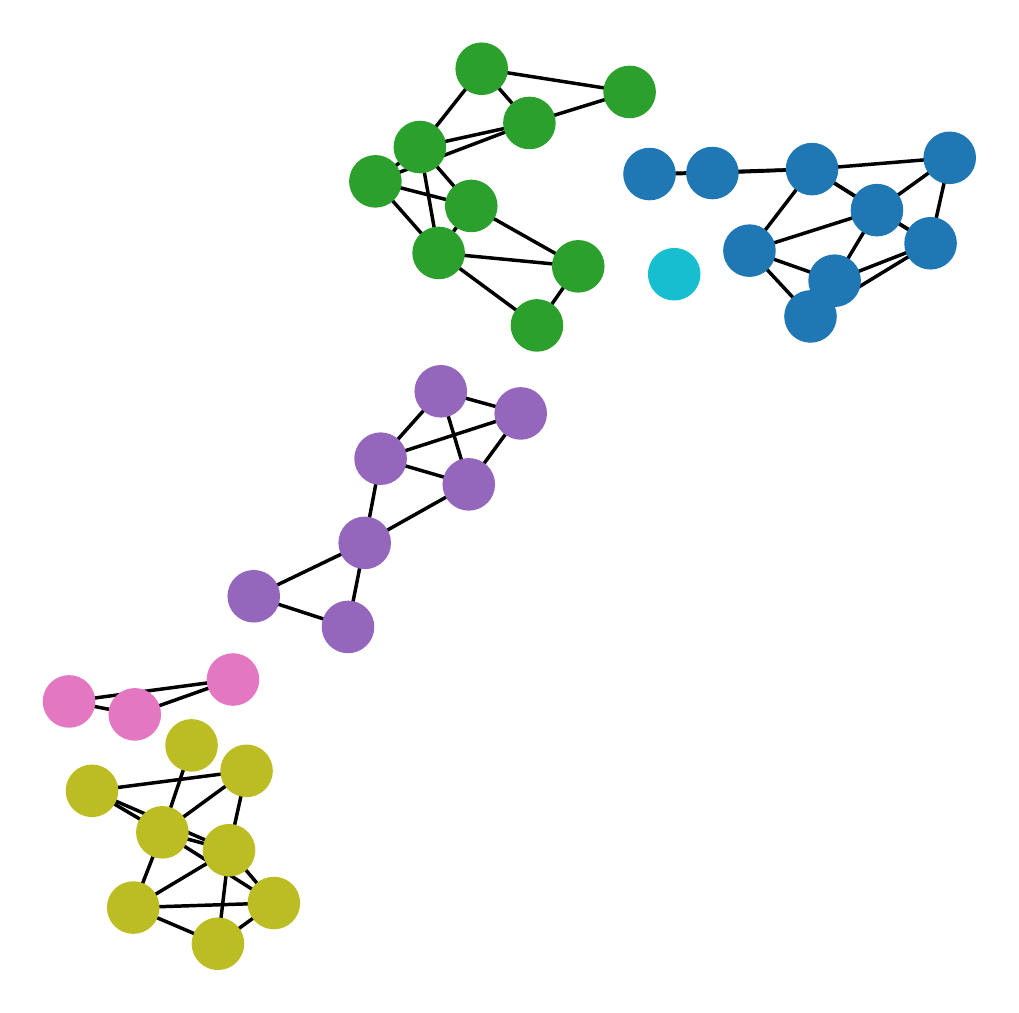}
         \caption{\blue{Frag. $k{=}2$}}
         \label{fig:perts-frag2}
     \end{subfigure}
     \hfill
     \begin{subfigure}[b]{0.12\textwidth}
         \centering
         \includegraphics[width=\textwidth]{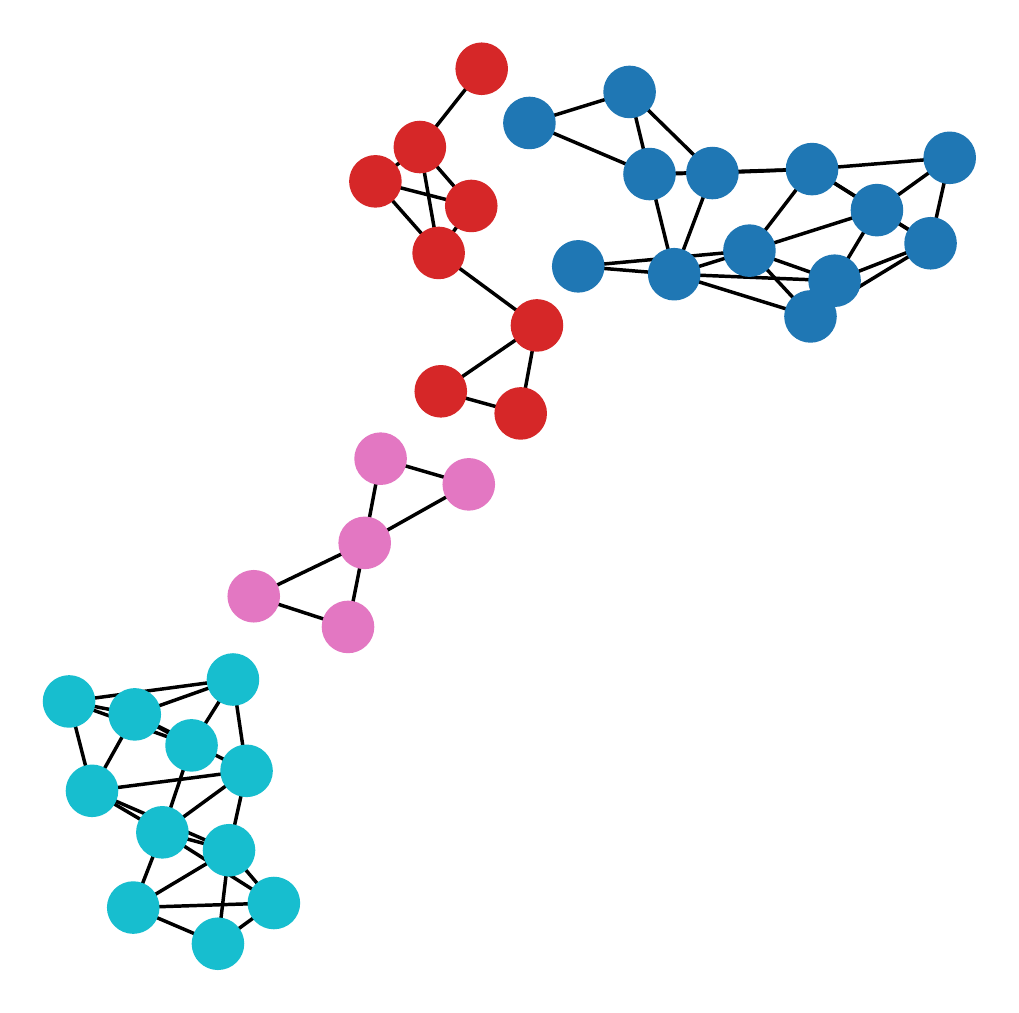}
         \caption{\blue{Frag. $k{=}3$}}
         \label{fig:perts-frag3}
     \end{subfigure}
     \hfill
     \begin{subfigure}[b]{0.12\textwidth}
         \centering
         \includegraphics[width=\textwidth]{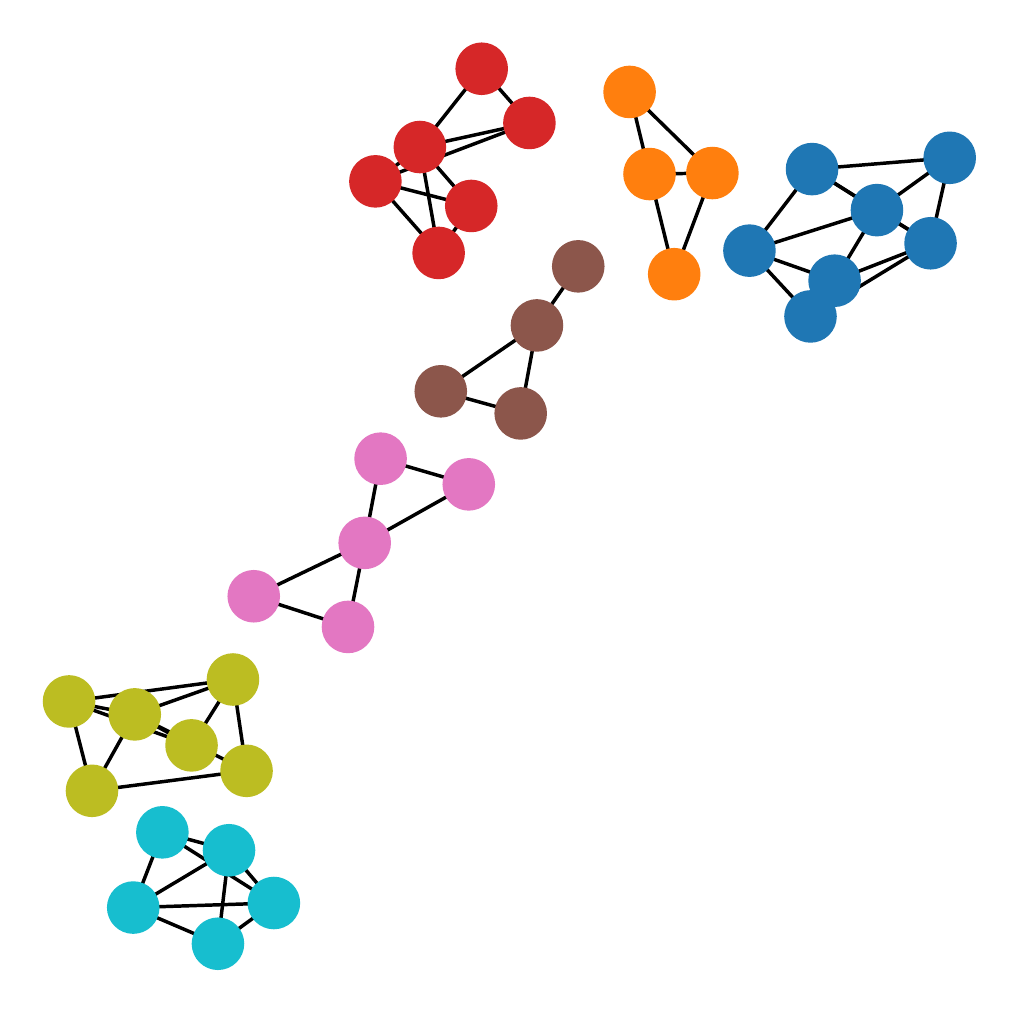}
         \caption{\blue{FiedlerFrag}}
         \label{fig:perts-fiedler}
     \end{subfigure}
\caption{\red{Node feature} and \blue{graph structure} perturbations of the first graph in ENZYMES. The color coding of nodes illustrates their feature values, except (k-n) where the fragment assignment is shown.}
\label{fig:perts}
\vspace{-3pt}
\end{figure}

As a proxy for invariance or sensitivity to graph perturbations, we study the changes in GNN performance on perturbed versions of each dataset.
These perturbations are designed to eliminate or emphasize particular types of information embedded in the graphs.
We define an empirical \emph{sensitivity profile} of a dataset as a vector where each element is the performance of a GNN after a given perturbation, reported as a percentage of the network's performance on the original dataset.
% For a given dataset and its prediction task, this sensitivity profile provides a comprehensive view of what information is important, which then enables one to thoughtfully design or select a GNN model for further use on that dataset. 
% For instance, if long-range interactions are important on a given graph, one might  use a network designed to capture such interactions for improved performance.
% We define an empirical \emph{sensitivity profile} of a dataset as a vector where each element is the percentage of performance of a GNN retains after a given perturbation is applied to the dataset.
% For example, if ignoring all node features causes the networks performance to become $3\%$ worse, the corresponding entry would be $97\%$.
% Then, we taxonomize the datasets based their sensitivity profiles via hierarchical clustering.
In particular, we use a set of 13 perturbations, visualized in Figure~\ref{fig:perts}. Of these perturbations, 6 are designed to perturb node features, while keeping the graph structure intact, whereas the remaining 7 keep the node attributes the same, but manipulate the graph structure.

For the purpose of these perturbations, we consider all graphs to be undirected and unweighted, and assume they all have node features, but not edge features. These assumptions hold for most datasets we use in this study. However, if necessary, we preprocess the data by symmetrizing each graph's adjacency matrix and dropping any edge attributes.
Formally, let $G = (V, E, \mathbf{X})$ be an undirected, unweighted, attributed graph with node set $V$ of cardinality $\vert V\vert=n$, edge set $E\subset V\times V$, and a matrix of $d$-dimensional node features $\mathbf{X} \in \mathbb{R}^{n\times d}$. 
We let $\mathbf{M} \in \mathbb{R}^{n\times n}$ denote the adjacency matrix of each graph, where $\mathbf{M}(u,v) = 1$ if $(u,v) \in E$ and zero otherwise.

Several of our perturbations are based on spectral graph theory, which represents graph signals in a spectral domain analogous to classical Fourier analysis. 
We define the graph Laplacian $\mathbf{L} \coloneqq \mathbf{D} - \mathbf{M}$ and the symmetric normalized graph Laplacian $\mathbf{N} \coloneqq \mathbf{D}^{-\frac12} \mathbf{L} \mathbf{D}^{-\frac12} = \mathbf{I} - \mathbf{D}^{-\frac12} \mathbf{M} \mathbf{D}^{-\frac12}$,
where $\mathbf{D}$ is the diagonal degree matrix.
Both  $\mathbf{L}$ and $\mathbf{N}$ are positive  semi-definite and have an orthonormal eigendecompositions $\mathbf{L} = \mathbf{\Phi} \mathbf{\Lambda} \mathbf{\Phi}^\top$ and $\mathbf{N} = \mathbf{\tilde \Phi} \mathbf{\tilde \Lambda} \mathbf{\tilde \Phi}^\top$.
By convention, we order the eigenvalues and corresponding eigenvectors $\{(\lambda_i, \mathbf{\phi}_i)\}_{0\leq i\leq n-1}$ of $\mathbf{L}$ (and similarly for $\mathbf{N}$) in ascending order $0=\lambda_0 \leq \lambda_1 \leq \dots \leq \lambda_{n-1}$.
The eigenvectors $\{\phi_i\}_{0\leq i\leq n-1}$ constitute a basis of the space of graph signals and can be considered as generalized Fourier modes. The eigenvalues $\{\lambda_i\}_{0\leq i\leq n-1}$ characterize the variation of these Fourier modes over the graph and can be interpreted as (squared) frequencies. 

%The perturbations we design can be categorized into perturbations that alter node features, and those that target graph structure; we will now elaborate on them in further detail.

\vspace{-3pt}
\subsection{Node Feature Perturbations}\label{sec:nf-pert}
We first consider two perturbations that alter local node features, setting them either to a fixed constant (w.l.o.g., one) for all nodes, or to a one-hot encoding of the degree of the node. We refer to these perturbations as \pert{NoNodeFtrs} (as constant node features carry no additional information) and \pert{NodeDeg}, respectively. Sensitivity to these perturbations, exhibited by a large decrease in predictive performance, may indicate that a task is dominated by highly informative node features.
Further, we consider a random node feature perturbation (\pert{RandFtrs}) by sampling a one-dimensional feature for each node from $\mathcal{U}_{[-1,1]}$, which has been shown to improve the WL expressiveness of MPNNs~\cite{abboud2021surprising, sato2021random}.

We also develop spectral node feature perturbations. As in Euclidean settings, the Fourier decomposition can be used to decompose graph signals into a set of canonical signals, called Fourier modes, which are organized according to increasing variation (or frequency). In Euclidean Fourier analysis, these modes are sinusoidal waves oscillating at different frequencies. %For example, audio signals are superpositions of such sinusoidal waves. 
A standard practice in audio signal processing is to remove noise from a signal by identifying and removing certain Fourier modes or \emph{frequency bands}.
We generalize this technique to graph datasets and systematically remove certain graph Fourier modes to probe the importance of the corresponding frequency bands.

In this perturbation, we use the frequencies derived from the symmetric normalized graph Laplacian $\mathbf{N}$ and split them into three roughly equal-sized frequency bands (\textit{low}, \textit{mid}, \textit{high}), i.e., bins of subsequent eigenvalues. To assess the importance of each of the frequency bands, we then apply \emph{hard} band-pass filtering to the graph signals (node feature vectors), i.e., we project the signals on the span of the selected Fourier modes.
More specifically, for each band, we let $\mathbf{I}_{\text{band}}$ be a diagonal matrix with diagonal elements equal to one if the corresponding eigenvalue is in the band, and zero otherwise.
%We recall that $\mathbf{\tilde \Lambda}$ and $\mathbf{\tilde \Phi}$ are sorted so that the entries increase and hence represent increasing frequencies.
Then, the hard band-pass filtered signal is computed as
\vspace{-2pt}
\begin{align}
\mathbf{X}_{\text{band}} = \mathbf{\tilde \Phi} \mathbf{I}_{\text{band}} \mathbf{\tilde\Phi}^\top \mathbf{X}.
\end{align}
\vspace{-2pt}
The above band-pass filtering perturbation enables a precise selection of the frequency bands. However, it requires a full eigendecomposition of the normalized graph Laplacian, which is impractical for large graphs. We therefore provide an alternative approach based on wavelet bank filtering \cite{coifman2006diffusion}. This leverages the fact that polynomial filters $h$ of the normalized graph Laplacian directly transform the spectrum via $h(\mathbf{N})=\mathbf{\tilde \Phi} h(\mathbf{\tilde \Lambda}) \mathbf{\tilde \Phi}^\top$, yielding the \emph{frequency response} $h(\lambda)$ for any eigenvalue $\lambda$ of $\mathbf{N}$. This is usually done by taking the symmetrized diffusion matrix
\vspace{-2pt}
\begin{align}
    \mathbf{T}
    &= \frac12(\mathbf{I} + \mathbf{D}^{-\frac12} \mathbf{M} \mathbf{D}^{-\frac12}) = \frac{1}{2} \left(2\mathbf{I} - \mathbf{\mathbf{N}}\right).
\end{align}
\vspace{-2pt}
%which is closely related to $\mathbf{N}$.
By construction, $\mathbf{T}$ admits the same eigenbasis as $\mathbf{N}$ but its eigenvalues are mapped from $[0,2]$ to $[0,1]$ via the frequency response $h(\lambda)=1-\lambda/2$. As a result, large eigenvalues are mapped to small values (and vice versa).
Next, we construct \emph{diffusion wavelets}~\cite{defferrard2016convolutional} that consist of differences of dyadic powers $2^k, k\in \mathbb{N}_0$ of $\mathbf{T}$, i.e., $\Psi_k=\mathbf{T}^{2^{k-1}}-\mathbf{T}^{2^{k}}$, which act as bandpass filters on the signal. Intuitively, this operator ``compares'' two neighborhoods of different sizes (radius $2^{k-1}$ and $2^k$) at each node.
Diffusion wavelets are usually maintained in a wavelet bank $\mathcal{W}_K=\{\mathbf{\Psi}_k,\mathbf{\Phi_K}\}_{k=0}^K$, which contains additional highpass $\mathbf{\Psi}_0=\mathbf{I}-\mathbf{T}$ and lowpass $\mathbf{\Psi_K}=\mathbf{T}^K$ filters.
In our experiments, we choose $K=1$, resulting in the following low, mid, and highpass filtered node features:
%\vspace{-2pt}
\begin{align}
    \mathbf{X}_{\text{high}} = (\mathbf{I} - \mathbf{T}) \mathbf{X}, \quad
    \mathbf{X}_{\text{mid}} = (\mathbf{T} - \mathbf{T}^2) \mathbf{X}, \quad
    \mathbf{X}_{\text{low}} = \mathbf{T}^2 \mathbf{X}.
    \label{eqn:wavelettransform}
\end{align}
\vspace{-2pt}
These filters correspond to frequency responses $h_{\text{high}}(\lambda)=\lambda/2$, $h_{\text{mid}}(\lambda)=(1-\lambda/2)-(1-\lambda/2)^2$ and $h_{\text{low}}(\lambda)=(1-\lambda/2)^2$. Therefore, the low-pass filtering preserves low-frequency information while suppressing high-frequency information, whereas high-pass filtering does the opposite. The mid-pass filtering suppresses all frequencies. However, it preserves much more middle-frequency information than it does high- or low-frequency information.

Therefore, this filtering may be interpreted as an approximation of the hard band-pass filtering discussed above. From the spatial message passing perspective, low-pass filtering is equivalent to local averaging of the node features, which has a profound implication on homophilic and heterophilic characteristics of the datasets (Sec.~\ref{sec:node-level-res}). Finally, since the computations needed in \eqref{eqn:wavelettransform} can be carried out via sparse matrix multiplications, they scale much better to large graphs. Therefore, we utilize the wavelet bank filtering for the datasets with larger graphs considered in Sec.~\ref{sec:node-level-res}, while for the smaller graphs, considered in Sec.~\ref{sec:graph-level-res}, we employ the direct band-pass filtering approach.

\subsection{Graph Structure Perturbations}\label{sec:gs-pert}
\vspace{-3pt}
The following perturbations act on the graph structure by altering the adjacency matrix. By removing all edges (\pert{NoEdges)} or making the graph fully-connected (\pert{FullyConn}), we can eliminate the structural information completely and essentially turn the graph into a set. The difference between the two perturbations lies in whether all nodes are processed independently or together. 
However, \pert{FullyConn} is only applied  to inductive datasets in Sec.~\ref{sec:graph-level-res} due to computational limitations.
Furthermore, we consider a degree-preserving random edge rewiring perturbation (\pert{RandRewire}). In each step, we randomly sample a pair of edges and randomly exchange their end nodes. We then repeat this process without replacement until 50\% of the edges have been randomly rewired.

To inspect the importance of local vs.\ global graph structure, we designed the \pert{Frag-k} perturbations, which randomly partition the graph into connected components consisting of nodes whose distance to a seed node is less than $k$. Specifically, we randomly draw one seed node at a time and extract its $k$-hop neighborhood by eliminating all edges between this new fragment and the rest of the graph; we repeat this process on the remaining graph until the whole graph is processed. A smaller $k$ implies smaller components, and hence discards the global structure and long-range interactions.

Graph fragmentations can also be constructed using spectral graph theory. In our taxonomization, we adopt one such method, which we refer to as Fiedler fragmentation (\pert{FiedlerFrag}) (see \cite{irion2016efficient} and the references therein). In the case when the graph $G$ is connected, $\phi_0$, the eigenvector of the graph Laplacian $\mathbf{L}$ corresponding to $\lambda_0=0$, is constant. The eigenvector $\phi_1$ corresponding to the next smallest eigenvalue, $\lambda_1$, is known as the \emph{Fiedler vector}
\cite{fiedler1975property}. Since $\phi_0$ is constant, it follows that $\phi_1$ has zero average. This motivates partitioning the graph into two sets of vertices, one where $\phi_1$ is positive and the other where $\phi_1$ is negative.
We refer to this process as binary Fiedler fragmentation. 
This heuristic is used to construct the ratio cut for a connected graph~\cite{hagen1992new}. The ratio cut partitions a connected graph into two disjoint connected components $V=U \cupdot W$, 
%\anna{(maybe just use $\cup$, or define $\cupdot$)} \textcolor{orange}{Added the word \emph{disjoint}. That might be enough to clarify?}
such that the objective $\vert E(U, W)\vert/(\vert U\vert \cdot \vert W\vert)$ is minimized, where $E(U, W)\coloneqq \{(u,w)\in E : u\in U, w\in W\}$ is the set of removed edges when fragmenting $G$ accordingly. This can be seen as a combination of the min-cut objective (numerator), while encouraging a balanced partition (denominator).

\pert{FiedlerFrag} is based on iteratively applying binary Fiedler fragmentation. In each step, we separate the graph into its connected components and apply binary Fiedler fragmentation to the largest component. We repeat this process until either we reach 200 iterations, or the size of the largest connected component falls below 20. In contrast to the random fragmentation \pert{Frag-k}, this perturbation preserves densely connected regions of the graph and eliminates connections between them. Thus, \pert{FiedlerFrag} tests the importance of \textit{inter community message flow}.
Due to computational limits, we only apply \pert{FiedlerFrag} to inductive datasets in Sec.~\ref{sec:graph-level-res} for which this computation is feasible.

\subsection{Data-driven Taxonomization by Hierarchical Clustering} \label{ss:taxonomization}
\vspace{-3pt}
To study a systematic classification of the graph datasets, we use Ward's method \cite{ward1963hierarchical} for hierarchical clustering analysis on their \emph{sensitivity profiles}. The \emph{sensitivity profiles} are established empirically by contrasting the performance of a GNN model on a perturbed dataset and on the original dataset. To quantify this performance change, we use $\log_2$-transformed ratio of test AUROC (area under the ROC curve). Thus a sensitivity profile is a 1-D vector with as many elements as we have in perturbation experiments. See Figure~\ref{fig:viz_abs} and Appendix~\ref{sec:method-details} for further details.

In order to generate \emph{sensitivity profiles}, we must select suitable GNN models based on several practical considerations:
(i) The model has to be expressive enough to efficiently leverage aspects of the node features and graph structure that we perturb. Otherwise, our analysis will not be able to uncover reliance on these properties. (ii) The model needs to be general enough to be applicable to a wide variety of datasets, avoiding dataset-specific adjustments that may lead to profiling that is not comparable between datasets.
Therefore, we did not aim for specialized models that maximize performance, but rather models that (i) achieve at least baseline performance comparable to published works over all datasets, (ii) have manageable computational complexity to facilitate large-scale experimentation, and (iii) use well-established and theoretically well-understood architectures.

With these criteria in mind, we focused on two popular MPNN models in our analysis: GCN~\cite{kipf2016GCN} and GIN~\cite{xu2018gin}. The original GCN serves as an ideal starting point as its abilities and limitations are well understood. However, we also wanted to perform taxonomization through a provably more expressive and recent method, which motivated our selection of GIN as the second architecture. We emphasize that the main focus here is not to provide a benchmarking of GNN models \textit{per se}, but rather to address the taxonomization of \emph{graph datasets} (and accompanying tasks) used in such benchmarks. Nevertheless, we have also generated sensitivity profiles by additional models in order to comparatively demonstrate the robustness of our approach: 2-Layer GIN, ChebNet~\cite{defferrard2016convolutional}, GatedGCN~\cite{bresson2017residual} and GCN~II~\cite{pmlr-v119-chen20v}; see Figure~\ref{fig:gnn_corr}.

%###########################################################
\section{Results}
%###########################################################

\begin{figure}[t]
% \vspace*{-10pt}
\centering
     \begin{subfigure}[b]{0.48\textwidth}
         \centering
         \includegraphics[width=1\textwidth]{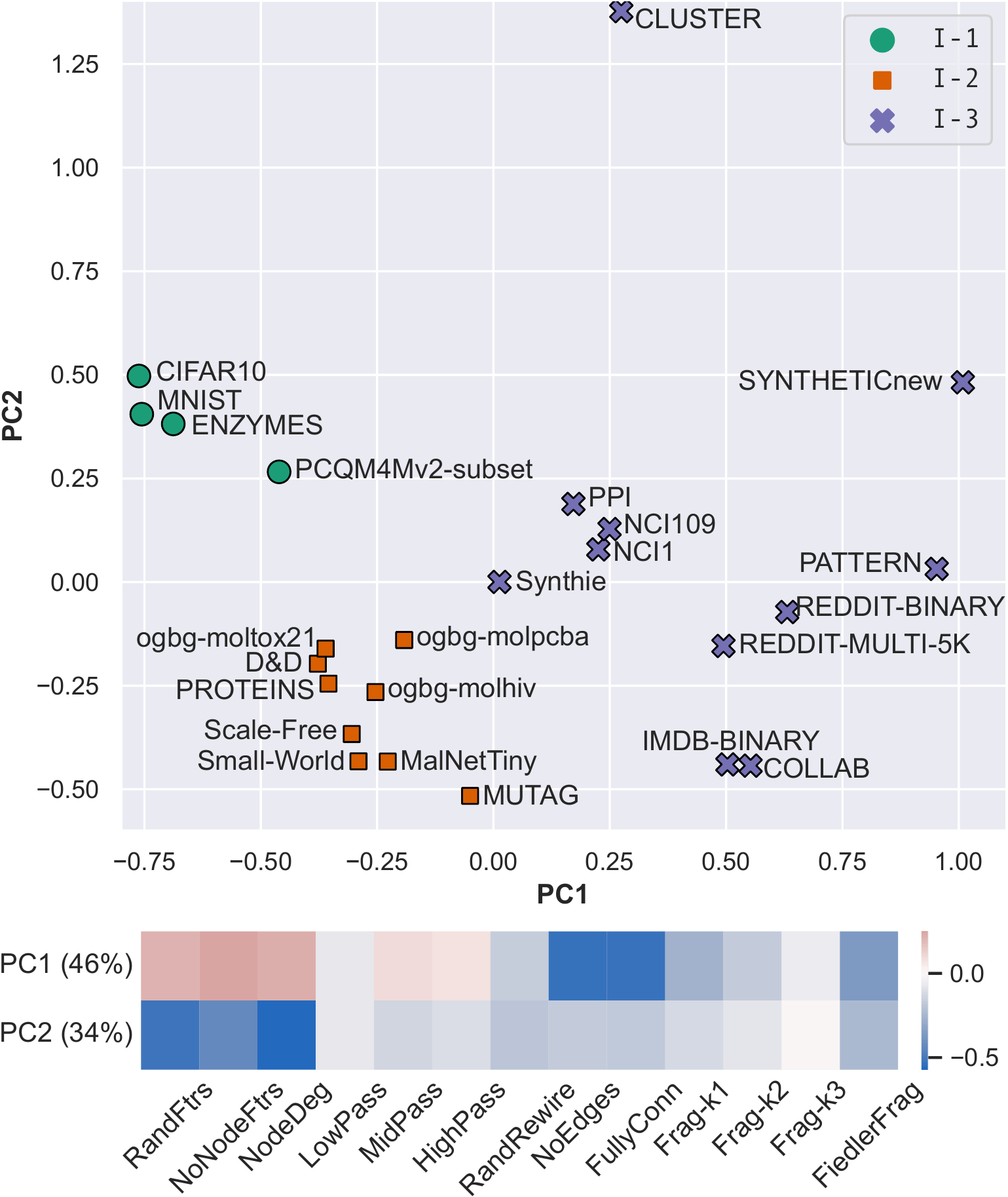}
         \vspace{-15pt}
         \caption{Inductive graph datasets}
         \label{fig:inductive-pca}
     \end{subfigure}
     \hfill
     \begin{subfigure}[b]{0.48\textwidth}
         \centering
         \includegraphics[width=1\textwidth]{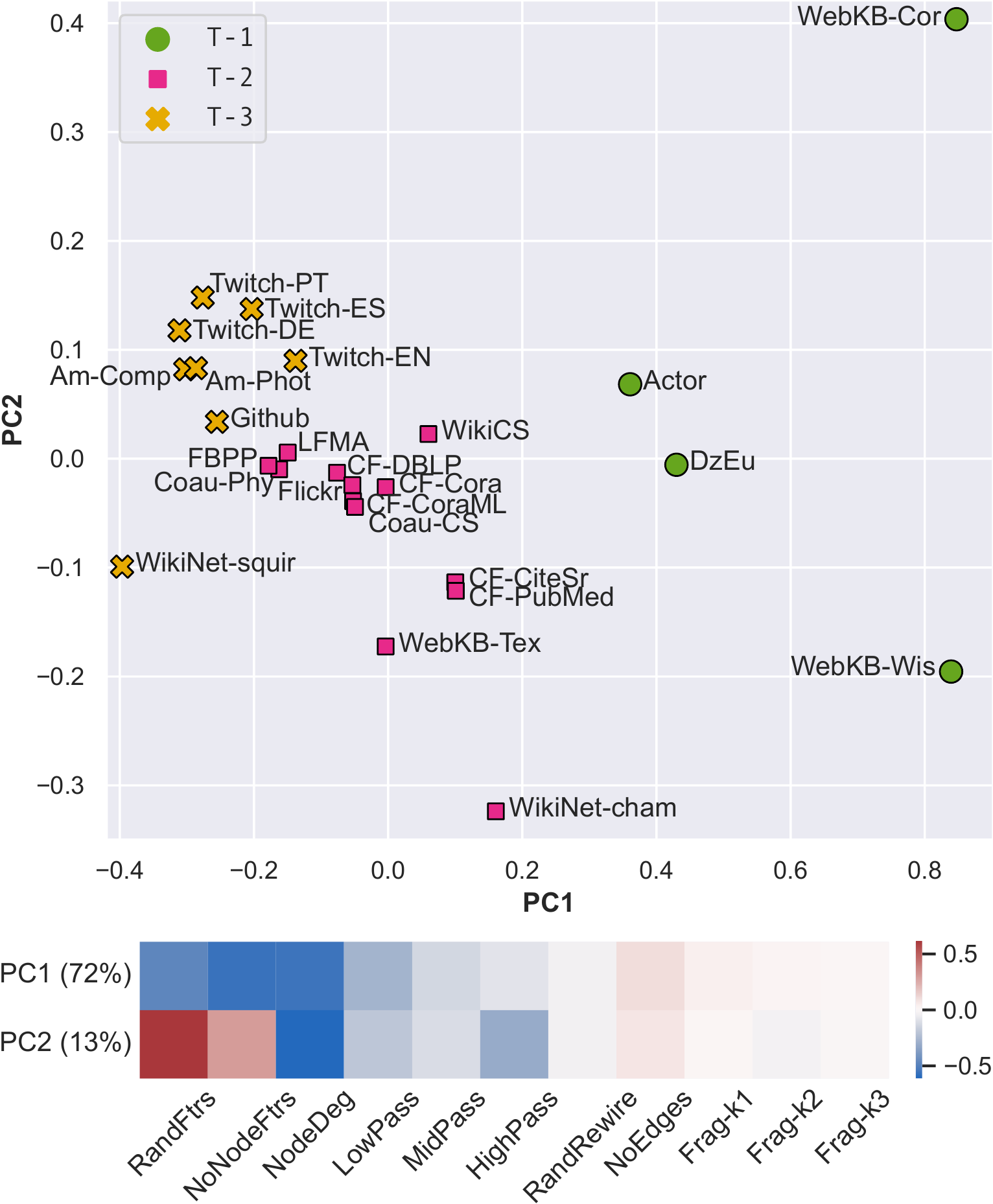}
         \vspace{-15pt}
         \caption{Transductive graph datasets}
         \label{fig:transductive-pca}
     \end{subfigure}
\caption{Visualization of (a) inductive and (b) transductive datasets based on PCA of their perturbation \emph{sensitivity profiles} according to a GCN model. The datasets are labeled according to their taxonomization by hierarchical clustering, shown in Figure~\ref{fig:gl-taxo}~and~\ref{fig:nl-taxo}, which corroborates with the emerging clustering in the PCA plots. In the bottom part are shown the loadings of the first two principal components and (in parenthesis) the percentage of variance explained by each of them.
}
\label{fig:pca}
% \vspace*{-10pt}
\end{figure}

Each of the 49 datasets we consider is equipped with either a node classification or graph classification task.
In the case of node classification, we further differentiate between the \emph{inductive} setting, in which learning is done on a set of graphs and the generalization occurs from a training set of graphs to a test set, and the \emph{transductive} setting, in which learning is done in one (large) graph and the generalization occurs between subsets of nodes in this graph.
Graph classification tasks, by contrast, always appear in an \emph{inductive} setting. The only major difference between graph classification and inductive node classification is that prior to final prediction, the hidden representations of all nodes are pooled into a single graph-level representation.
%Therefore we analyze  inductive and transductive tasks separately, as opposed to a graph-level vs. node-level split.
In the following two subsections, we provide an analysis of the sensitivity profiles for datasets with inductive and transductive tasks.

\vspace*{-3pt}
\subsection{Taxonomy of Inductive Benchmarks} \label{sec:graph-level-res}
\vspace*{-5pt}
\myparagraph{Datasets.}
We examine a total of 24 datasets, 21 of which are equipped with a graph-classification task (inductive by nature) and the other three are equipped with an inductive node-classification task. Of these datasets, 18 are derived from real-world data, while the other six are synthetically generated. % (Table~\ref{table:graph_datasets}).

For real-world data, we consider several domains. Biochemistry tasks are the most ubiquitous, including compound classification based on effects on cancer or HIV inhibition (\ds{NCI1} \& \ds{NCI109} \cite{wale2008}, \ds{ogbg-molhiv} \cite{hu_open_2021}), protein-protein interaction \ds{PPI} \cite{zitnik2017predicting,hamilton2017inductive}, multilabel compound classification based on toxicity on biological targets (\ds{ogbg-moltox21} \cite{hu_open_2021}), and multiclass classification of enzymes (\ds{ENZYMES} \cite{hu_open_2021}). We also consider superpixel-based graph classification as an extension of image classification (\ds{MNIST} \& \ds{CIFAR10} \cite{dwivedi_benchmarking_2020}), collaboration datasets (\ds{IMDB-BINARY} \& \ds{COLLAB} \cite{yanardag2015DGK}), and social graphs (\ds{REDDIT-BINARY} \& \ds{REDDIT-MULTI-5K} \cite{yanardag2015DGK}).

For synthetic data, we have a concrete understanding of their graph domain properties and how these properties relate to their respective prediction tasks. This allows us to derive a deeper understanding of their \emph{sensitivity profiles}.
The six synthetic datasets in our study make use of a varied set of graph generation algorithms. \ds{Small-world} \cite{you2020design} is based on graph generation with the Watz-Strogatz (WS) model; the task is to classify graphs based on average path length. \ds{Scale-free} \cite{you2020design} retains the same task definition, but the graph generation algorithm is an extension of the Barab\'asi-Albert (BA) model proposed by \citet{2002}. \ds{PATTERN} and \ds{CLUSTER} are node-level classification tasks generated with stochastic block models (SBM) \cite{Holland83_SBM}.
\ds{Synthie} \cite{7837955} graphs are derived by first sampling graphs from the well-known Erd\"os-R\'enyi (ER) model, then deriving each class of graphs by a specific graph surgery and sampling of node features from a distinct distribution per each class.
Similarly, \ds{SYNTHETICnew} \cite{FeragenSyntheticNEW} graphs are generated from a random graph, where different classes are formed by specific modifications to the original graph structure and node features. 
Further details of dataset definitions and synthetic graph generation algorithms are provided in Appendix~\ref{sec:datasets-details}.

\begin{figure}[t]
    \centering
    \includegraphics[trim=0 10 10 -2, clip, width=0.85\textwidth]{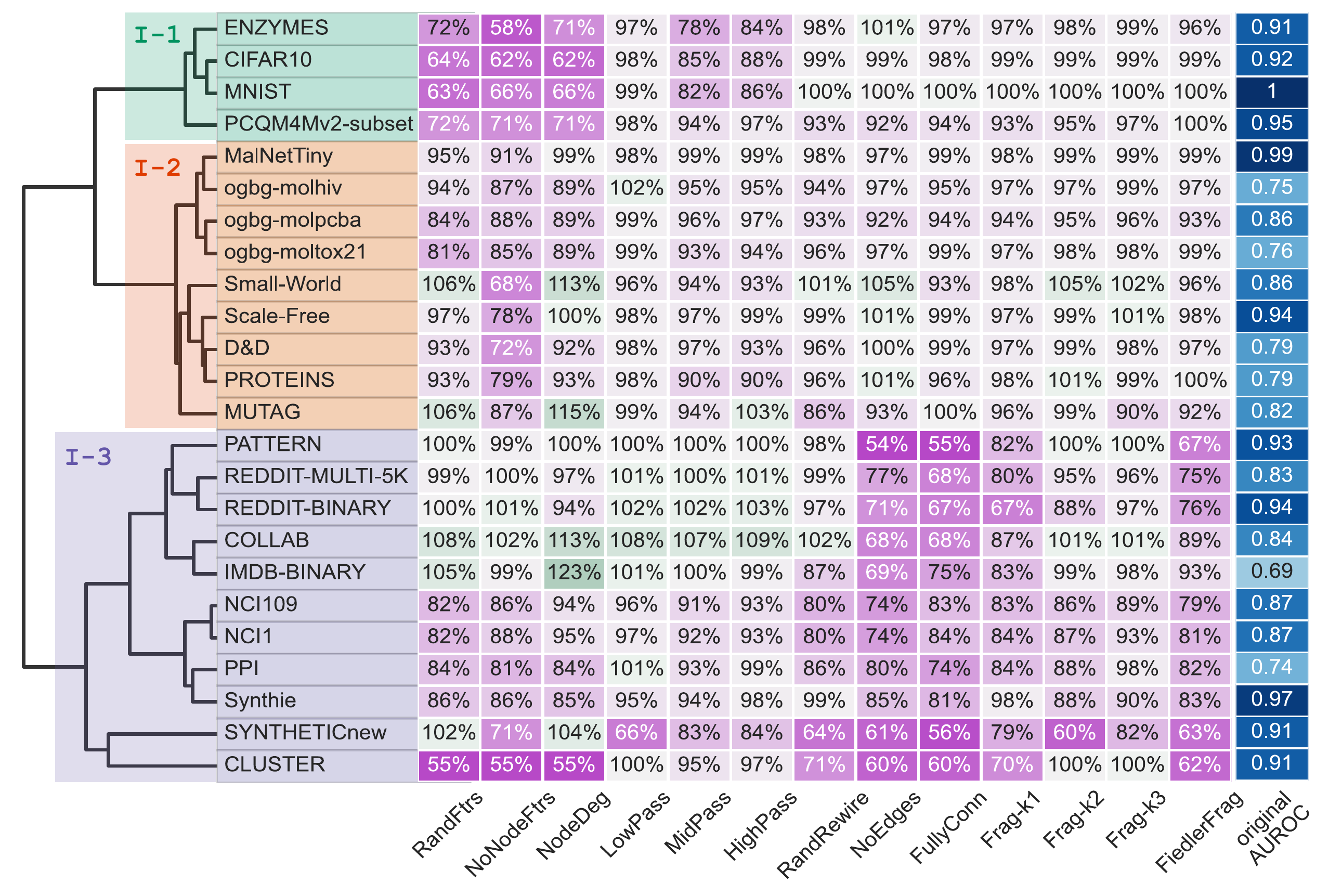}
    \vspace{-5pt}
    \caption{Taxonomy of inductive graph learning datasets via graph perturbations. For each dataset and perturbation combination, we show the GCN model performance relative to its performance on the unmodified dataset.
    }
    \label{fig:gl-taxo}
    % \vspace*{-5pt}
\end{figure}

\myparagraph{Insights.} Here we itemize the main insights into inductive datasets. Our full taxonomy is shown in Figures~\ref{fig:gl-taxo}~and~\ref{fig:inductive-pca}, with a detailed analysis of individual clusters given in Appendix~\ref{sec:inductive-results-details}.
\vspace{-5pt}
\begin{itemize}[leftmargin=1em, itemsep=0.05em]
    \item\textbf{Three distinct groups of datasets.} 
    We identify a categorization into three dataset clusters \I{\{1,2,3\}} that emerge from both the hierarchical clustering and PCA. The datasets in \I{\{1,2\}} exhibit stronger node feature dependency and do not encode crucial information in the graph structure.
    The main differentiating factor between \I{1} and \I{2} is their relative sensitivity to node feature perturbations -- in particular, how well \pert{NodeDeg} can substitute the original node features.
    On the other hand, datasets in \I{3} rely considerably more on graph structure for correct task prediction. This is also reflected by the first two principal components (Figure~\ref{fig:inductive-pca}), where PC1 approximately corresponds to structural perturbations and PC2 to node feature perturbations.
    \item\textbf{No clear clustering by dataset domain.}
    While datasets that are derived in a similar fashion cluster together (e.g., \ds{REDDIT-*} datasets), in general, each of the three clusters contains datasets from a variety of application domains. Not all molecular datasets behave alike; e.g., \ds{ogbg-mol*} datasets in \I{2} considerably differ from \ds{NCI*} datasets in \I{3}.
    \item\textbf{Synthetic datasets do not fully represent real-world scenarios.} \ds{CLUSTER}, \ds{SYNTHETICnew}, and \ds{PATTERN} lie at the periphery of the PCA embeddings, suggesting that existing synthetic datasets do not resemble the type of complexity encountered in real-world data. Hence, one should use synthetic datasets in conjunction with real-world datasets to comprehensively evaluate GNN performance rather than solely relying on synthetic ones. We also note that the sensitivity profiles of all synthetic datasets are well-accounted for w.r.t. their respective design criteria which validate our approach; we refer the reader to Appendix~\ref{sec:inductive-results-details} for a more detailed analysis.
    \item\textbf{Representative set.}
    One can now select a representative subset of all datasets to cover the observed heterogeneity among the datasets. Our recommendation: \ds{PCQM4Mv2-subset}, \ds{CIFAR10} from \I{1}; \ds{D\&D}, \ds{ogbg-molpcba} from \I{2}; \ds{NCI1}, \ds{COLLAB}, \ds{REDDIT-MULTI-5K}, \ds{CLUSTER} from \I{3}.
    \item 
      \parbox[t]{\dimexpr\textwidth-\leftmargin}{%
      \vspace{-2mm}
      \begin{wrapfigure}[14]{R}{0.4\textwidth}
        \centering
        \vspace{-1.1\baselineskip}
        \vspace{1.5mm}
        \includegraphics[trim=2 8 2 2, width=\linewidth]{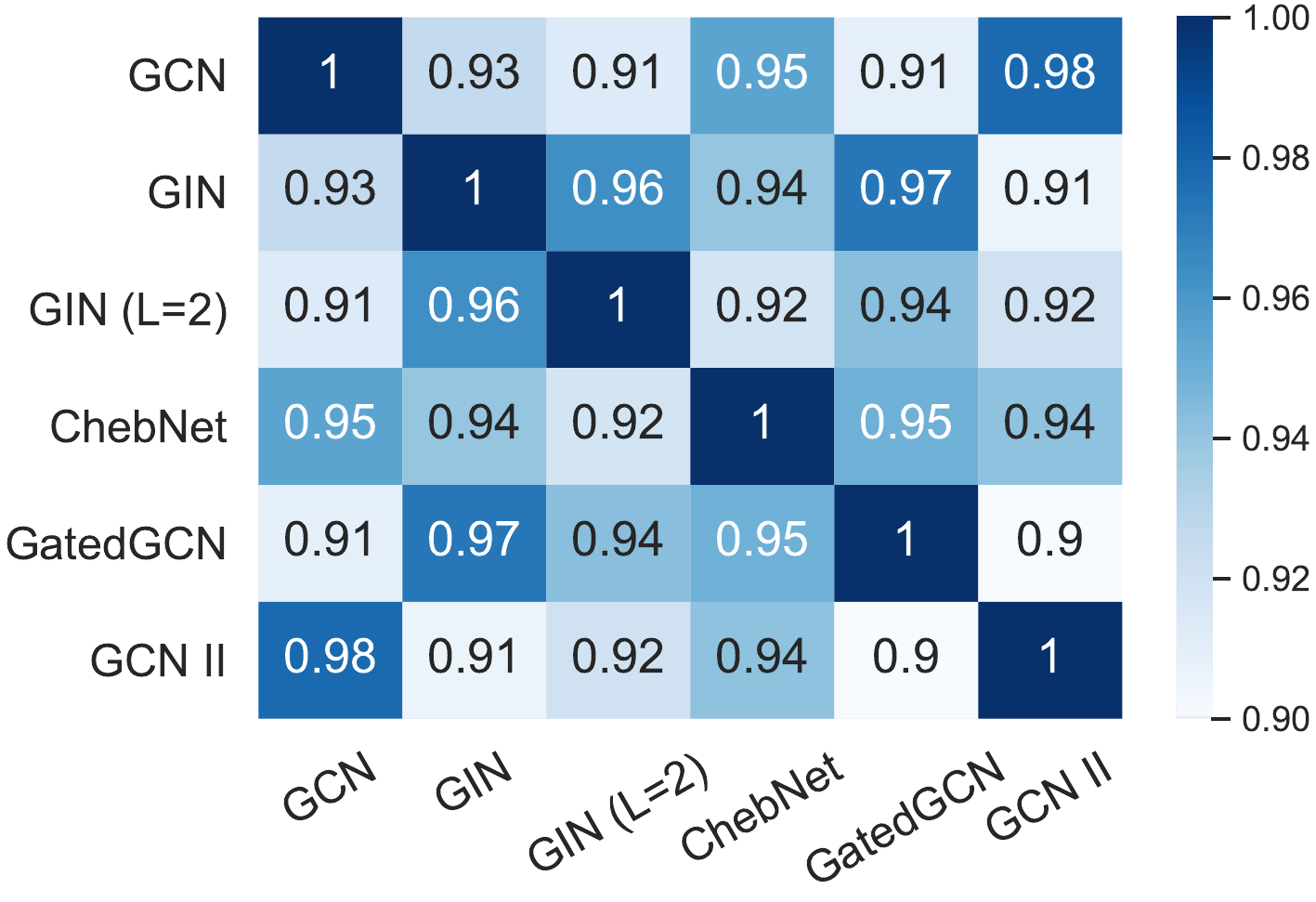}
        \caption{Pearson correlation between profiles derived by six GNN models. %; described in Sec.~\ref{ss:taxonomization}.
        }\label{fig:gnn_corr}
    \end{wrapfigure}
    \textbf{Robustness w.r.t. GNN choice.}
    In addition to GCN, we have performed our perturbation analysis w.r.t. GIN~\cite{xu2018gin}, 2-Layer GIN, ChebNet~\cite{defferrard2016convolutional}, GatedGCN~\cite{bresson2017residual} and GCN~II~\cite{pmlr-v119-chen20v}. These models were selected to cover a variety of inductive model biases: GIN is provably 1-WL expressive, ChebNet uses a higher-order approximation of the Laplacian, GatedGCN employs gating akin to attention, and GCN~II leverages skip connections and identity mapping to alleviate oversmoothing. We have also tested a 2-layer GIN to probe the robustness to the number of message-passing layers. The taxonomies w.r.t. other models (Figure~\ref{fig:gl-taxo-full}) are congruent with that of GCN. Given the differing inductive biases and representational capacity, some differences in the sensitivity profiles are not only expected but desired to validate their functions in benchmarking. The resulting profiles can be used for a detailed comparative analysis of these models, but the overall conclusions remain consistent. This consistency is further validated by our correlation analysis amongst these models, shown in Figure~\ref{fig:gnn_corr}. The Pearson correlation coefficients of all pairs are above 90\%, implying that our taxonomy is sufficiently robust w.r.t. different GNNs and the number of layers.
    }
\end{itemize}

\subsection{Taxonomy of Transductive Benchmarks} \label{sec:node-level-res}

\myparagraph{Datasets.}
We selected a wide variety of 25 transductive datasets with node classification tasks, including citation networks, social networks, and other web page derived networks (see Appendix~\ref{sec:datasets-details}).
In citation networks, such as CitationFull (\ds{CF}) \cite{bojchevski2018deep}, nodes and edges correspond to papers that are linked via citation.
In web page derived networks, like \ds{WikiNet} \cite{pei2020Geom-GCN}, \ds{Actor} \cite{pei2020Geom-GCN}, and \ds{WikiCS} \cite{mernyei2020wikics}, they correspond to hyperlinks between pages.
In social networks, like Deezer (\ds{DzEu}) \cite{rozemberczki2021multiscale}, LastFM (\ds{LFMA}) \cite{rozemberczki2021multiscale}, \ds{Twitch} \cite{rozemberczki2020characteristic}, Facebook (\ds{FBPP}) \cite{rozemberczki2020characteristic}, \ds{Github} \cite{rozemberczki2020characteristic}, and \ds{Coau} \cite{shchur2018pitfalls}, nodes and edges are based on a type of relationship, such as mutual-friendship and co-authorship.
% Due to suboptimal baseline performance even after a thorough hyperparameter search, we excluded \ds{Twitch-RU} and \ds{Twitch-FR} from our main analysis.
\ds{Flickr} \cite{zeng2020graphsaint} and \ds{Amazon} \cite{shchur2018pitfalls} are constructed based on other notions of similarity between entities, such as co-purchasing and image property similarities.
\ds{WebKB} \cite{pei2020Geom-GCN} contains networks of university web pages connected via hyperlinks. It is an example of a \emph{heterophilic} dataset \cite{mostafa_local_2021}, since immediate neighbor nodes do not necessarily share the same labels (which correspond to a user's role, such as faculty or graduate student). By contrast,  \ds{Cora}, \ds{CiteSeer}, and \ds{PubMed} are known to be  \emph{homophilic} datasets where nodes within a neighborhood are likely to share the same label. In fact, no less than 60\% of nodes in these networks have neighborhoods that share the same node label as the central node~\cite{mernyei2020wikics}.
% In particular, no less than 60\% of the node neighborhoods in \ds{Cora}, \ds{CiteSeer}, and \ds{PubMed} are completely homogeneous, meaning that all nodes within a neighborhood share the same node label as the central node \cite{mernyei2020wikics}.

\begin{figure}[t]
    \centering
    % trim: left bottom right top
    \includegraphics[trim=10 10 10 5, clip, width=0.85\textwidth]{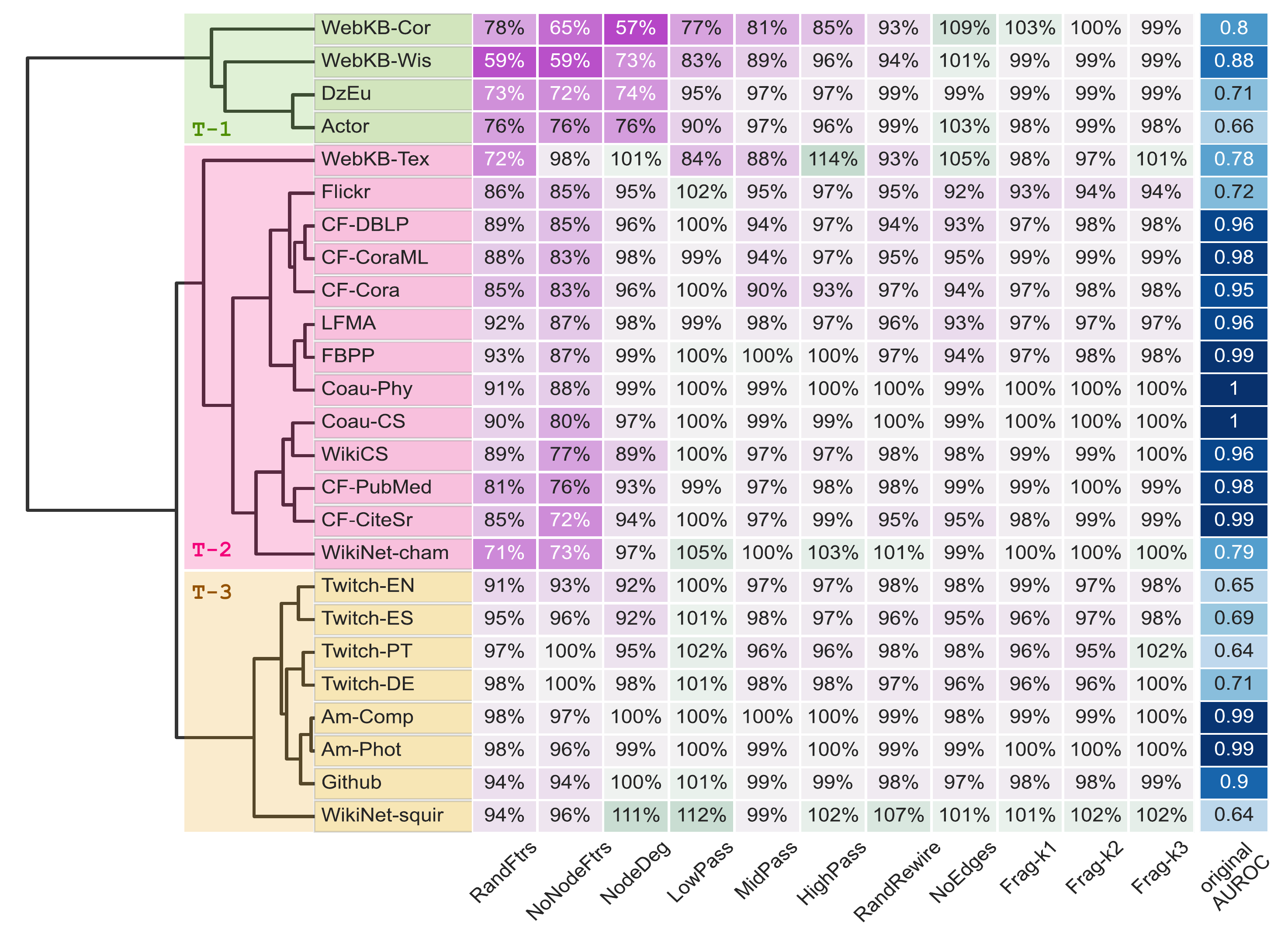}
    \vspace{-5pt}
    \caption{Taxonomization of transductive datasets based on sensitivity profiles w.r.t. a GCN model.
    }
    \label{fig:nl-taxo}
    \vspace{-5pt}
\end{figure}

\myparagraph{Insights.} Below we list the main insights into transductive graph datasets and their taxonomy (Figures~\ref{fig:nl-taxo}~and~\ref{fig:transductive-pca}). We refer the reader to Appendix~\ref{sec:transductive-results-details} for the analysis of individual clusters.
%\vspace{-5pt}
\begin{itemize}[leftmargin=1em, itemsep=0.05em]
    \item\textbf{Transductive datasets are uniformly insensitive to structural perturbations.}
    Sensitivity profiles of \emph{all} transductive datasets show high robustness to \emph{all} graph structure perturbations. This is in stark contrast with the inductive datasets, where the largest cluster \I{3} is defined by high sensitivity to structural perturbations. The graph connectivity may not be vital to every dataset/task, e.g., in \ds{WikiCS} word embeddings of Wikipedia pages may be sufficient for categorization without hyperlinks. While the observation that no dataset significantly depends on structural information is startling, it corroborates with the reported strong performance of MLP or similar models augmented with label propagation to outperform GNNs in several of these transductive datasets \cite{gasteiger2018APPNP,huang2020CorrectAndSmooth}. % This might have several implications: (i) the benchmark datasets may tend to embed structural information as node features, rendering most actual structure redundant, and (ii) the current benchmarking datasets are deficient in capturing reliance on long-distance interactions.
    
    \item\textbf{Three distinct groups of datasets.} 
    The transductive datasets are also categorized into three clusters as \T{\{1,2,3\}}. \T{1} consists of heterophilic datasets, such as \ds{WebKB} and \ds{Actor}~\cite{mostafa_local_2021,ma2021homophily}. These are well-separated from others, as seen in the right half of the PCA plot (Figure~\ref{fig:transductive-pca}), primarily via PC1 and characterized by performance drop due to removal of the original node features (\pert{NoNodeFtrs}, \pert{RandFtrs}) and their replacement by node degrees (\pert{NodeDeg}). \T{3} is indifferent to both node and structure removal, implying redundancies between node features and graph structure for their tasks. \T{2} datasets, on the other hand, experience significant performance degradation on \pert{NoNodeFtrs} and \pert{RandFtrs}, yet these drops are recovered in \pert{NodeDeg}. This indicates that \T{2} datasets have tasks for which structural summary information is sufficient, perhaps due to homophily.
    %rely on structural summary information that is encoded in their node features.
    
    %\item\textbf{Heterophilic datasets are well-separated from others.}
    %Located in the right half of the PCA plot (Figure~\ref{fig:transductive-pca}), \T{1} consists of heterophilic datasets, such as \ds{WebKB} and \ds{Actor}~\cite{mostafa_local_2021,ma2021homophily}. In particular, \T{1} datasets are separated from \T{2} and \T{3} via PC1, which is primarily characterized by performance drop due to removal of the original node features (\pert{NoNodeFtrs}) and their replacement by node degrees (\pert{NodeDeg}).
    % Since node degree encodes immediate neighborhood information, it is expected that it does not work for heterophilic datasets that require higher order neighborhood structures. The distinct profiles of these datasets imply that their inclusion to such processes may lead to a more complete benchmarking.
    
    \item\textbf{Representative set.} Many datasets have very close sensitivity profiles, thus factoring in also the graph size and original AUROC (avoiding saturated datasets), we make the following recommendation: \ds{WebKB-Wis}, \ds{Actor} from \T{1}; \ds{WikiNet-cham}, \ds{WikiCS}, \ds{Flickr} from \T{2}; \ds{WikiNet-squir}, \ds{Twitch-EN}, \ds{GitHub} from \T{3}.
\end{itemize}

%###########################################################
\vspace{-5pt}
\section{Discussion}\label{sec:discussion}
\vspace{-3pt}
%###########################################################

%########
% BRAINSTORMING AREA -- potential discussion points

% The dependency on task for similar looking datasets: Interestingly, even though PROTEINS and ENZYMES contain similar protein structure graphs, their different prediction tasks lead to notably different perturbation sensitivities; this illustrates the task dependency of the optimal graph representations.

% as validation of our approach: similarly derived dataset pairs with similar tasks such as \ds{NCI1} vs. \ds{NCI109} and \ds{REDDIT-BINARY} vs. \ds{REDDIT-MULTI-5K} produce highly similar sensitivity profiles. this was dismissed at some point but could be reintegrated if we also assert that the tasks remain similar.

% NodeDeg beating original AUROC is a common phenomenon especially in inductive tasks. It indicates that it is possible to improve existing datasets (especially when using GCNs/less expressive models) by data augmentation by adding more structure-derived node features (NodeDeg, clustering coefficients, pagerank of each node etc.). Especially for datasets that have no node features, this seems to lead to a cleaner structural signal and better than AUROC performance. So in a way, NodeDeg is a litmus test to see if performance of a given dataset can be improved via data augmentation (exploiting structural info as node features). not exactly novel, but a low hanging fruit perhaps.

%########
%%%%% GENERAL INSIGHTS %%%%%

Our results quantify the extent to which graph features or structures are more important for the downstream tasks; a vital question brought up in classical works on graph kernels~\cite{kriege2020survey,schulz2019necessity}.
We observed that more than half of the datasets contain rich node features. On average, excluding these features reduces GNN prediction performance more than excluding the entire graph structures, especially for transductive node-level tasks. Furthermore, low-frequency information in node features appears to be essential in most datasets that rely on node features.
Historically, most graph data aimed to capture closeness among entities,
%and the corresponding tasks are sufficiently inferred from these relationships, 
which has prompted the development of local aggregation approaches, such as label propagation, personalized page rank, and diffusion kernels \cite{kohler2008walking,cowen2017networkpropagation}, all of which share a common principle of low pass filtering. High-frequency information, on the other hand, may be important in recently emerging application areas, such as combinatorial optimization, logical reasoning or biochemical property prediction, which require complex non-local representations.
%is more intensively utilized in many image processing tasks such as texture synthesis \cite{he2021texture,gatys2015texture} and style transfer \cite{gatys2016image,yoo2019photorealistic}.
% (Only until recently, graph datasets equipped with tasks that require more sophisticated handling of the graphs have emerged, such as combinatorial optimization, logical reasoning, and complex biomedical properties prediction \textcolor{blue}{Need help with citation, not really familiar with comb opt and logreas. Also not sure how does this tie in with our analysis}.)
% This is likely to be connected to the ``basic'' nature of classification tasks where each data input (e.g., image) is classified as a whole; more intensive tasks such as texture synthesis \cite{he2021texture,gatys2015texture}, and style transfer \cite{gatys2016image,yoo2019photorealistic}, are bound to rely on high frequency information much more than image classification.
% \textcolor{blue}{(add a sentence to suggest that maybe we need to move towards creating some more ``interesting'' datasets?)}
% Meanwhile, despite recent interest in different aspects of graph datasets such as long-range dependencies and heterophily, the availability of such datasets remain lacking.

Further, despite the recent interest in the development of new methods that could leverage long-range dependencies and heterophily, the availability of adequate benchmarking datasets remains lacking or less readily accessible.
Meanwhile, some recent efforts, such as GraphWorld~\cite{palowitch2022graphworld}, aim to comprehensively profile a GNN's performance using a collection of synthetic datasets that cover an entire parametric space. Notably, our analysis demonstrates that synthetic tasks do not fully resemble the complexity of real-world applications. Hence, benchmarking made purely by synthetic datasets should be taken with caution, as the behavior might not be representative of real-world scenarios.
% Although synthetic datasets enable direct testing of certain properties of GNN models, they generally do not resemble the full complexity of real world data and appear to be best used as a complement to the real world derived datasets rather than a sufficient replacement.
%\textcolor{blue}{(this has not been discussed explicitly in the results section currently, need to either add this discussion or remove this final sentence)}.

As a comprehensive benchmarking framework, our work provides several potential use cases beyond the taxonomy analysis presented here.
One such usage is understanding the characteristics of any new datasets and how they are related to existing ones.
For example, DeezerEurope (\ds{DzEu}) is a relatively new dataset~\cite{rozemberczki2021multiscale} that is less commonly benchmarked and studied than the other datasets we consider.
The inclusion of \ds{DzEu} in \T{1} suggested its heterophilic nature, which indeed has been recently demonstrated \cite{lim2021large}.
On the other hand, since the sensitivity profiles naturally suggest the invariances that are important for different datasets from a practical standpoint, they could provide valuable guidance to the development of self-supervised learning and data augmentations for GNNs~\cite{xie2022self}.
% For example, \pert{NodeDeg} perturbation considerably increases the performance for \ds{WikiNet-squir} (\T{3}) and \ds{IMDB-BINARY} (\I{3}), indicating that augmenting the node features with node degree features for these datasets can be helpful.
% Meanwhile,

Finally, we observed that overall patterns in \emph{sensitivity profiles} remain similar regardless of whether we used GCN, GIN, or the other 4 models to derive them. Subtle differences in sensitivity profiles w.r.t. different GNN models are not only expected but also desired when comparing models that have distinct levels of expressivity. While we expect overall patterns to be similar, more expressive models should provide enhanced resolution.
%as dataset properties do not change over models; we see that our framework manages this trade-off well. In the future, comparison of multiple architectures could also yield insights about the datasets when various theoretical implications of the architectures are taken into consideration.
One could then contrast taxonomization w.r.t. first-order GNNs (such as those we used) with more expressive higher-order GNNs, Transformer-based models with global attention, and others. We hope our work will also inspire future work to empirically validate the expressivity of new graph learning methods in this vein beyond classical benchmarking.

\myparagraph{Limitations and Future Work.}
Our perturbation-based approach is fundamentally limited in that we cannot test the significance of a property that we cannot perturb or that the reference GNN model cannot capture. Therefore, designing more sophisticated perturbation strategies to gauge specific relations could bring further insight into the datasets and GNN models alike. New perturbations may gauge the usefulness of geometric substructures such as cycles~\cite{bevilacqua2021ESAN} or the effects of graph bottlenecks, e.g., by rewiring graphs to modify their ``curvatures'' \cite{topping2021understanding}.
Other perturbations could include graph sparsification (edge removal) \cite{spielman2010spectral} and graph coarsening (edge contraction) \cite{brugnone2019coarse,bodnar2021deep}.

A number of OGB node-level datasets are not included in this study due to the memory cost of typical MPNNs. Conducting an analysis based on recent scalable GNN models~\cite{fey2021gnnautoscale} would be an interesting avenue of future research. Further, we only considered classification tasks, omitting regression tasks, as their evaluation metrics are not easily comparable. One way to circumvent this issue would be to quantize regression tasks into classification tasks by binning their continuous targets.
Additionally, we disregarded edge features in two OGB molecular datasets we used. In a future work, edge features could be leveraged by an edge-feature-aware generalization of MPNNs. The importance of edge features can then be analyzed by introducing new edge-feature perturbations. We also limited our analysis to node-level and graph-level tasks, but this framework could be further extended to link-prediction or edge-level tasks. While our perturbations could be used in this new scenario as well, new perturbations, such as graph sparsification, would need to be considered. Similarly, hallmark models for link and relation predictions, outside MPNNs, should be considered.

% Finally, our taxonomy can be further automated by removing the reliance on human decisions in drawing cluster boundaries. Ideally, these boundaries should be determined automatically based on statistical significance notions in hierarchical cluster assignments. Such an approach was recently proposed in~\cite{kimes2017statistical}, but it relies on fitting a Gaussian to each cluster, thus requiring more samples than available in our collection of datasets. Further, one needs to be careful not to include highly similar perturbations, as these could introduce very similar sensitivity profiles across all datasets, thus artificially up-weighting certain aspects over others. One potential solution would be to perform hierarchical clustering on only a few top principal components rather than the full sensitivity profiles to effectively truncate highly correlated perturbations. We regard such extensions as potentially interesting, but ultimately out of scope for this study.

% %###########################################################
\vspace{-5pt}
\section{Conclusion}
\vspace{-2pt}
% %###########################################################
% As the field of graph representation learning develops, the importance of setting up comprehensive and informative benchmarking processes that evaluate these developments rapidly increases in tandem. Nevertheless, we have argued that it has been difficult today to make informed decisions about benchmarking data selection in the relatively short history of graph learning, primarily due to a limited understanding of how datasets differ in their encoding and propagation of information. 

% In this paper, we have presented a solution to this problem by constructing a method for defining and analyzing the information flow in graph data, and have come up with the most complete taxonomy of graph datasets in literature based on the analysis of this information flow on a multitude of graph datasets, covering graph and node prediction tasks, on inductive and transductive datasets. We have taken care to ensure that our methodology provides not just a snapshot of the current state of benchmarking in graph representation learning, but also a dynamic framework that can grow as the field progresses further by its application to future graph datasets and models. We also facilitate the dynamic nature of our methodology by providing the source code for our taxonomy; we hope that this will make it easier for researchers to both use our framework in developing their benchmarks, and further extend it to suit their needs.

We provide a systematic data-driven approach for taxonomizing a large collection of graph datasets -- the first study of its kind.
The core principle of our approach is to gauge the essential characteristics of a given dataset with respect to its accompanying prediction task by inspecting the downstream effects caused by perturbing its graph data. The resulting sensitivities to the diverse set of perturbations serve as ``fingerprints'' that allow identifying datasets with similar characteristics.
We derive several insights into the current common benchmarks used in the field of graph representation learning and make recommendations on the selection of representative benchmarking suits. Our analysis also puts forward a foundation for evaluating new benchmarking datasets that will likely emerge in the field.

%###########################################################
% Do NOT include Acknowledgments in the anonymized submission, only in the final paper.
%###########################################################
\section*{Acknowledgements}
This work was partially funded by \mbox{Fin-ML} CREATE graduate studies scholarship for PhD~[\emph{F.W.}]; IVADO (Institut de valorisation des données) grant PRF-2019-3583139727, FRQNT (Fonds de recherche du Québec - Nature et technologies) grant 299376, Canada CIFAR AI Chair~[\emph{G.W.}]; NSF (National Science Foundation) grant DMS-1845856~[\emph{M.H.}]; and NIH (National Institutes of Health) grant NIGMS-R01GM135929~[\emph{M.H.,G.W.}] The content provided here is solely the responsibility of the authors and does not necessarily represent the official views of the funding agencies.

%%%%%%%%%%%%%%%%%%%%%%%%%%%%%%%%%%%%%%%%%%%%%%%%%%%%%%%%%%%%
% \clearpage
\bibliographystyle{plainnat}
\bibliography{references}

\clearpage
\appendix
\renewcommand\thefigure{\thesection.\arabic{figure}}
\renewcommand\thetable{\thesection.\arabic{table}}
% Optionally include extra information (complete proofs, additional experiments and plots) in the appendix.
% This section will often be part of the supplemental material.

%%%%%%%%%%%%%%%%%%%%%%%%%%%%%%%%%%%%%%%%%%%%%%%%%%%%%%%%%%%%%%%%%%%%%%%%%%%%%%%
\section{Extended Methods} \label{sec:method-details}
\setcounter{figure}{0}
\setcounter{table}{0}
%%%%%%%%%%%%%%%%%%%%%%%%%%%%%%%%%%%%%%%%%%%%%%%%%%%%%%%%%%%%%%%%%%%%%%%%%%%%%%%

\begin{figure}[!ht]
    \centering
    \includegraphics[width=0.8\textwidth]{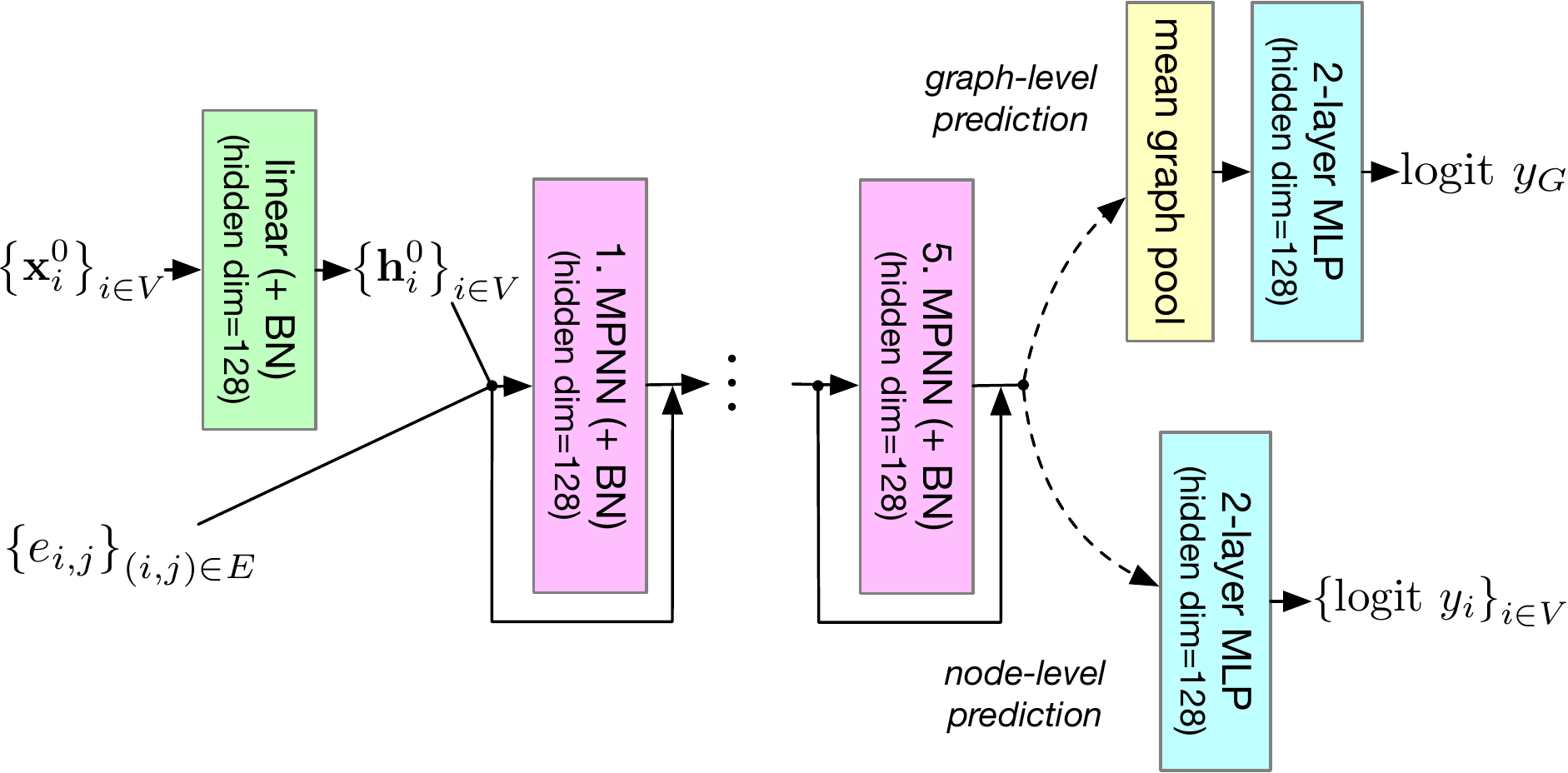}
    % \vspace{-3pt}
    \caption{MPNN model blueprint used for all datasets.}
    \label{fig:mpgnn}
\end{figure}

\subsection{Taxonomization by Hierarchical Clustering}
To study a systematic classification of the graph datasets, we use Ward's method \cite{ward1963hierarchical} for hierarchical clustering analysis on their \emph{sensitivity profiles}.
Specifically, we first construct a perturbation sensitivity matrix where each row represents a dataset and each column represents a perturbation. 
An entry in this matrix is computed by taking the ratio between the test score achieved with the perturbed dataset and the test score achieved with the original dataset. As our performance metric we use the area under the receiver operating characteristic (AUROC) averaged over 10 random seed runs or 10 cross-validation folds, depending on whether a dataset has predefined data splits or not. Row-wise hierarchical clustering provides us a data-driven taxonomization of the datasets.

Using AUROC as our metric, the values of the perturbation sensitivity matrix range from $0.5$ to $1$ when a perturbation causes a loss in predictive performance, and from $1$ to $2$ when it improves it. Therefore we element-wise $\log_2$-transform the matrix to balance the two ranges and map the values onto $[ -1, 1 ]$ before hierarchical clustering. Yet, for a more intuitive presentation, we show the original ratio values as percentages throughout this paper.

\subsection{MPNN Hyperparameter Selection}
We keep the model hyperparameters, illustrated in Figure~\ref{fig:mpgnn}, identical for each dataset and perturbation combination. We use a linear node embedding layer, 5 graph convolutional layers with residual connections and batch normalization (only for inductive datasets), followed by global mean pooling (in case of graph-level prediction tasks), and finally a 2-layer MLP classifier. For training we use Adam optimizer \cite{kingma2017adam} with learning rate reduction by 0.5 factor upon reaching a validation loss plateau.
% We run each graph-level experiment for 300 epochs with an initial learning rate of 0.001 and a scheduler patience of 15 epochs; node-level experiments are run for 5000 epochs with an initial learning rate of 0.01 and a scheduler patience of 100 epochs.
Early stopping is done based on validation split performance.

\myparagraph{Implementation.} Our pipeline is built using PyTorch~\cite{NEURIPS2019_9015} and
PyG~\cite{Fey/Lenssen/2019} with GraphGym~\cite{you2020design} (provided under MIT License). Its modular \& scalable design facilitated here one of the most extensive experimental evaluations of graph datasets to date.

\myparagraph{Computing environment and used resources.}\label{app:cluster}
All experiments were run in a shared computing cluster environment with varying CPU and GPU architectures. These involved a mix of NVidia V100 (32GB), RTX8000 (48GB), and A100 (40GB) GPUs. The resource budget for each experiment was 1 GPU, 4 CPUs, and up to 32GB system RAM.

%%%%%%%%%%%%%%%%%%%%%%%%%%%%%%%%%%%%%%%%%%%%%%%%%%%%%%%%%%%%%%%%%%%%%%%%%%%%%%%
\clearpage
\section{Extended Results} \label{sec:results-details}
\setcounter{figure}{0}
\setcounter{table}{0}
%%%%%%%%%%%%%%%%%%%%%%%%%%%%%%%%%%%%%%%%%%%%%%%%%%%%%%%%%%%%%%%%%%%%%%%%%%%%%%%

\subsection{Taxonomy of Inductive Benchmarks} \label{sec:inductive-results-details}

% I-1
\myparagraph{\I{1}: Node-feature reliance.} The top-most cluster \I{1}, while mostly indifferent to structural perturbations, is highly sensitive to node feature perturbations that comprise the left-hand-side columns in Figure~\ref{fig:gl-taxo}. The presence of image-based datasets \ds{MNIST} and \ds{CIFAR10} in this cluster is not surprising, as for superpixel graphs the structure loosely follows a grid layout for all classes, meaning determining class solely based on structure is difficult. Additionally, the coordinate information of superpixels is encoded also in the node features, together with average pixel intensities.
A model with powerful enough classifier component is then sufficient for achieving high accuracy using these node features alone.
Furthermore, the sensitivity of these datasets to \pert{MidPass} and \pert{HighPass} indicates that the overall shape of the signals encoded by low-frequencies is more informative for classifying the image content than sharp superpixel transitions encoded by high-frequencies. The presence of \ds{ENZYMES} in \I{1} is likely due to the fact that some of the node features are precomputed using graph kernels, and therefore are sufficient to distinguish the enzyme classes in the dataset when structural information is removed. Last but not least, \ds{PCQM4Mv2-subset} dataset appears to have a complex task that is dominated by the node feature information, yet the graph structure encodes non-negligible information as well. Out of all datasets in the \I{1} cluster, \ds{PCQM4Mv2-subset} is the most sensitive one to structural perturbations. This corroborates with the expectation that predicting the HOMO-LUMO gap, which is the energy difference between the highest occupied molecular orbital (HOMO) and lowest unoccupied molecular orbital (LUMO), is a complex task that heavily depends on atom types, their bonds, and relative distances.

% I-2 -- Main message: node features are dominant, but unlike in I-1 they can be partially replaced by structural information extracted by GNN from the graphs. This corroborates with what we know about the datasets in this cluster, that they often have structure-derived features already present in the original node features. This structural information is important and can be recovered, however some of the other important information (e.g. atom types) cannot.
\myparagraph{\I{2}: Node features contain majority of necessary structural information.} 
For datasets in \I{2}, the graph structural information is again not necessary for achieving the baseline performance if the original node features are present, while the performance deteriorates noticably if \pert{NoNodeFtrs} is applied. However, unlike \I{1}, these datasets are much less affected overall by the perturbations on node features. Many of the node features on these datasets are themselves derived from the graph's geometry, and it seems MPNNs are able to use either the graph structure or the node features to compensate for the absence of the other when encountering perturbed graphs. It appears that the low/mid/high-pass filterings in particular are able to retain a significant amount of  geometric information. 
% I-2, synthetic

The synthetic graphs of \ds{Scale-Free} and \ds{Small-world} (both \I{2} datasets) are generated through different algorithms (WS and BA, respectively), but the node features and tasks are equivalent: The features are the local clustering coefficient and PageRank score of each node and the task is to classify graphs based on average path length. Since the encoded features are derived from graph structure itself, MPNNs are still able to exploit them when the original graph structure is perturbed. When the MPNNs are forced to rely on graph structure instead, they are still able to attain AUROCs above random despite some decrease.

For many of the \I{2} datasets, \pert{NodeDeg} allows one to replace geometric information of original node features with new geometric information, the degree of each vertex, to large success -- for some of them the original AUROC scores are recovered and even surpassed, possibly due to \pert{NodeDeg} reinforcing the existing structural signal. This trend is not as pronounced when the GIN-based model is used, since GIN achieves a comparatively high level of performance even in the face of \pert{NoNodeFtrs}, likely due to the higher expressiveness of GIN compared to GCN in distinguishing of structural patterns.

On the other hand, there are datasets of biochemical origin in this cluster, whose node features encode chemical and physical attributes, such as atom or amino acid type. Except \ds{MUTAG}, there appears to be some information encoded in these node features that is irreplaceable by graph structure or node degree information.

% I-3, subcluster 1
\myparagraph{\I{3}: Graph-structure reliance.} The \I{3} cluster is characterized by strong structural dependencies, and can be further divided into two subgroups based on their sensitivities to node feature perturbations.

The first subgroup, which consists of \ds{PATTERN}, \ds{COLLAB}, \ds{IMDB-BINARY} and \ds{REDDIT}, is not affected by node feature perturbations. These datasets do not have any original informative node features and their tasks appear to be purely structure-based. Indeed, in the case of \ds{PATTERN} the task is to detect structural patterns in graphs, rendering node features irrelevant for the task. 
On the other hand, structural perturbations such as \pert{NoEdges} and \pert{FullyConn} cause drastic performance drops in this group, since most of its task signals are sourced from graph structures.
% On the other hand, since most of the task signal here is sourced from the graph structure, structural perturbations such as \pert{NoEdges} and \pert{FullyConn} cause drastic drops in performance.
This group also exhibits limited to no sensitivity towards \pert{Frag-k2} and \pert{Frag-k3} perturbations, which test for degrees of reliance on longer range interactions by limiting information propagation to $\{2, 3\}$ hops. We still see prominent sensitivity to \pert{Frag-k1}, though, implying reliance on information from immediate neighbors. We can attribute the insensitivity for $k > 1$ to inherent graph properties for some of these datasets: For dense networks like \ds{PATTERN} or ego-nets such as \ds{IMDB-BINARY} and \ds{COLLAB}, just 1 or 2 hops recover the original graph -- for these graphs, the notion of long-range information does not exist.

% I-3, subcluster 2
The second \I{3} subgroup, formed by \ds{NCI} datasets and \ds{Synthie}, are the datasets that are notably affected by \emph{all} perturbations. For \ds{Synthie}, this sensitivity stems from its construction. The four synthetic classes in \ds{Synthie} are formed by combinations of two distributions of graph structures and two distributions of node features -- elimination of either leads to a \emph{partial} collapse in the distinguishability of two classes. The \ds{NCI} classification tasks, similarly to related bioinformatics datasets in \I{2}, show a degree of reliance on the high-dimensional node features, but additionally, they are also dependent on non-local structure as they are among the datasets most adversely affected by \pert{Frag-k2} and \pert{Frag-k3}.

% I-3, outliers
Synthetic datasets \ds{CLUSTER} and \ds{SYNTHETICnew} are also adversely affected by both structural and node feature perturbations. However, they stand out due to the magnitude of this effect. Many of the perturbations lead to a major decrease in AUROC and close-to-random performance. A closer inspection can provide an explanation. The task of \ds{CLUSTER} is semi-supervised clustering of unlabeled nodes into six clusters, and the true cluster labels are given as node features in only a single node per cluster. \pert{NoEdges} and \pert{FullyConn} remove the cluster structure altogether, while \pert{NoNodeFtrs} and \pert{NodeDeg} remove the given cluster labels, rendering the task  unsolvable in either case. In \ds{SYNTHETICnew}, the two classes are derived from a ``base'' graph by a class-specific edge rewiring and node feature permutation, hence either graph structure or node features should differentiate the classes. Despite such expectation, we observe that the original node features alone are not sufficient, as structure perturbations have detrimental impact on the prediction performance. On the other hand GIN and GCN with \pert{NodeDeg} can learn to distinguish the two classes even without the original node features. Thus, the original node features appear to be unnecessary, while after bandpass-filtering even provide misleading signal.

\begin{figure}[H]
\centering
     \begin{subfigure}[b]{\textwidth}
     \begin{adjustwidth}{-1.1 cm}{-1.1 cm}
         \centering
         % trim: left bottom right top
         \includegraphics[align=c, trim=0 120 0 0, clip, width=0.47\textwidth]{graph_tasks/graph-tasks-PCA.pdf}
         \hfill
         \includegraphics[align=c, trim=0 10 10 0, clip, width=0.68\textwidth]{graph_tasks/graph-tasks-gcn.pdf}
     \end{adjustwidth}
     \vspace{-3pt}
     \caption{Sensitivity profiles by \textbf{GCN} model (reprint of Figure~\ref{fig:inductive-pca}~and~\ref{fig:gl-taxo}).}
     \label{fig:gl-taxo-gcn}
     \end{subfigure}
     \begin{subfigure}[b]{\textwidth}
     \begin{adjustwidth}{-1.1 cm}{-1.1 cm}
         \centering
         \vspace{15pt}
         \includegraphics[align=c, trim=-10 -10 -10 -10, clip, width=0.47\textwidth]{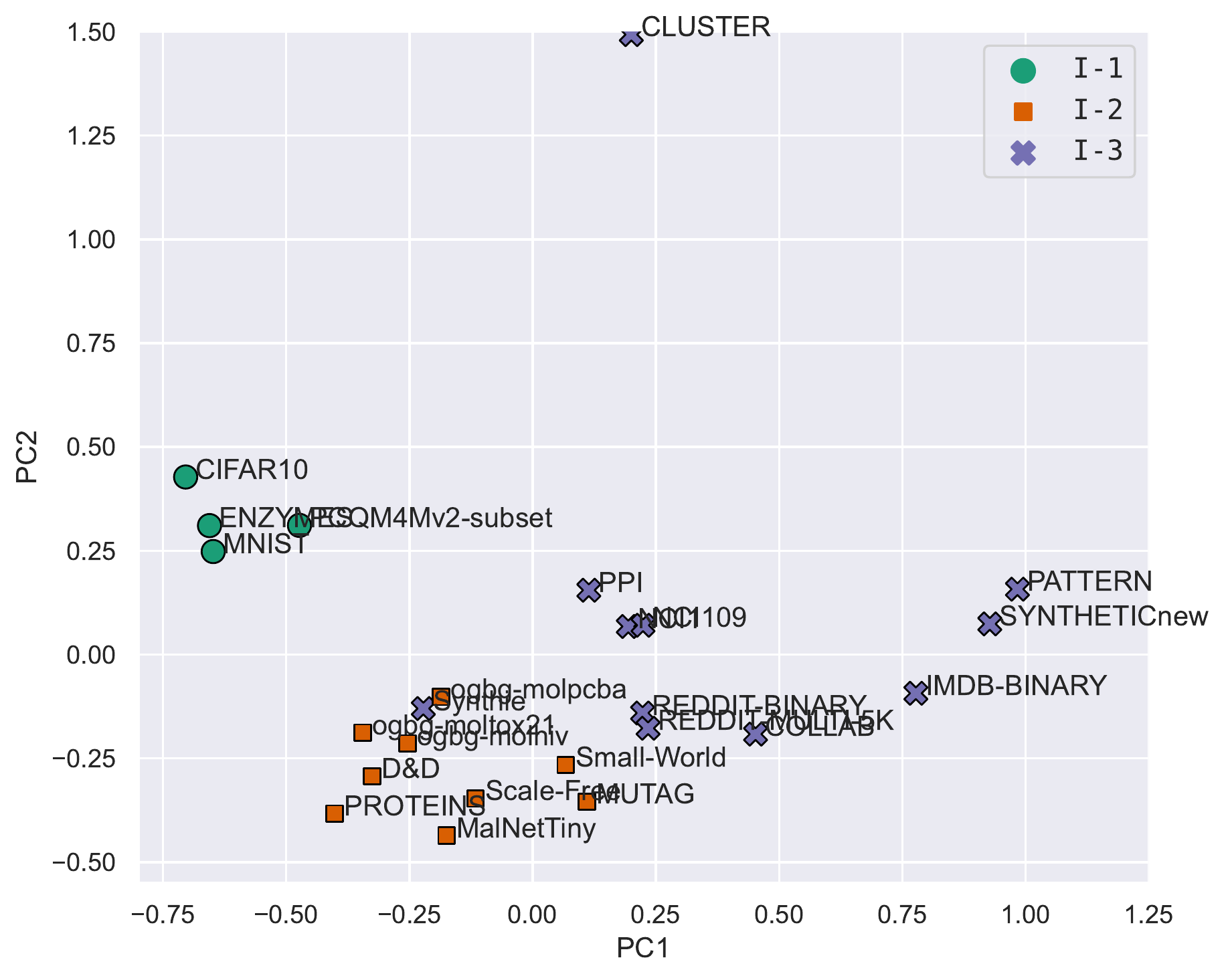}
         \hfill
         \includegraphics[align=c, trim=0 10 10 -2, clip, width=0.68\textwidth]{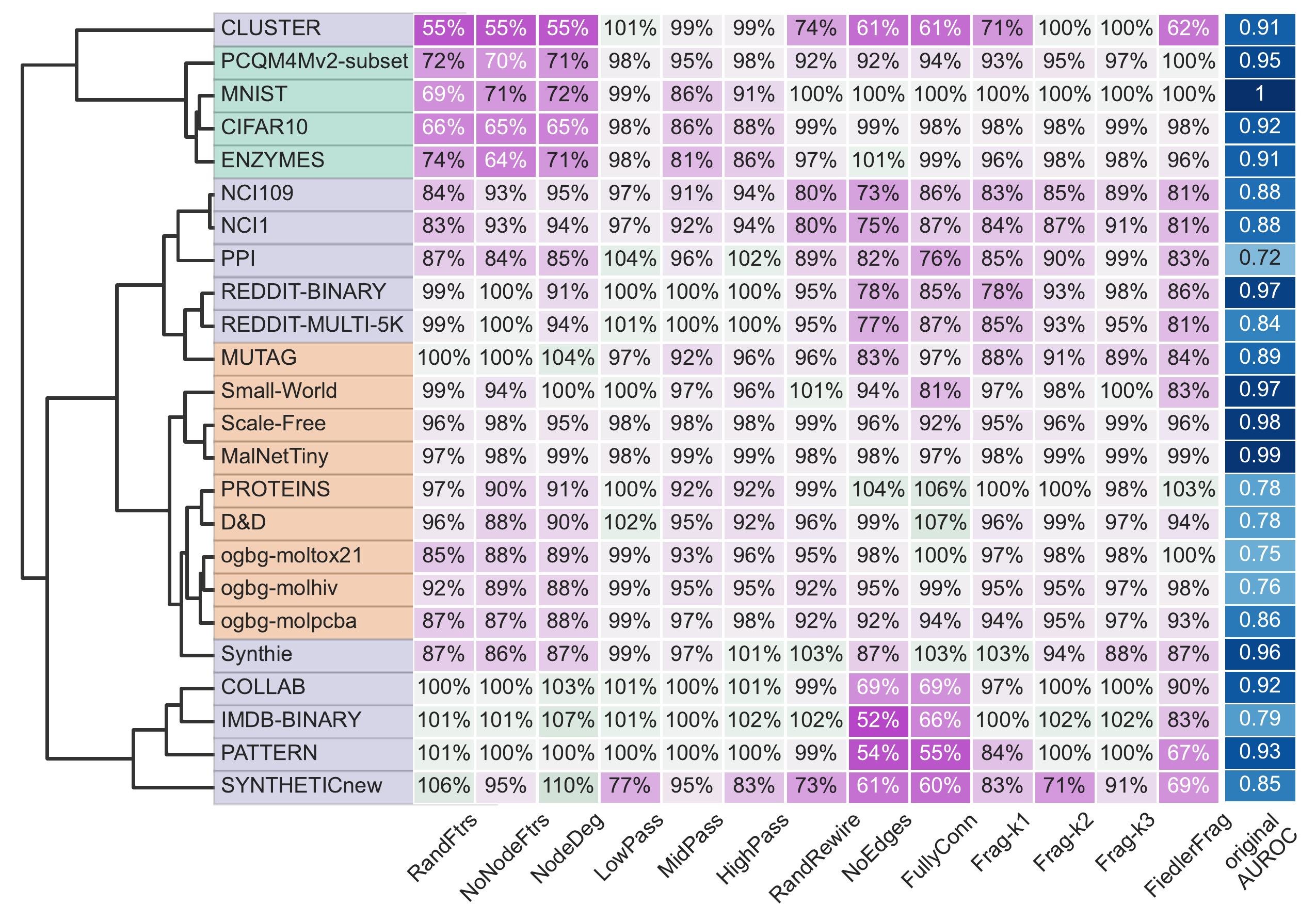}
     \end{adjustwidth}
     \vspace{-3pt}
     \caption{Sensitivity profiles by \textbf{GIN} model; annotated by cluster assignment w.r.t. GCN model.}
     \label{fig:gl-taxo-gin}
     \end{subfigure}
    \begin{subfigure}[b]{\textwidth}
     \begin{adjustwidth}{-1.1 cm}{-1.1 cm}
         \centering
         \vspace{15pt}
         \includegraphics[align=c, trim=0 0 0 0, clip, width=0.47\textwidth]{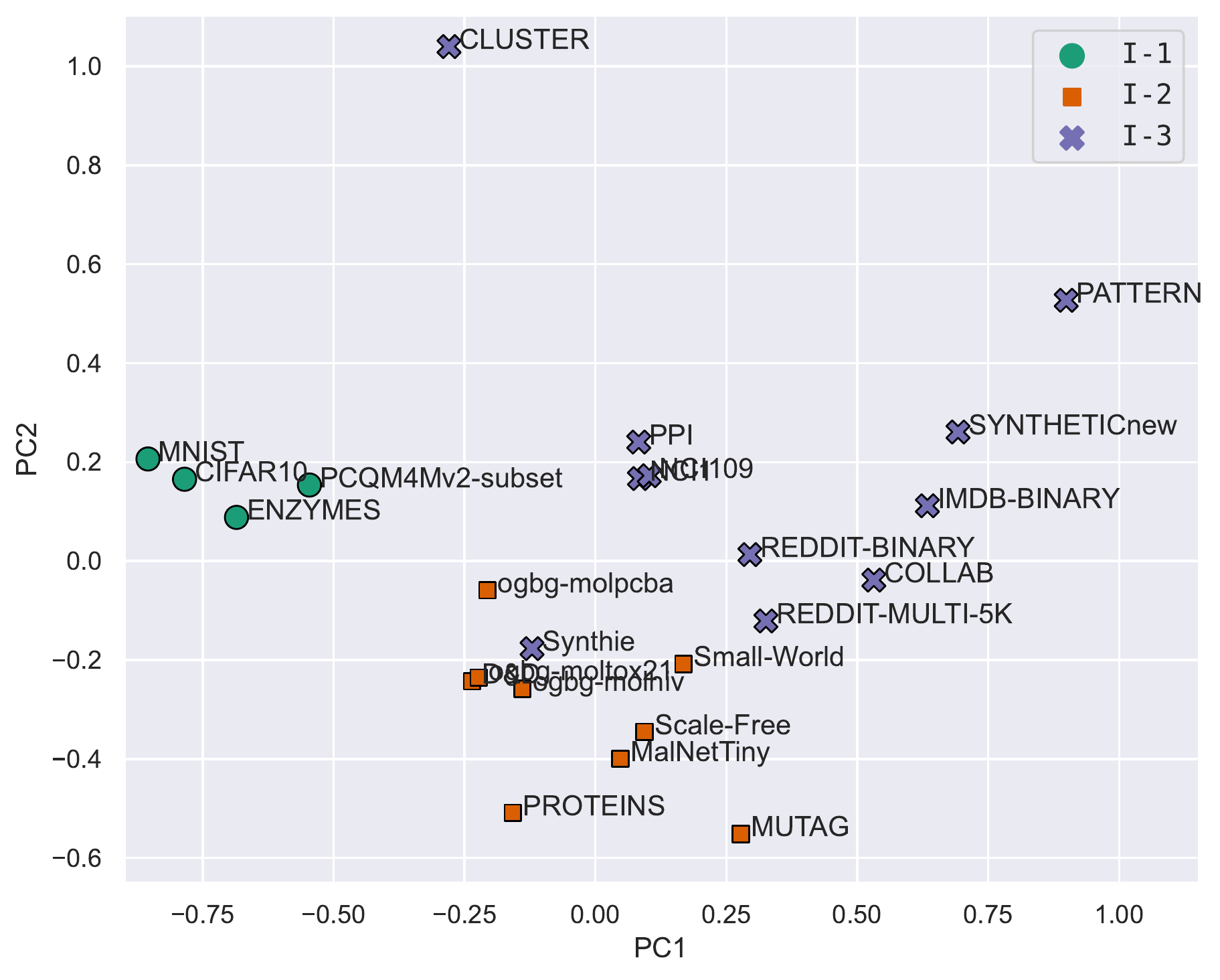}
         \hfill
         \includegraphics[align=c, trim=0 10 10 -2, clip, width=0.68\textwidth]{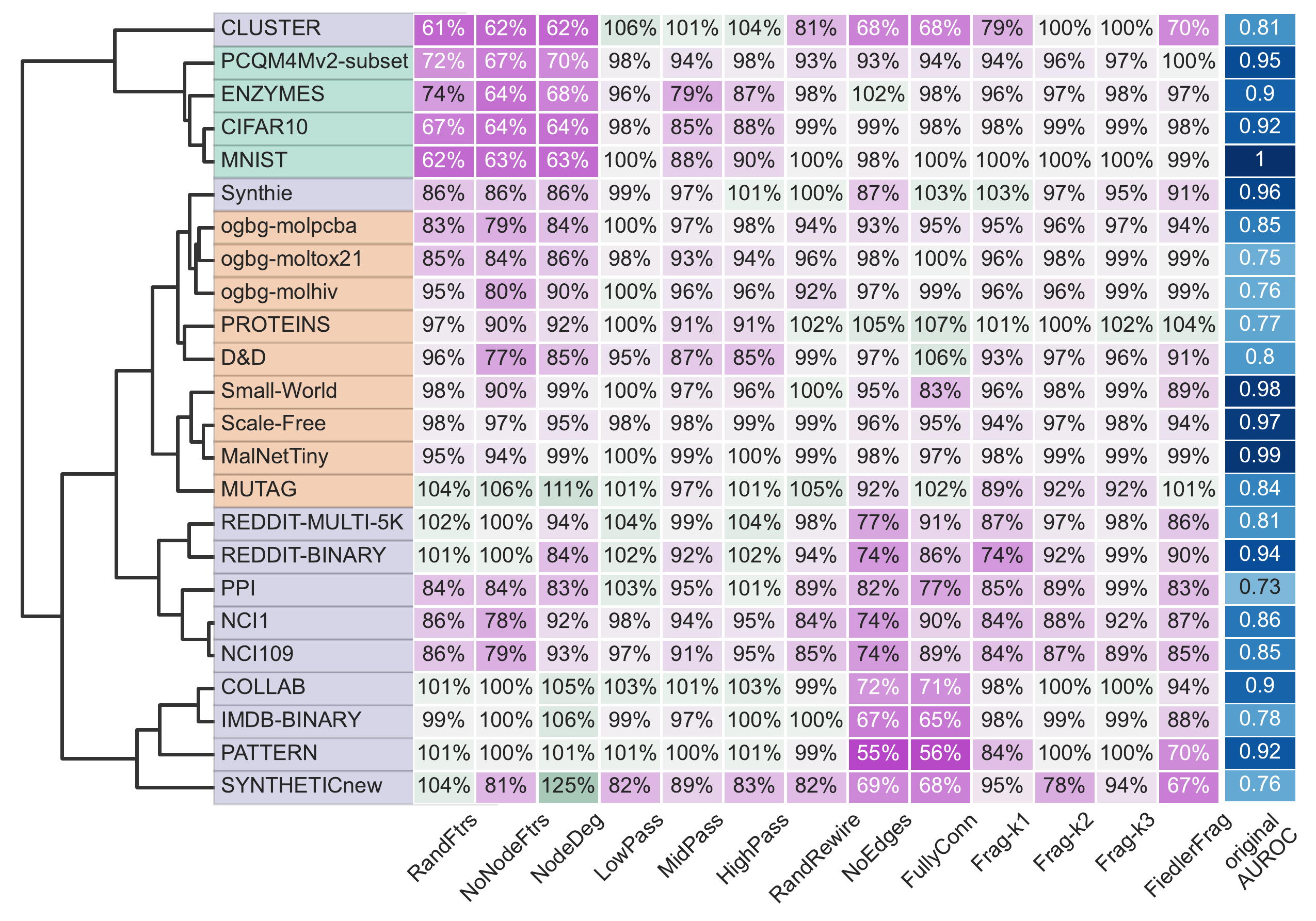}
     \end{adjustwidth}
     \vspace{-3pt}
     \caption{Sensitivity profiles by \textbf{2-Layer GIN} model; annotated by cluster assignment w.r.t. GCN model.}
     \label{fig:gl-taxo-gin_small}
     \end{subfigure}
\end{figure}
\begin{figure}[H]
\ContinuedFloat
\centering
    \vspace{-15pt}
     \begin{subfigure}[b]{\textwidth}
     \begin{adjustwidth}{-1.1 cm}{-1.1 cm}
         \centering
         \includegraphics[align=c, trim=0 0 0 0, clip, width=0.47\textwidth]{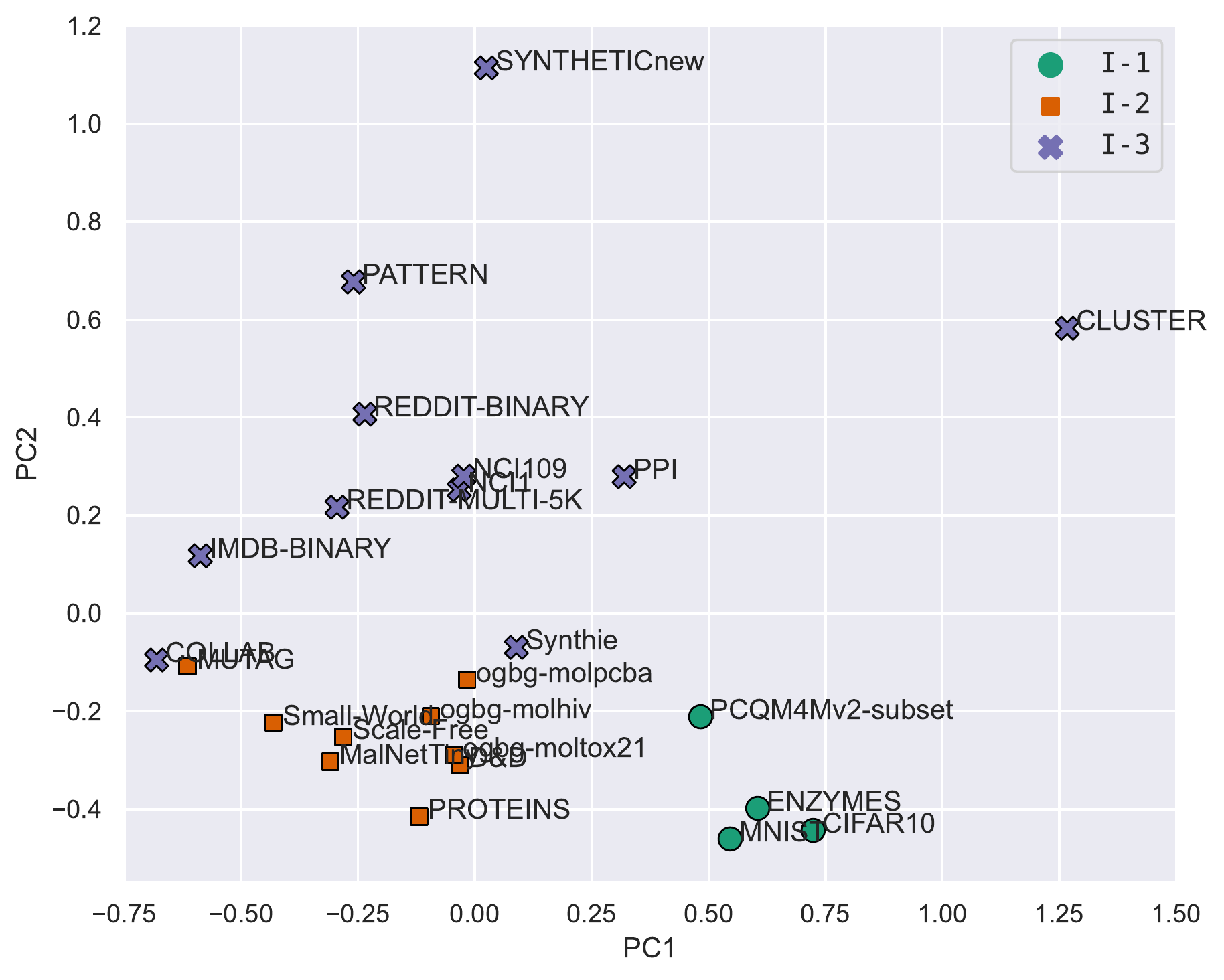}
         \hfill
         \includegraphics[align=c, trim=0 10 10 0, clip, width=0.68\textwidth]{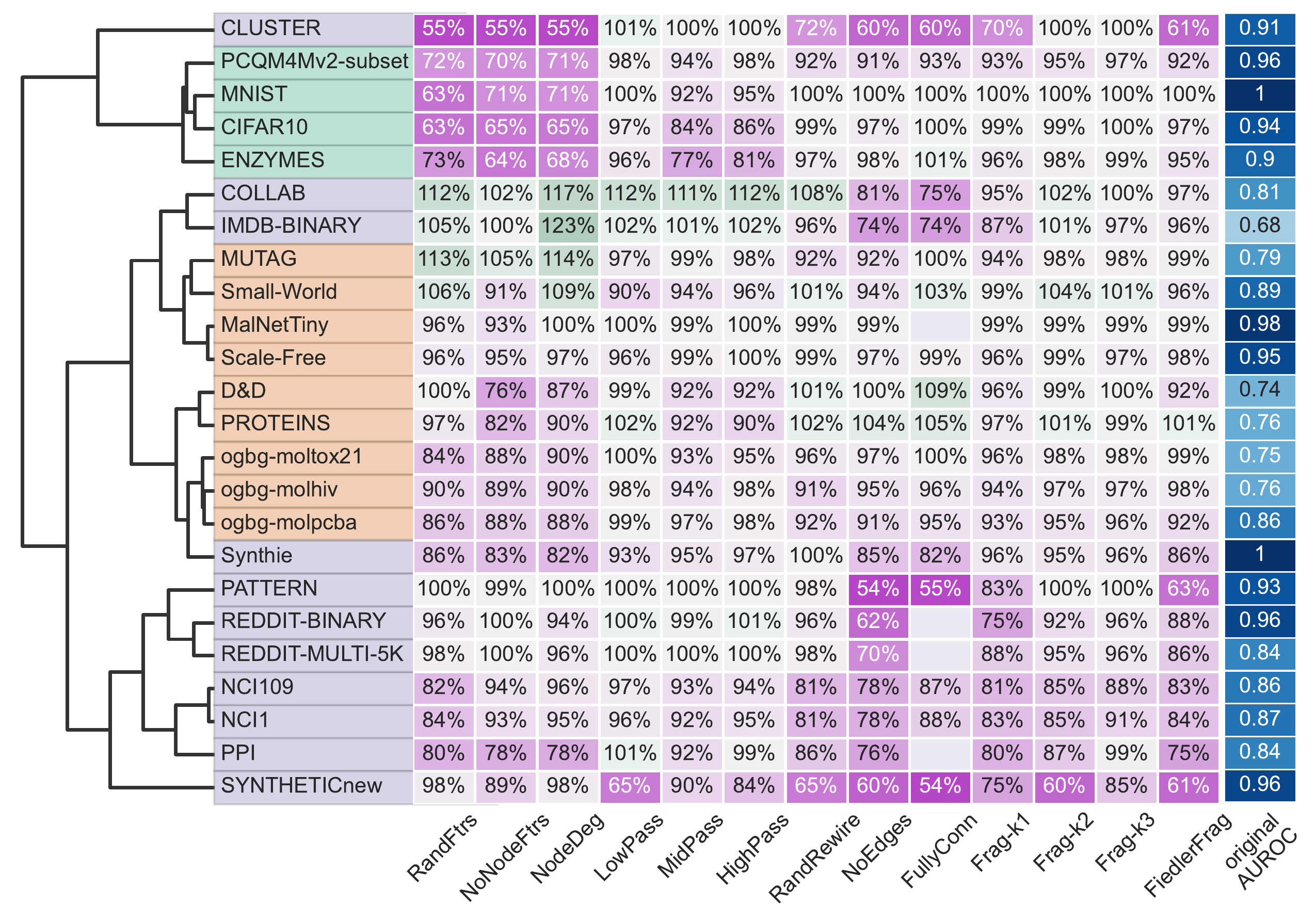}
     \end{adjustwidth}
     \vspace{-8pt}
     \caption{Sensitivity profiles by \textbf{ChebNet} model; annotated by cluster assignment w.r.t. GCN model.}
     \label{fig:gl-taxo-cheb}
     \end{subfigure}
     \begin{subfigure}[b]{\textwidth}
     \begin{adjustwidth}{-1.1 cm}{-1.1 cm}
         \centering
         \vspace{12pt}
         \includegraphics[align=c, trim=0 0 0 0, clip, width=0.47\textwidth]{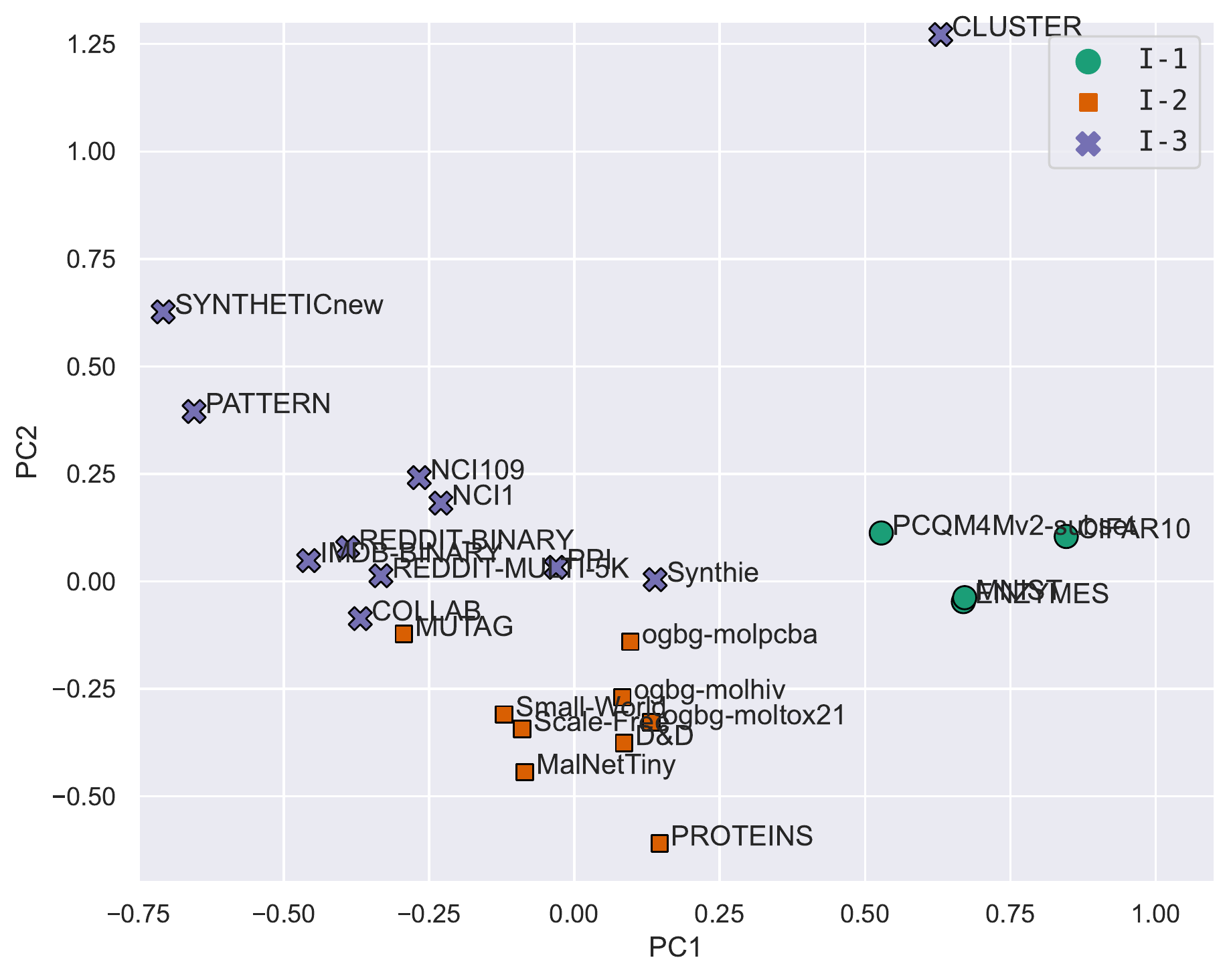}
         \hfill
         \includegraphics[align=c, trim=0 10 10 -2, clip, width=0.68\textwidth]{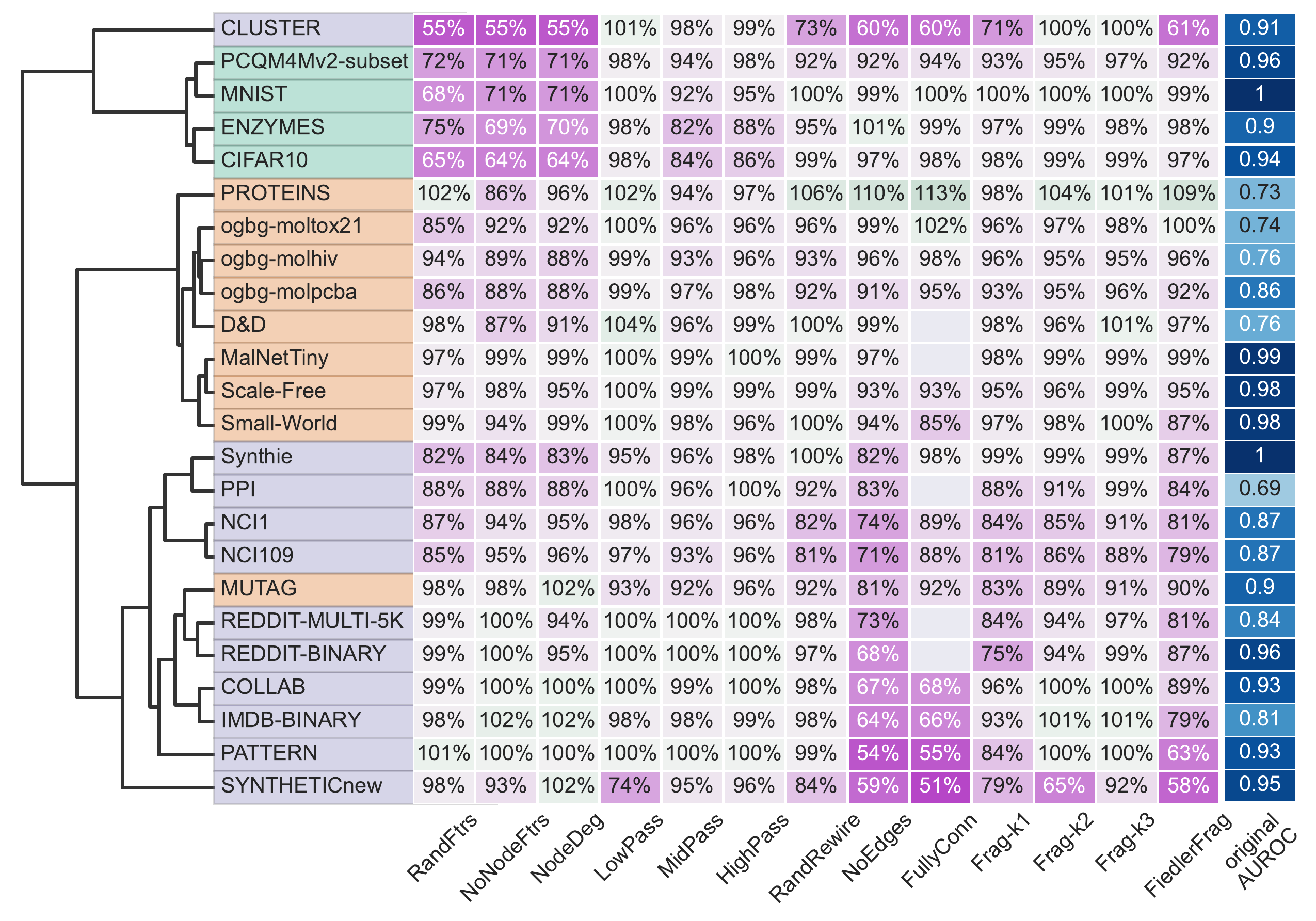}
     \end{adjustwidth}
     \vspace{-8pt}
     \caption{Sensitivity profiles by \textbf{GatedGCN} model; annotated by cluster assignment w.r.t. GCN model.}
     \label{fig:gl-taxo-gatedgcn}
     \end{subfigure}
     \begin{subfigure}[b]{\textwidth}
     \begin{adjustwidth}{-1.1 cm}{-1.1 cm}
         \centering
         \vspace{12pt}
         \includegraphics[align=c, trim=0 0 0 0, clip, width=0.47\textwidth]{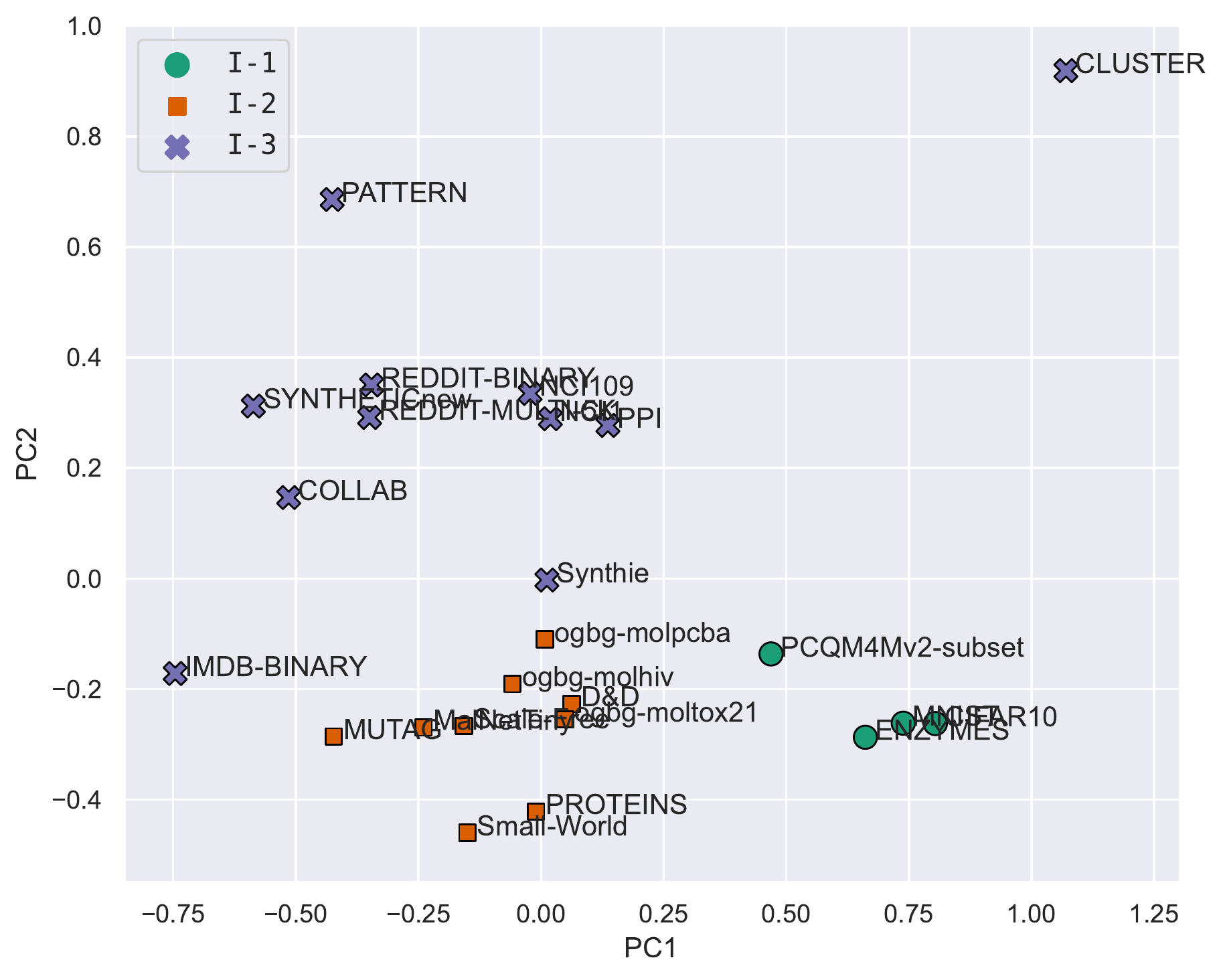}
         \hfill
         \includegraphics[align=c, trim=0 10 10 -2, clip, width=0.68\textwidth]{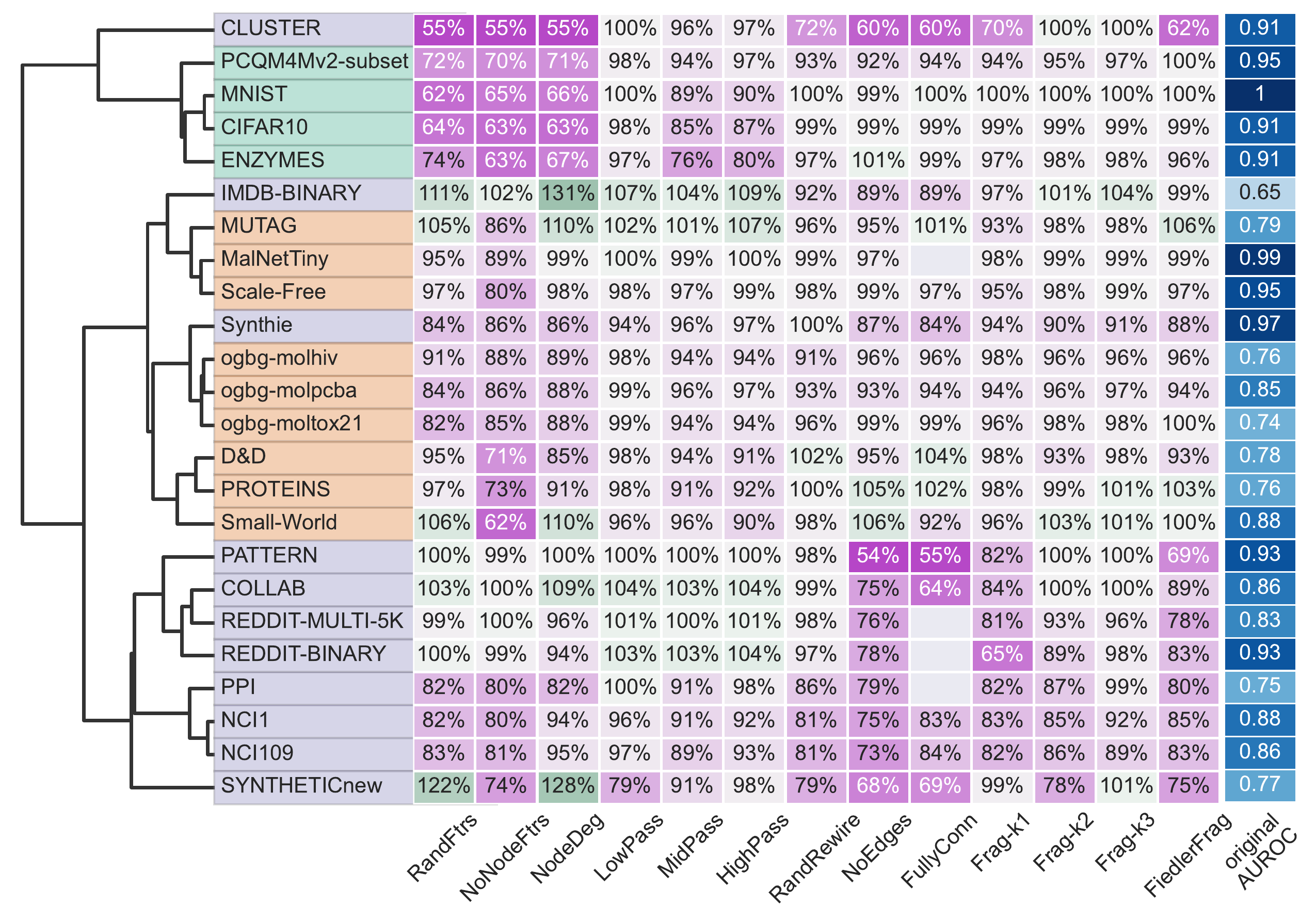}
     \end{adjustwidth}
     \vspace{-3pt}
     \caption{Sensitivity profiles by \textbf{GCNII} model; annotated by cluster assignment w.r.t. GCN model.}
     \label{fig:gl-taxo-gcn2}
     % \vspace*{8pt}
     \end{subfigure}
\begin{adjustwidth}{-1.1 cm}{-1.1 cm}
\hspace*{5pt}
\parbox{1.14\textwidth}{
\caption{Taxonomy of inductive graph learning datasets via graph perturbations. The categorization into 3 dataset clusters is stable across the following models with only minor deviations: (a) GCN, (b) GIN, (c) 2-Layer GIN, (d) ChebNet, (e) GatedGCN, (f) GCNII. Panel (a) left and right is as shown in Figure~\ref{fig:inductive-pca}~and~\ref{fig:gl-taxo}, respectively, shown here for ease of comparison. Missing performance ratios (due to out-of-memory error) are shown in gray.
}
\label{fig:gl-taxo-full}
}
\end{adjustwidth}
\vspace{20pt}
\end{figure}

\newpage
\subsection{Taxonomy of Transductive Benchmarks} \label{sec:transductive-results-details}

\myparagraph{All transductive datasets are relatively insensitive to structural perturbations.}
% \myparagraph{\T{\{1,2,3\}}: Inconsequence of structure.}
Unlike many of the inductive datasets that show significant reliance on the graph structure (\I{3}), the lowest performance achieved for a transductive dataset due to graph structure removal is still as high as 92\% (\ds{Flickr}), suggesting a weak dependence on the full graph structure.
Furthermore, on average, considering only the neighborhoods of up to 3-hops (\pert{Frag-k3}) nearly retains the full potential of the model (99\% ± 1.6\%), revealing the lack of long-range dependencies in these node-level datasets.
Such negligence of the full graph structure might be attributed to the limitations of the GCN expressivity and issues such as oversquashing \cite{topping2021understanding}. While these limitations are fundamentally true, our observation of long-range dependencies on some graph-level tasks like \ds{NCI}, coupled with our architecture being 5 layers deep with residual connections, indicate that our GCN model is capable of capturing non-local information in the 3-hop neighborhoods. Furthermore, our observed long-range \emph{independence} in transductive node-level datasets is consistent with the promising results presented by recent development of scalable GNNs that operate on subgraphs \cite{2019clustergcn,zeng2020graphsaint,fey2021gnnautoscale}, breaking or limiting long-range connections.

\myparagraph{\T{3}: Indifference to node and structure removal.}
%Apart from the structural indifference,
The datasets in \T{3} are relatively insensitive to perturbations of graph structure and also to the removal of node features (\pert{NoNodeFtrs} and \pert{NodeDeg}).
For example, the Amazon datasets (\ds{Am-Phot} and \ds{Am-Comp}) always achieve near perfect classification performance regardless of the perturbations applied, suggesting redundancy between node features and graph structure for the corresponding tasks.
For these datasets, in particular, \ds{GitHub}, \ds{Am}, and \ds{Twitch}, more sophisticated, or combinations of, perturbations might be needed to gauge their essential characteristics.

% \ds{WikiNet-squir} and \ds{WebKB-Tex}, although clustered into \T{3} due to the overall inconsequence of \pert{NoNodeFtrs} and structural perterbations, demonstrate some distinct characteristics. First, \ds{WikiNet-squir} shows a significant performance boost (+11\% AUROC) by replacing its node features (binary bag-or-words) with the node degree information. Such performance boost is reasonable because a web page linked by many other web pages is likely to receive more traffic, which is the corresponding task, than those linked by only a few. On the other hand, as we will discuss more in \T{1}, \ds{WebKB-Tex} considerably benefits from \pert{HighPass}, while \pert{LowPass} and \pert{MidPass} severely decrease its performance.

\myparagraph{\T{2}: Rich node features but substitutable for  structural (summary) information.}
\T{2} contains a broad spectrum of datasets from citation networks (\ds{CF}), social networks (\ds{Coau}, \ds{FBPP}, \ds{LFMA}), to web pages (\ds{WikiNet}, \ds{WikiCS}).
The considerable performance decrease due to node feature removal suggests the relevance of the node features for their tasks.
For example, it is not surprising that the binary bag-of-words features of \ds{CF} datasets provide relevant information to classify papers into different fields of research, as one might expect some keywords to appear more likely in one field than in another.
Furthermore, using the one-hot encoded node degrees (\pert{NodeDeg}) always results in better performance over \pert{NoNodeFtrs}. And in many cases such as Facebook (\ds{FBPP}), \pert{NodeDeg} nearly retains the baseline performance, suggesting the relevance of node degree information, as a form of structural summary, for the respective tasks.

\ds{WebKB-Tex}, although clustered into \T{2} is more of an outlier that does not clearly fit into any of the existing clusters. As we will discuss more in \T{1}, \ds{WebKB-Tex} considerably benefits from \pert{HighPass}, while \pert{LowPass} and \pert{MidPass} severely decrease its performance.

\myparagraph{\T{1}: Heterophilic datasets.}
Three of the four datasets in \T{1} (\ds{Actor}, \ds{WebKB-Cor}, and \ds{WebKB-Wis}) are commonly referred to as heterophilic datasets \cite{mostafa_local_2021,ma2021homophily}.
While \ds{WebKB-Tex} (\T{2}) is also known to be heterophilic, it is isolated from \T{1} mainly due to its insensitivity to node feature removal, suggesting the structure alone is sufficient for its prediction task.

Our results show that in heterophilic datasets such as \T{1} and \ds{WebKB-Tex}, \pert{LowPass} node feature filtering, realized by local aggregation (Eq.~\ref{eqn:wavelettransform}), significantly degrades the performance, unlike other homophilic datasets.
By contrast, \pert{HighPass} results in better performance than \pert{LowPass}. In the case of \ds{WekbKB-Tex}, \pert{HighPass} significantly improves the performance over the baseline.
%which is related to some recent findings that there are ``good'' types of heterophily \cite{ma2021homophily}, in which case it is possible to exploit the neighborhood structural patterns to infer the correct node labels.
This observation is related to recent findings \cite{ma2021homophily} that in the case of extreme heterophily, local information, this time in form of the neighborhood patterns, may suffice to infer the correct node labels.

Finally, despite heterophilic datasets \cite{ma2021homophily,alon2021bottleneck,topping2021understanding,mostafa_local_2021} attracting much recent attention, this type of datasets (\T{1} and \ds{WebKB-Tex}) is lacking in availability compared to the others (\T{\{2,3\}}), which exhibit homophily but with different levels of reliance on node features. Thus, there is a need to collect and generate more real-world heterophilic datasets.

\subsection{Correlations of Perturbations}\label{sec:pert-corr}

\begin{figure}[H]
    \centering
    \begin{subfigure}[t]{0.49\textwidth}
        \centering
        \includegraphics[width=\textwidth]{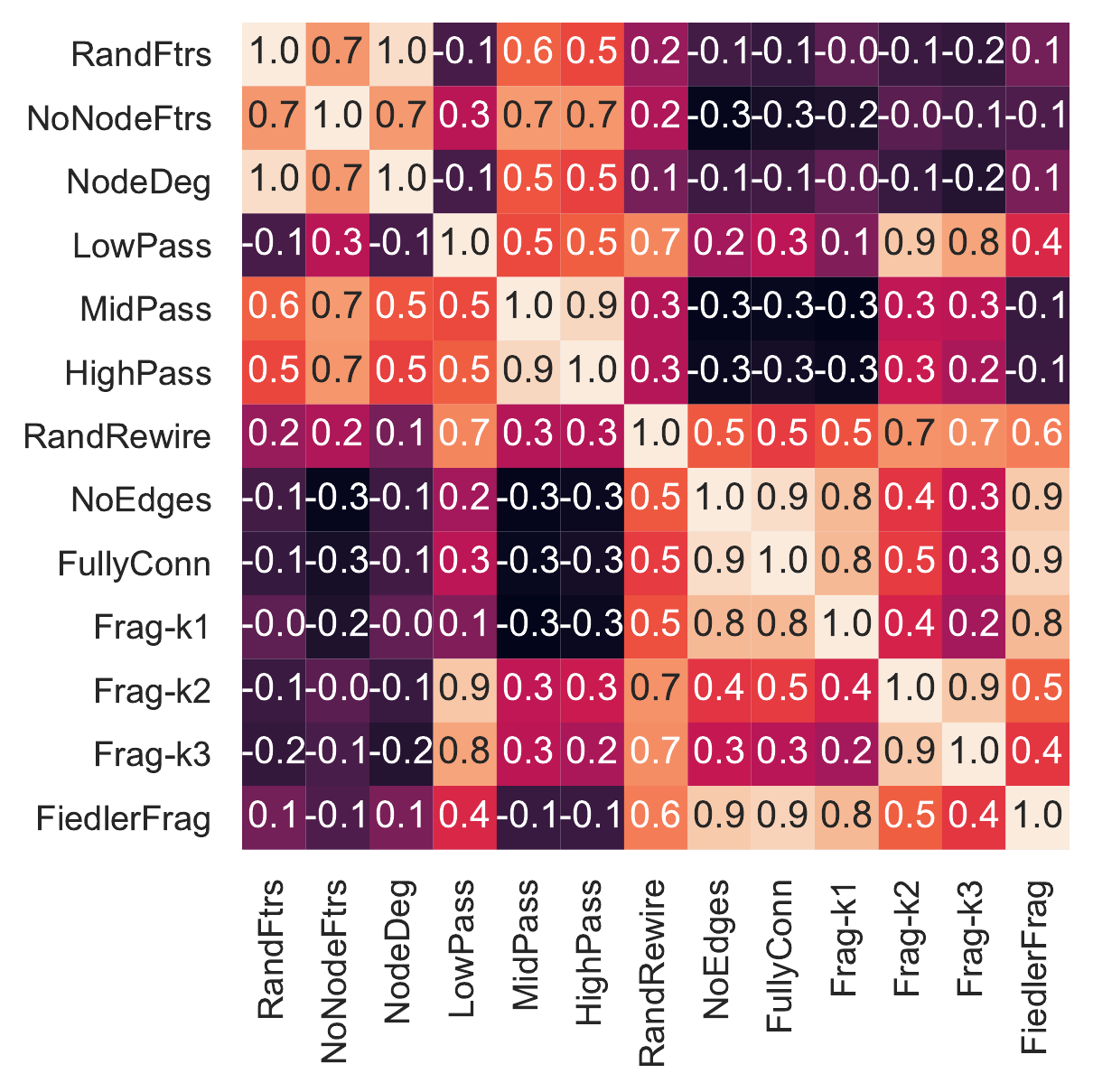}
        \caption{Inductive benchmarks}
        \label{fig:pert-corr-ind}
    \end{subfigure}
    \hfill
    \begin{subfigure}[t]{0.49\textwidth}
        \centering
        \includegraphics[width=\textwidth]{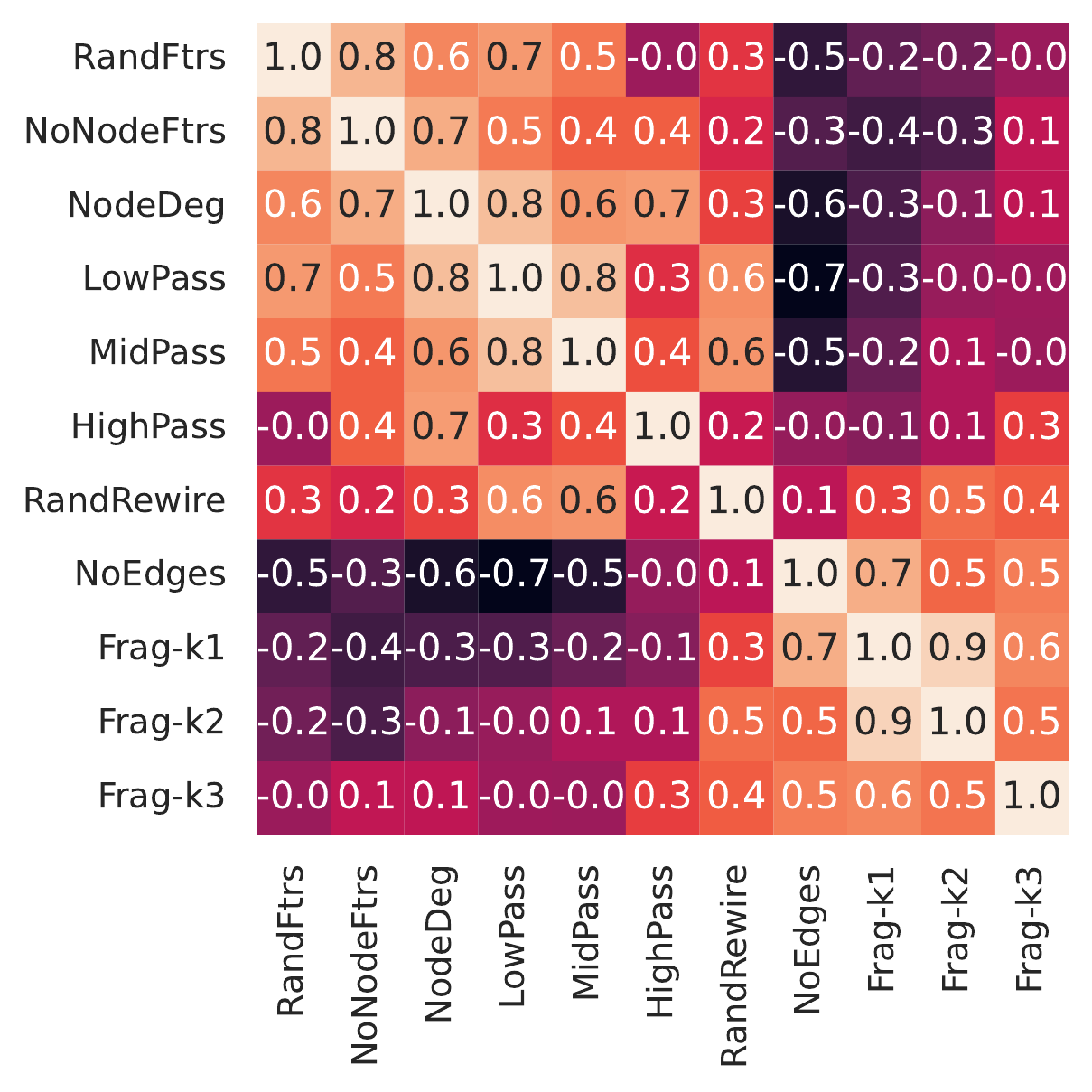}
        \caption{Transductive benchmarks}
        \label{fig:pert-corr-trans}
    \end{subfigure}
    \caption{Pearson correlation coefficients of the log2 performance fold change between different perturbations (w.r.t. a GCN model).}
    \label{fig:pert-corr}
\end{figure}

We compute the Pearson correlation between all pairs of perturbations based on the log2 performance fold change. The results in Figure~\ref{fig:pert-corr} indicate that many perturbations correlate with each other to some extend. For both transductive and inductive benchmarks, the perturbations roughly cluster into two groups, separating node feature perturbations (see Section~\ref{sec:nf-pert}) and graph structure perturbations (see Section~\ref{sec:gs-pert}). In particular, perturbations that replace the original node features with other less informative features, including \pert{RandFtrs}, \pert{NoNodeFtrs}, and \pert{NodeDeg}, highly correlate with one another (Pearson $r \ge 0.6$). Similarly, perturbations that severely break the graphs apart, including \pert{NoEdges}, \pert{Frag-k1}, and \pert{FiedlerFrag}, are highly correlated (Pearson $r \ge 0.8$).
% Note that, however, for inductive benchmarks, \pert{Frag-k2} and \pert{Frag-k3} highly correlate with each other (Pearson $r=0.9$), but not with other graph structure perturbations. We hypothesize that this is because 2 and 3-hop neighborhood fragmentation preserve much more information than \pert{Frag-k1} and other perturbations that more severely degrade the graph structure, and hence show a different characteristics.

%%%%%%%%%%%%%%%%%%%%%%%%%%%%%%%%%%%%%%%%%%%%%%%%%%%%%%%%%%%%%%%%%%%%%%%%%%%%%%%
\section{Graph Learning Benchmarks} \label{sec:datasets-details}
\setcounter{figure}{0}
\setcounter{table}{0}
%%%%%%%%%%%%%%%%%%%%%%%%%%%%%%%%%%%%%%%%%%%%%%%%%%%%%%%%%%%%%%%%%%%%%%%%%%%%%%%

\subsection{Inductive Datasets}

\noindent\textbf{\ds{MNIST} and \ds{CIFAR10}} \cite{dwivedi_benchmarking_2020} are derived from the well-known image classification datasets. The images are converted to graphs by SLIC superpixelization; node features are the average pixel coordinates and intensities; edges are constructed based on kNN criterion.

\noindent\textbf{\ds{PATTERN} and \ds{CLUSTER}} \cite{dwivedi_benchmarking_2020} are node-level inductive datasets generated from SBMs \cite{Holland83_SBM}. In PATTERN, the task is to identify nodes of a structurally specific subgraph; CLUSTER has a semi-supervised clustering task of predicting the true cluster assignment of nodes while observing only one labelled node per cluster.

\noindent\textbf{\ds{IMDB-BINARY}} \cite{yanardag2015DGK} is a dataset of ego-networks, where nodes represent actors/actresses and an edge between two nodes means that the two artists played in a movie together. The task is to determine which genre (action or romance) each ego-network belongs to.

\noindent\textbf{\ds{D\&D}} \cite{Dobson2003} is a protein dataset where each protein is represented by a graph with rich node feature set. The task is to classify proteins as enzymes or non-enzymes.

\noindent\textbf{\ds{ENZYMES}} \cite{Borgwardt2005} is a dataset of tertiary structures from six enzymatic classes (determined by Enzyme Commission numbers). Each node represents a secondary structure element (SSE), and has an edge between its three spatially closest nodes. Node features are the type of SSE, and the physical and chemical information.

\noindent\textbf{\ds{PROTEINS}} \cite{Borgwardt2005} is a modification of the \ds{D\&D} \cite{Dobson2003}; the task is the same but the protein graphs are generated as in \ds{ENZYMES}.

\noindent\textbf{\ds{NCI1} and \ds{NCI109}} \cite{wale2008} consist of graph representations of chemical compounds; each graph represents a molecule in which nodes represent atoms and edges represent atomic bonds. Atom types are one-hot encoded as node features. The tasks are to determine whether a given compound is active or inactive in inhibiting non-small cell lung cancer (\ds{NCI1}) or ovarian cancer (\ds{NCI109}).

\noindent\textbf{\ds{COLLAB}} \cite{yanardag2015DGK} is an ego-network dataset of researchers in three different fields of physics. Each graph is a researcher's ego-network, where nodes are researchers and an edge between two nodes means the two researchers have collaborated on a paper. The task is to determine which field a given researcher ego-network belongs to.

\noindent\textbf{\ds{REDDIT-BINARY} and \ds{REDDIT-MULTI-5K}} \cite{yanardag2015DGK} graphs are derived from Reddit communities (subreddits). These subreddits are Q\&A based or discussion-based. Each graph represents a set of interactions between users through posts and comments; nodes represent users while an edge implies an interaction between two users. The task for \ds{REDDIT-BINARY} is to determine whether the given interaction graph belongs to a Q\&A or discussion subreddit. In \ds{REDDIT-MULTI-5K}, the graphs are drawn from 5 specific subreddits instead, and the task is to predict the subreddit a graph belongs to.

\noindent\textbf{\ds{MUTAG}} \cite{debnath1991structure} is a dataset of Nitroaromatic compounds. Each compound is represented by a graph in which nodes represent atoms with their types one-hot encoded as node features, and edges represent atomic bonds. The task is to determine whether a given compound has mutagenic effects on Salmonella typhimurium bacteria.

\noindent\textbf{\ds{MalNet-Tiny}} \cite{freitas2021a} is a smaller version of MalNet dataset, consisting of function call graphs of various malware on Android systems using Local Degree Profiles as node features. In MalNet-Tiny, the task is constrained to classification into 5 different types of malware.

\noindent\textbf{\ds{ogbg-molhiv}, \ds{ogbg-molpcba}, \ds{ogbg-moltox21}} \cite{hu_open_2021} datasets, adopted from MoleculeNet~\cite{wu2018moleculenet}, are composed of molecular graphs, where nodes represent atoms and edges represent atomic bonds in-between. Node features include atom type and physical/chemical information such chirality and charge. The task is to classify molecules on whether they inhibit HIV replication (\ds{ogbg-molhiv}) or their toxicity on on 12 different targets such as receptors and stress response pathways in a multilabel classification setting (\ds{ogbg-moltox21}). In \ds{ogbg-molpcba} the task is 128-way multi-task binary classification derived from 128 bioassays from PubChem BioAssay.

\noindent\textbf{\ds{PCQM4Mv2-subset}} is our derivative of the OGB-LSC PCQM4Mv2~\cite{hu2021ogblsc} molecular dataset. The original task is a regression of a quantum physical property -- the HOMO-LUMO gap. For compatibility with our analysis, we quantized the regression task into 20-way classification task based on quantils of the training set. As true labels of the original ``test-dev'' and ``test-challange'' dataset splits are kept private by the OGB-LSC challenge organizers, and for efficiency of our analysis, we created a custom reduced splits as follows: \textit{train set}: random 10\% of the original train set; \textit{validation set}: another random 50,000 graphs from the original train set; \textit{test set}: the original validation set. The molecular graphs are featurized the same way as in \ds{ogbg-mol*} datasets.

\noindent\textbf{\ds{PPI}} \cite{zitnik2017predicting,hamilton2017inductive} dataset contains a collection of 24 tissue-specific protein-protein interaction networks derived from the STRING database~\cite{szklarczyk2015string} using tissue-specific gold-standards from~\cite{greene2015understanding}. 20 of the networks are used for training, 2 used for validation, and 2 used for testing. In each network, each protein (node) is associated with 50 different gene signatures as node features. The multi-label node classification task was to classify each gene (node) in a graph based on its gene ontology terms.

\noindent\textbf{\ds{SYNTHETICnew}} \cite{FeragenSyntheticNEW} is a dataset where each graph is based on a random graph $G$ with scalar node features drawn from the normal distribution. Two classes of graphs are generated from $G$ by randomly rewiring edges and permuting node attributes; the number of rewirings and permuted attributes are distinct for the two classes. Noise is added to the node features to make the tasks more difficult. The task is to determine which class a given graph belongs to.

\noindent\textbf{\ds{Synthie}} \cite{7837955} dataset is generated from two Erd\"os-R\'enyi graphs $G_{1, 2}$: Two sets of graphs $S_{1, 2}$ are then generated by randomly adding and removing edges from $G_{1, 2}$. Then, 10 graphs were sampled from these sets and connected by randomly adding edges, resulting in a single graph. Two classes of these graphs, $C_{1, 2}$ are generated by using distinct sampling probabilities for the two sets. The two classes are then in turn split into two by generating two sets of vectors $A$ and $B$; nodes from a given graph were appended a vector from $A$ as node features if they were sampled from $S_1$, and $B$ for $S_2$ for one class, and vice versa for the other. The task is to classify which of these four classes a given graph belongs to. 

\noindent\textbf{\ds{Small-world} and \ds{Scale-free}} \cite{you2020design} datasets are generated by tweaking graph generation parameters for the real-world-derived small-world \cite{watts1998} and scale-free \cite{2002} graphs. Graphs are generated using a range of Averaging Clustering Coefficient and Average Path Length parameters. In our experiments, clustering coefficients and PageRank scores constitute node features while the task is to classify graphs based on average path length, where the continuous path length variable is rendered discrete by 10-way binning.

\begin{table}[t]
\centering
\caption{Inductive benchmarks. All datasets are equipped with graph-level classification tasks, except PATTERN and CLUSTER that are equipped with inductive node-level classification tasks.}\label{table:graph_datasets}
% \scriptsize
\fontsize{8pt}{8pt}\selectfont
\begin{adjustwidth}{-2.5 cm}{-2.5 cm}\centering
\setlength\tabcolsep{4pt} % default value: 6pt
\rowcolors{1}{}{lightgray}
\renewcommand{\arraystretch}{1.2}
\begin{tabular}{ lrrrrrcc }\toprule
    \textbf{Dataset} & \hspace*{-2em}\textbf{\# Graphs} & \textbf{Avg \# Nodes} & \textbf{Avg \# Edges} & \textbf{\# Features} & \textbf{\# Classes} & \textbf{Predef. split} & \textbf{Ref.} \\
    \midrule
    MNIST       & 70,000 & 70.57  & 564.53  & 3 & 10 & Yes & \cite{dwivedi_benchmarking_2020} \\
    CIFAR10     & 60,000 & 117.63 & 941.07  & 5 & 10 & Yes & \cite{dwivedi_benchmarking_2020} \\
    PATTERN     & 14,000 & 118.89 & 6,078.57 & 3 & 2  & Yes & \cite{dwivedi_benchmarking_2020} \\
    CLUSTER     & 12,000 & 117.20 & 4,301.72 & 7 & 6  & Yes & \cite{dwivedi_benchmarking_2020} \\
    IMDB-BINARY & 1,000  & 19.77  & 96.53   & -- & 2  & No  & \cite{yanardag2015DGK} \\
    D\&D        & 1,178  & 284.32 & 715.66  & 89 & 2  & No  & \cite{Dobson2003} \\
    ENZYMES     & 600   & 32.63  & 62.14   & 21 & 6  & No  & \cite{Borgwardt2005} \\
    PROTEINS    & 1,113  & 39.06  & 72.82   & 4 & 2  & No  & \cite{Borgwardt2005} \\
    NCI1        & 4,110  & 29.87  & 32.3    & 37 & 2  & No  & \cite{wale2008} \\
    NCI109      & 4,127  & 29.68  & 32.13   & 38 & 2  & No  & \cite{wale2008} \\
    COLLAB      & 5,000  & 74.49  & 2,457.78 & -- & 3  & No  & \cite{yanardag2015DGK} \\
    REDDIT-BINARY & 2,000& 429.63 & 497.75  & -- & 2  & No  & \cite{yanardag2015DGK} \\
    REDDIT-MULTI-5K     & 4,999   & 508.52  & 594.87&  -- & 5  & No  & \cite{yanardag2015DGK} \\
    MUTAG       & 188   & 17.93  & 19.79   & 7 & 2  & No  & \cite{debnath1991structure} \\
    MalNet-Tiny       & 5,000   & 1,410.3  &  2,859.94   & 5 & 5  & No  &  \cite{freitas2021a}\\
    ogbg-molhiv & 41,127  & 25.5  & 27.5    & 9 sets & 2  & Yes  & \cite{hu_open_2021} \\
    ogbg-molpcba & 437,929  & 26.0  & 28.1    & 9 sets & 128x binary  & Yes  & \cite{hu_open_2021} \\
    ogbg-moltox21   & 7,831  & 18.6  & 19.3    & 9 sets & 12x binary  & Yes  & \cite{hu_open_2021} \\
    PCQM4Mv2-subset & 446,405 &14.1 &14.6 &9 sets &\scriptsize{quantized to} 20 &Custom &\cite{hu2021ogblsc} \\
    PPI& 24   & 2,372.67    & 66,136     & 50 & 121  & Yes  &  \cite{zitnik2017predicting}\\
    SYNTHETICnew& 300   & 100    & 196     & 1 & 2  & No  &  \cite{FeragenSyntheticNEW}\\
    Synthie     & 400   & 95     & 196.25  & 15 & 4  & No  & \cite{7837955} \\
    Small-world & 256   & 64     & 694     & 2 & 10 & No  & \cite{you2020design} \\
    Scale-free  & 256   & 64     & 501.56  & 2 & 10 & No  & \cite{you2020design} \\
    \bottomrule
\end{tabular}
\end{adjustwidth}\end{table}

\subsection{Transductive Node-level Datasets}

\noindent\textbf{\ds{WikiNet}} \cite{pei2020Geom-GCN} contains two networks of Wikipedia pages, where edges indicate mutual links between pages, and node features are bag-of-words (BOW) of informative nouns. The task is to classify the web pages based on their average monthly traffic bins.

\noindent\textbf{\ds{WebKB}} \cite{pei2020Geom-GCN} contains networks of web pages from different universities, where an (directed) edge is a hyperlink between two web pages, with BOW node features. The task is to classify the web pages into five categories: student, project, course, staff, and faculty.

\noindent\textbf{\ds{Actor}} \cite{pei2020Geom-GCN} is a network of actors, where an edge indicate co-occurrence of two actors on a same Wikipedia page, with node features represented by keywords about the actor on Wikipedia. The task is to classify the actor into one of five categories.

\noindent\textbf{\ds{WikiCS}} \cite{mernyei2020wikics} is a network of Wikipedia articles related to Computer Science, where edges represent hyperlinks between them, with 300-dimensional word embeddings of the articles. The task is to classify the articles into one of ten branches of the field.
    
\noindent\textbf{\ds{Flickr}} \cite{zeng2020graphsaint} is a network of images, where the edges represent common properties between images, such as locations, gallery, and comments by the same users. The node features are BOW of image descriptions, and the task is to predict one of 7 tags for an image.

\noindent\textbf{\ds{CF}} (CitationFull) \cite{bojchevski2018deep} contains citation networks where nodes are papers and edges represent citations, with node features as BOW of papers. The task is to classify the papers based on their topics.
    
\noindent\textbf{\ds{DzEu}} (DeezerEurope) \cite{rozemberczki2021multiscale} is a network of Deezer users from European countries where nodes are the users and edges are mutual follower relationships. The task is to predict the gender of users.

\noindent\textbf{\ds{LFMA}} (LastFMAsia) \cite{rozemberczki2021multiscale} is a network of LastFM users from Asian countries where edges are mutual follower relationships between them. The task is to predict the location of users.
    
\noindent\textbf{\ds{Amazon}} \cite{shchur2018pitfalls} contains Amazon Computers and Amazon Photo. They are segments of the Amazon co-purchase graph, where nodes represent goods, edges indicate that two goods are frequently bought together, node features are bag-of-words encoded product reviews, and class labels are given by the product category.
    
\noindent\textbf{\ds{Coau}} (Coauthor) \cite{shchur2018pitfalls} contains Coauthor CS and Coauthor Physics. They are co-authorship graphs based on the Microsoft Academic Graph from the KDD Cup 2016 challenge 3. Nodes are authors, and are connected by an edge if they co-authored a paper; node features represent paper keywords for each author’s papers, and class labels indicate most active fields of study for each author.

\noindent\textbf{\ds{Twitch}} \cite{rozemberczki2020characteristic} contains Twitch user-user networks of gamers who stream in a certain language where nodes are the users themselves and the edges are mutual friendships between them. The task is to to predict whether a streamer uses explicit language. Due to low baseline performance even after a thorough hyperparameter search, we excluded \ds{Twitch-RU} and \ds{Twitch-FR} from our main analysis.
    
\noindent\textbf{\ds{Github}} \cite{rozemberczki2020characteristic} is a network of GitHub developers where nodes are developers who have starred at least 10 repositories and edges are mutual follower relationships between them. The task is to predict whether the user is a web or a machine learning developer.
    
\noindent\textbf{\ds{FBPP}} (FacebookPagePage) \cite{rozemberczki2020characteristic} is a network of verified Facebook pages that liked each other, where nodes correspond to official Facebook pages, edges to mutual likes between sites. The task is multi-class classification of the site category.

\begin{table}[t]
\centering
\small
\caption{Transductive benchmarks with node-level classification tasks.}\label{table:node_datasets}
\fontsize{8pt}{8pt}\selectfont
\rowcolors{3}{}{lightgray}
\renewcommand{\arraystretch}{1.2}
\begin{tabular}{ lrrrrcc }\toprule
    \multirow{2}{*}{\textbf{Dataset}} & \multirow{2}{*}{\textbf{\# Nodes}} & \multirow{2}{*}{\textbf{\# Edges}} & \textbf{\# Node} & \textbf{\# Pred.} & \textbf{Predef.} & \multirow{2}{*}{\textbf{Ref.}}\\
    & & & \textbf{feat.} & \textbf{classes} & \textbf{split} & \\\midrule
    % Dataset & \# Nodes & \# Edges & \# Features & \# Classes & Reference\\
     WikiNet-cham & 2,277 & 72,202 & 128 & 5 & Yes & \cite{pei2020Geom-GCN}\\
     WikiNet-squir & 5,201 & 434,146 & 128 & 5 & Yes & \cite{pei2020Geom-GCN}\\
     WebKB-Cor & 183 & 298 & 1,703 & 10 & Yes & \cite{pei2020Geom-GCN}\\
     WebKB-Wis & 251 & 515 & 1,703 & 10 & Yes & \cite{pei2020Geom-GCN}\\
     WebKB-Tex & 183 & 325 & 1,703 & 10 & Yes & \cite{pei2020Geom-GCN}\\
     Actor & 7,600 & 30,019 & 932 & 10 & Yes & \cite{pei2020Geom-GCN}\\
     WikiCS & 11,701 & 297,110 & 300 & 10 & Yes & \cite{mernyei2020wikics}\\
     Flickr & 89,250 & 899,756 & 500 & 7 & Yes & \cite{zeng2020graphsaint}\\
     CF-Cora & 19,793 & 126,842 & 8,710 & 70 & No & \cite{bojchevski2018deep}\\
     CF-CoraML & 2,995 & 16,316 & 2,879 & 7 & No & \cite{bojchevski2018deep}\\
     CF-CiteSeer & 4,230 & 10,674 & 602 & 6 & No & \cite{bojchevski2018deep}\\
     CF-DBLP & 17,716 & 105,734 & 1,639 & 4 & No & \cite{bojchevski2018deep}\\
     CF-PubMed & 19,717 & 88,648 & 500 & 3 & No & \cite{bojchevski2018deep}\\
     DzEu & 28,281 & 185,504 & 128 & 2 & No & \cite{rozemberczki2021multiscale}\\
     LFMA & 7,624 & 55,612 & 128 & 18 & No & \cite{rozemberczki2021multiscale}\\
     Am-Comp & 13,752 & 491,722 & 767 & 10 & No & \cite{shchur2018pitfalls}\\
     Am-Phot & 7,650 & 238,162 & 745 & 8 & No & \cite{shchur2018pitfalls}\\
     Coau-CS & 18,333 & 163,788 & 6,805 & 15 & No & \cite{shchur2018pitfalls}\\
     Coau-Phy & 34,493 & 495,924 & 8,415 & 5 & No & \cite{shchur2018pitfalls}\\
     Twitch-EN & 7,126 & 77,774 & 128 & 2 & No & \cite{rozemberczki2020characteristic}\\
     Twitch-ES & 4,648 & 123,412 & 128 & 2 & No & \cite{rozemberczki2020characteristic}\\
     Twitch-DE & 9,498 & 315,774 & 128 & 2 & No & \cite{rozemberczki2020characteristic}\\
     Twitch-PT & 1,912 & 64,510 & 128 & 2 & No & \cite{rozemberczki2020characteristic}\\
     Github & 37,700 & 578,006 & 128 & 2 & No & \cite{rozemberczki2020characteristic}\\
     FBPP & 22,470 & 342,004 & 128 & 4 & No & \cite{rozemberczki2020characteristic}\\
     \bottomrule
\end{tabular}
\end{table}

%%%%%%%%%%%%%%%%%%%%%%%%%%%%%%%%%%%%%%%%%%%%%%%%%%%%%%%%%%%%%%%%%%%%%%%%%%%%%%%
\section{Distribution of Classical Graph Properties in Benchmarking Datasets} \label{sec:datasets-graph-props}
\setcounter{figure}{0}
\setcounter{table}{0}
%%%%%%%%%%%%%%%%%%%%%%%%%%%%%%%%%%%%%%%%%%%%%%%%%%%%%%%%%%%%%%%%%%%%%%%%%%%%%%%

In this work we use \textit{perturbation sensitivity profiles} derived from a GNN's prediction performance in order to gauge \textit{how task-related information} is encoded in the graph datasets. In this section we explore an alternative approach. We analyze classical graph properties in multiple datasets and their classes to investigate whether we can establish a meaningful taxonomy without a dependence on a particular GNN method, while using well-established graph properties.

A static analysis of the graph properties alone is insufficient without taking into account the prediction task as well. The graph domain that a dataset $X$ is sampled from (e.g., drug-like molecules, proteins, ego networks, citation networks) may exhibit varying range of properties (e.g., density, node degree distribution, local/global clustering coefficients, number of triangles, graph diameter, girth, maximum clique, etc.), however these do not take into account node features in attributed graphs, and could be irrelevant to the prediction task $Y$. Therefore, we look at the difference in graph properties compared among the individual classes of  $Y$.

\begin{table}[t]
\centering
\caption{Classical graph properties among positive and negative classes of 9 graph-classification datasets. The difference between datasets dominates within-dataset differences between classes.}
\label{table:classical_props}
\fontsize{8pt}{8pt}\selectfont 
\setlength\tabcolsep{4pt} % default value: 6pt
\begin{adjustwidth}{-2.5 cm}{-2.5 cm}\centering
\rowcolors{3}{}{lightgray}
\renewcommand{\arraystretch}{1.2}
\begin{tabular}{lrrrrrrrrrr}\toprule
    &Num. &Num. &\multirow{2}{*}{Density} &\multirow{2}{*}{Connectivity} &\multirow{2}{*}{Diameter} &Approx. &\multirow{2}{*}{Centrality} &Cluster. &Num. \\
&nodes &edges & & & &max clique & &coeff. &triangles \\\midrule
    IMDB-BINARY (class=0) &20.11 &96.78 &0.559 &3.828 &1.838 &10.30 &0.559 &0.943 &307.73 \\
    IMDB-BINARY (class=1) &19.43 &96.29 &0.482 &3.388 &1.884 &10.01 &0.482 &0.951 &476.25 \\
    REDDIT-BINARY (class=0) &641.25 &735.95 &0.012 &0.556 &5.646 &3.22 &0.012 &0.054 &35.96 \\
    REDDIT-BINARY (class=1) &218.00 &259.56 &0.032 &0.423 &3.778 &2.95 &0.032 &0.041 &13.71 \\
    D\&D (class=0) &341.88 &870.23 &0.019 &1.110 &20.843 &4.95 &0.019 &0.479 &617.07 \\
    D\&D (class=1) &183.72 &449.43 &0.040 &1.140 &17.460 &4.79 &0.040 &0.480 &302.55 \\
    PROTEINS (class=0) &50.00 &94.06 &0.142 &1.196 &13.837 &3.85 &0.142 &0.473 &34.30 \\
    PROTEINS (class=1) &22.94 &41.52 &0.315 &1.420 &7.278 &3.80 &0.315 &0.575 &17.24 \\
    NCI1 (class=0) &25.65 &27.65 &0.100 &0.924 &11.265 &2.02 &0.100 &0.002 &0.03 \\
    NCI1 (class=1) &34.07 &36.94 &0.078 &0.796 &11.917 &2.05 &0.078 &0.004 &0.07 \\
    NCI109 (class=0) &25.61 &27.61 &0.100 &0.913 &11.061 &2.02 &0.100 &0.002 &0.02 \\
    NCI109 (class=1) &33.69 &36.59 &0.079 &0.794 &11.644 &2.05 &0.079 &0.004 &0.07 \\
    MUTAG (class=0) &13.94 &14.62 &0.169 &1.000 &7.016 &2.00 &0.169 &0.000 &0.00 \\
    MUTAG (class=1) &19.94 &22.40 &0.123 &1.000 &8.824 &2.00 &0.123 &0.000 &0.00 \\
    SYNTHETICnew (class=0) &100.00 &196.42 &0.040 &0.993 &7.333 &3.00 &0.040 &0.024 &5.39 \\
    SYNTHETICnew (class=1) &100.00 &196.08 &0.040 &0.993 &7.213 &3.00 &0.040 &0.022 &4.54 \\
    ogbg-molhiv (class=0) &25.20 &27.13 &0.104 &0.931 &11.016 &2.02 &0.104 &0.002 &0.03 \\
    ogbg-molhiv (class=1) &34.18 &36.69 &0.084 &0.824 &12.183 &2.01 &0.084 &0.001 &0.01 \\
    \bottomrule
\end{tabular}
\end{adjustwidth}\end{table}

\begin{wrapfigure}[28]{R}{0.45\textwidth}
% \begin{figure}[t]
    \centering
    \includegraphics[width=0.46\textwidth]{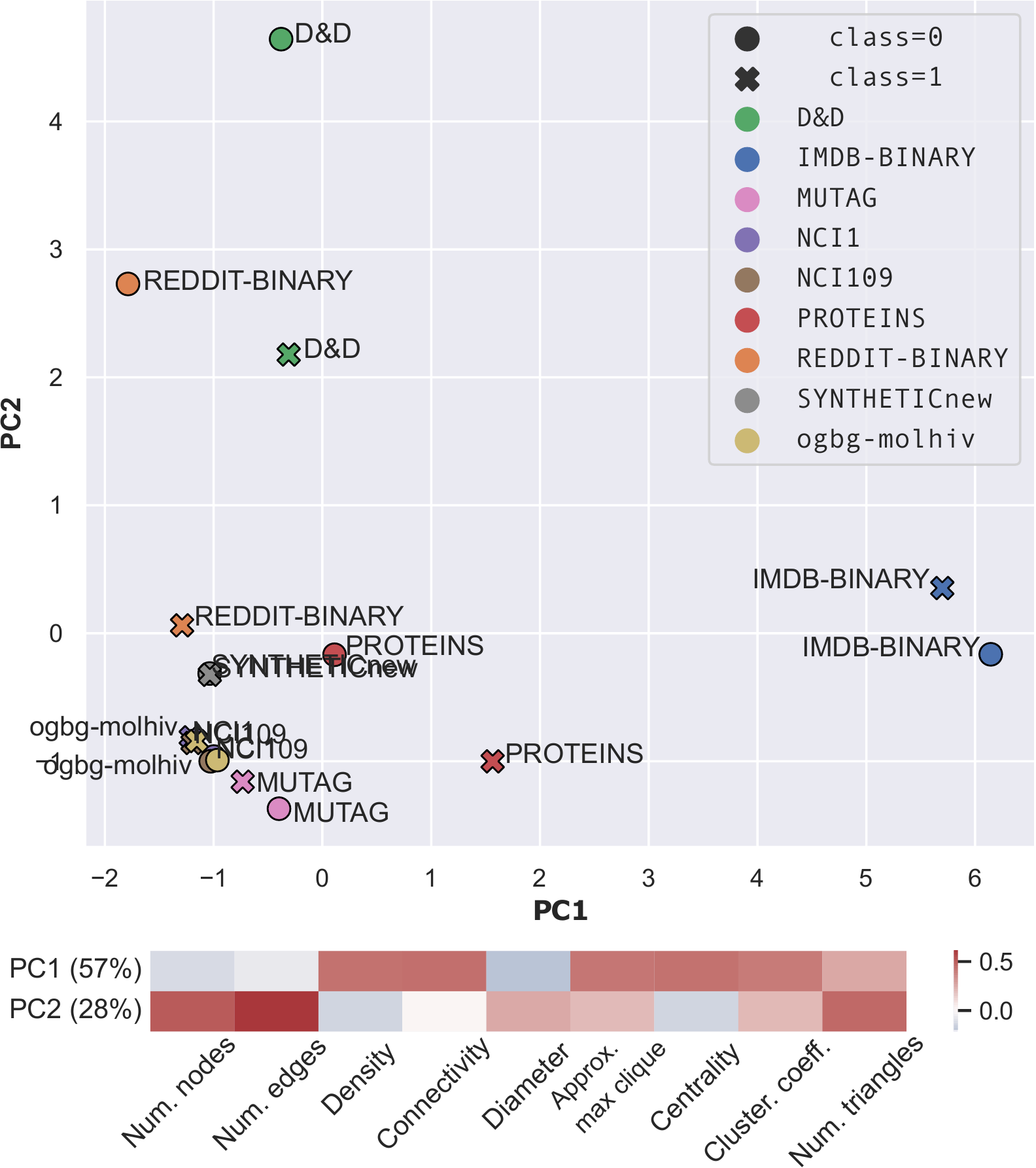}
    \vspace{-2pt}
    \caption{PCA plot of 9 binary graph-level classification datasets represented by their per-class graph properties. In the bottom, the loadings of the first two principal components are shown. 
    }
    \label{fig:pca_classical_props}
% \end{figure}
\end{wrapfigure}

Particularly, we look at all 9 inductive binary-classification datasets from our dataset selection (Table \ref{table:graph_datasets}). Within each class (the negative and positive label) of these 9 datasets we computed the average value of 9 graph properties computed by the NetworkX package~\cite{hagberg2008exploring}. The results are presented in Table~\ref{table:classical_props} and Figure~\ref{fig:pca_classical_props}.
Primarily, the computed graph properties vary more between datasets than between classes. The marginal graph properties of the positive and negative class are very similar to each other, especially for the SYNTHETICnew dataset. The largest difference between the classes appears to be the average size of the graphs, which is captured by the average number of nodes and edges. Therefore we argue that basing a taxonomy on dataset or class-level marginal graph properties is grossly insufficient as it completely fails to capture the nature of the prediction task.

Alternatively, one could conduct a correlation analysis between classical graph properties (averaged per class) and the outcome $Y$. However, that would again only take into account the marginal properties, assume linear relationship (as correlation captures only a linear relationship), and would rely on a fixed set of computable graph properties. These appear to be fundamental limitations compared to the perturbation analysis presented in the main text, that would result in a grossly skewed taxonomy.

%%%%%%%%%%%%%%%%%%%%%%%%%%%%%%%%%%%%%%%%%%%%%%%%%%%%%%%%%%%%%%%%%%%%%%%%%%%%%%%
% \section{Performance variation of \pert{Frag-k} perturbations} \label{sec:fragk_sensitivity}
% \section{Results from \pert{Frag-k} perturbations are stable across random seeds} 
\section{Impact of random initialization on \pert{Frag-k} perturbations} \label{sec:fragk_sensitivity}
\setcounter{figure}{0}
\setcounter{table}{0}
%%%%%%%%%%%%%%%%%%%%%%%%%%%%%%%%%%%%%%%%%%%%%%%%%%%%%%%%%%%%%%%%%%%%%%%%%%%%%%%

% We show that the \pert{Frag-k} perturbation results in relatively stable performance across ten different random seeds, despite it
Our \pert{Frag-k} perturbation is potentially sensitive to the random initializations of the initial seed nodes used in the fragmentation procedure. To measure this sensitivity of \pert{Frag-k} perturbations to node initializations, we computed the variance of AUROC results across ten experiments with different random seeds for both GCN and GIN models. Here, we analysed five datasets, the performance on which was significantly altered by \pert{Frag-k} in the original analysis, namely, CLUSTER, PATTERN, PPI, Synthie, and SYNTHETICnew.  The variances are within 5\%, with the only exception being \ds{SYNTHETICnew}. We hypothesize that this is due to the randomness of the constructions of the \ds{SYNTHETICnew} dataset. Thus, overall, the \pert{Frag-k} approach is sufficiently stable for datasets whose constructions involve little randomness.

\begin{table}[ht]
\centering
\caption{Variances of AUROC across ten different random seeds for \pert{Frag-k} for GCN.}
\label{table:fragk_sensitivity_gcn}
\fontsize{8pt}{8pt}\selectfont 
\setlength\tabcolsep{4pt} % default value: 6pt
\begin{adjustwidth}{-2.5 cm}{-2.5 cm}\centering
\rowcolors{3}{}{lightgray}
\renewcommand{\arraystretch}{1.2}
    \begin{tabular}{lrrrr}\toprule
        \textbf{Dataset} & \textbf{Perturbation} & \textbf{AUROC Avg.} & \textbf{AUROC Std.} & \textbf{AUC Std./Avg. (\%)} \\ \midrule
        CLUSTER & \pert{Frag-k1} & 0.637 & 0.001 & 0.165 \\
        CLUSTER & \pert{Frag-k2} & 0.913 & 0.000 & 0.039 \\
        CLUSTER & \pert{Frag-k3} & 0.913 & 0.000 & 0.037 \\
        % CLUSTER & none & 0.913 & 0.000 & 0.041 \\
        PATTERN & \pert{Frag-k1} & 0.769 & 0.001 & 0.095 \\
        PATTERN & \pert{Frag-k2} & 0.933 & 0.000 & 0.016 \\
        PATTERN & \pert{Frag-k3} & 0.933 & 0.000 & 0.021 \\
        % PATTERN & none & 0.933 & 0.000 & 0.023 \\
        PPI & \pert{Frag-k1} & 0.620 & 0.003 & 0.529 \\
        PPI & \pert{Frag-k2} & 0.647 & 0.012 & 1.807 \\
        PPI & \pert{Frag-k3} & 0.720 & 0.011 & 1.519 \\
        % PPI & none & 0.739 & 0.007 & 0.882 \\
        SYNTHETICnew & \pert{Frag-k1} & 0.704 & 0.126 & 17.908 \\
        SYNTHETICnew & \pert{Frag-k2} & 0.533 & 0.078 & 14.701 \\
        SYNTHETICnew & \pert{Frag-k3} & 0.715 & 0.089 & 12.492 \\
        Synthie & \pert{Frag-k1} & 0.962 & 0.015 & 1.581 \\
        Synthie & \pert{Frag-k2} & 0.870 & 0.029 & 3.334 \\
        Synthie & \pert{Frag-k3} & 0.876 & 0.036 & 4.164 \\
     \bottomrule
    \end{tabular}
\end{adjustwidth}
\end{table}

\begin{table}[ht]
\centering
\caption{Variances of AUROC across ten different random seeds for \pert{Frag-k} for GIN.}
\label{table:fragk_sensitivity_gin}
\fontsize{8pt}{8pt}\selectfont 
\setlength\tabcolsep{4pt} % default value: 6pt
\begin{adjustwidth}{-2.5 cm}{-2.5 cm}\centering
\rowcolors{3}{}{lightgray}
\renewcommand{\arraystretch}{1.2}
    \begin{tabular}{lrrrr}\toprule
        \textbf{Dataset} & \textbf{Perturbation} & \textbf{AUROC Avg.} & \textbf{AUROC Std.} & \textbf{AUC Std./Avg. (\%)} \\ \midrule
        CLUSTER & \pert{Frag-k1} & 0.643 & 0.001 & 0.162 \\
        CLUSTER & \pert{Frag-k2} & 0.910 & 0.001 & 0.101 \\
        CLUSTER & \pert{Frag-k3} & 0.910 & 0.001 & 0.130 \\
        % CLUSTER & none & 0.909 & 0.001 & 0.078 \\
        PATTERN & \pert{Frag-k1} & 0.780 & 0.001 & 0.091 \\
        PATTERN & \pert{Frag-k2} & 0.934 & 0.000 & 0.013 \\
        PATTERN & \pert{Frag-k3} & 0.934 & 0.000 & 0.019 \\
        % PATTERN & none & 0.934 & 0.000 & 0.017 \\
        PPI & \pert{Frag-k1} & 0.617 & 0.002 & 0.376 \\
        PPI & \pert{Frag-k2} & 0.644 & 0.009 & 1.476 \\
        PPI & \pert{Frag-k3} & 0.704 & 0.013 & 1.843 \\
        % PPI & none & 0.721 & 0.009 & 1.251 \\ 
        SYNTHETICnew & \pert{Frag-k1} & 0.708 & 0.081 & 11.407 \\
        SYNTHETICnew & \pert{Frag-k2} & 0.532 & 0.071 & 13.276 \\
        SYNTHETICnew & \pert{Frag-k3} & 0.757 & 0.064 & 8.411 \\
        Synthie & \pert{Frag-k1} & 0.985 & 0.008 & 0.810 \\
        Synthie & \pert{Frag-k2} & 0.945 & 0.011 & 1.213 \\
        Synthie & \pert{Frag-k3} & 0.920 & 0.025 & 2.677 \\
     \bottomrule
    \end{tabular}
\end{adjustwidth}
\end{table}

\end{document}